\pdfoutput=1
\documentclass{article}
\usepackage{iclr2027_arxiv,times}

\usepackage{amsmath,amsfonts,bm}

\def\eqref#1{equation~\ref{#1}}

\def\1{\bm{1}}

\DeclareMathAlphabet{\mathsfit}{\encodingdefault}{\sfdefault}{m}{sl}
\SetMathAlphabet{\mathsfit}{bold}{\encodingdefault}{\sfdefault}{bx}{n}

\usepackage{hyperref}
\hypersetup{hidelinks,pdftitle={Pretraining Transformers with Quantized Softmax in Attention}}
\usepackage{url}
\usepackage[T1]{fontenc}
\usepackage[utf8]{inputenc}
\usepackage{amsmath,amssymb,array,booktabs,float,graphicx,microtype,multirow,enumitem}
\usepackage[export]{adjustbox}
\usepackage{placeins}
\usepackage{longtable}
\renewcommand{\eqref}[1]{\textup{(\ref{#1})}}
\usepackage{xurl}
\usepackage{xspace}
\usepackage{dsfont}
\newcommand{\ind}{\mathds{1}}
\newcommand{\sg}{\operatorname{sg}}
\newcommand{\FW}{FWM\xspace}
\newcommand{\CH}{Nearest\xspace}
\newcommand{\dNLL}{\Delta\mathrm{NLL}}
\makeatletter
\newcommand{\rbnset}[2]{\expandafter\def\csname rbn@#1\endcsname{#2}}
\newcommand{\rbn}[1]{\ifcsname rbn@#1\endcsname\csname rbn@#1\endcsname\else\PackageError{rbn}{undefined number #1}{}\fi}
\makeatother
\rbnset{E1b.final3}{$+0.007$}%
\rbnset{E1b.final}{$+0.0069$}%
\rbnset{E1b.final.ci}{[$+0.0055$, $+0.0083$]}%
\rbnset{E1b.at125}{$+0.018$}%
\rbnset{E1b.at250}{$+0.004$}%
\rbnset{E1.K4.mean}{$-0.0035$}%
\rbnset{E1.K4.ci}{[$-0.0234$, $+0.0164$]}%
\rbnset{E1.K16.mean}{$-0.0082$}%
\rbnset{E1.K16.ci}{[$-0.0316$, $+0.0153$]}%
\rbnset{E11.K4.lo}{$+0.52$}%
\rbnset{E11.K4.hi}{$+1.14$}%
\rbnset{E11.K16.lo}{$+0.06$}%
\rbnset{E11.K16.hi}{$+0.09$}%
\rbnset{E4.WmP.mean}{$+0.010$}%
\rbnset{E4.WmP.ci}{[$-0.014$, $+0.034$]}%
\rbnset{E4.vsSoftmax.lo}{0.2}%
\rbnset{E4.vsSoftmax.hi}{0.3}%
\rbnset{E5.250.maxabs}{0.0023}%
\rbnset{E3a25.final3}{$+0.017$}%
\rbnset{E3a25.final}{$+0.0172$}%
\rbnset{E3a25.final.ci}{[$+0.0156$, $+0.0189$]}%
\rbnset{E3b.g.mean3}{$+0.067$}%
\rbnset{E3b.d.mean3}{$+0.088$}%
\rbnset{E2.50.closes}{97.3}%
\rbnset{E2.50.opens}{93.7}%
\rbnset{E2.50.freeze}{64.4}%
\rbnset{E2.125.closes}{65.6}%
\rbnset{E2.125.opens}{53.8}%
\rbnset{E2.125.freeze}{19.5}%
\rbnset{E2d.WP.mean}{$-0.074$}%
\rbnset{E2d.WP.ci}{[$-0.098$, $-0.049$]}%
\rbnset{E2d.PW.mean}{$+0.067$}%
\rbnset{E2d.PW.ci}{[$+0.054$, $+0.080$]}%
\rbnset{E2d.target.max}{0.015}%
\rbnset{E8.val.s7}{$+0.0011$}%
\rbnset{E8.val.s7.ci}{[$-0.0003$, $+0.0024$]}%
\rbnset{E8.val.s1337}{$+0.0003$}%
\rbnset{E8.val.s1337.ci}{[$-0.0013$, $+0.0020$]}%
\rbnset{E8.ss.wt}{$-0.008$}%
\rbnset{E8.ss.ptb}{$-0.016$}%
\rbnset{E8.ss.c4}{$+0.001$}%
\rbnset{E8.ss.lambada}{$-1.57$}%
\rbnset{E8.ss.lambada.ci}{[$-2.50$, $-0.68$]}%
\rbnset{E8.blimp.s7}{$-0.57$}%
\rbnset{E8.blimp.s7.ci}{[$-1.18$, $+0.01$]}%
\rbnset{E8.blimp.s1337}{$-1.26$}%
\rbnset{E8.blimp.s1337.ci}{[$-1.85$, $-0.66$]}%
\rbnset{E3a25.swap.lo}{0.008}%
\rbnset{E3a25.swap.hi}{0.009}%
\rbnset{E6.online.max}{$+0.0041$}%
\rbnset{E6.online.max.ci}{[$+0.0038$, $+0.0046$]}%
\rbnset{calgrad.exaq.mean}{$+0.0205$}%
\rbnset{calgrad.exaq.ci}{[$+0.0002$, $+0.0408$]}%
\rbnset{calgrad.exaq.250}{$+0.055$}%
\rbnset{I.runs}{98}%
\rbnset{I.gpuh}{454.1}%
\rbnset{I.evalh}{7.8}%
\rbnset{E1.250.vsW}{$-0.0347$}%
\rbnset{E1.250.vsW.ci}{[$-0.0361$, $-0.0333$]}%
\rbnset{E1.250.vsP}{$-0.0028$}%
\rbnset{E1.250.vsP.ci}{[$-0.0039$, $-0.0016$]}%
\rbnset{E1b.vsLERP}{$+0.0022$}%
\rbnset{E1b.vsLERP.ci}{[$+0.0007$, $+0.0037$]}%
\rbnset{E1b.vsW}{$-0.0358$}%
\rbnset{E1b.vsW.ci}{[$-0.0375$, $-0.0342$]}%
\rbnset{E1b.vsP}{$-0.0124$}%
\rbnset{E1b.vsP.ci}{[$-0.0137$, $-0.0110$]}%
\rbnset{E1.250.final}{$+0.0041$}%
\rbnset{E1.250.final.ci}{[$+0.0030$, $+0.0051$]}%
\rbnset{E1b.held.wt}{$+0.0145$}%
\rbnset{E1b.held.ptb}{$+0.0185$}%
\rbnset{E1b.held.c4}{$+0.0086$}%
\rbnset{E11.zs.lo}{0.03}%
\rbnset{E11.zs.hi}{0.25}%
\rbnset{E11.zs.full}{$10^{-8}$}%
\rbnset{E11.06.vsC2}{$+0.0000$}%
\rbnset{E11.250}{$+1.71$}%
\rbnset{E11.250.span}{2970}%
\rbnset{E11.K4.mean}{$+0.675$}%
\rbnset{E11.K16.mean}{$+0.076$}%
\rbnset{E2.50.valid.ww}{$+0.0072$}%
\rbnset{E2.50.valid.pp}{$-0.0002$}%
\rbnset{E2.50.span.ww}{55.8}%
\rbnset{E2.50.span.wp}{21.6}%
\rbnset{E2.50.span.freeze}{18.9}%
\rbnset{E2.50.span.pp}{21.9}%
\rbnset{E2.50.span.pw}{62.9}%
\rbnset{E2.50.freezeleft}{$+0.1150$}%
\rbnset{E2.125.valid.ww}{$+0.0040$}%
\rbnset{E2.125.valid.pp}{$+0.0001$}%
\rbnset{E2.125.span.ww}{54.4}%
\rbnset{E2.125.span.wp}{30.3}%
\rbnset{E2.125.span.freeze}{37.1}%
\rbnset{E2.125.span.pp}{22.1}%
\rbnset{E2.125.span.pw}{48.4}%
\rbnset{E2.125.freezeleft}{$+0.2573$}%
\rbnset{E2c.ww.peak}{8.36}%
\rbnset{E2c.ww.peakm}{525}%
\rbnset{E2c.orig.peak}{7.35}%
\rbnset{E2c.orig.peakm}{375}%
\rbnset{E2c.freeze.first}{8.50}%
\rbnset{E2c.freeze.firstm}{255}%
\rbnset{E2c.freeze.end}{4.53}%
\rbnset{E2c.wp.end}{3.72}%
\rbnset{E2c.wp.vsww}{$-1.45$}%
\rbnset{E2d.WPmPP.mean}{$-0.0056$}%
\rbnset{E2d.WPmPP.ci}{[$-0.0254$, $+0.0143$]}%
\rbnset{E2d.PWmWW.mean}{$-0.0013$}%
\rbnset{E2d.PWmWW.ci}{[$-0.0176$, $+0.0150$]}%
\rbnset{E4.hist.100}{$+0.068$}%
\rbnset{E4.hist.250}{$+0.316$}%
\rbnset{E4.WmP.250}{$+0.076$}%
\rbnset{E5.mean.lo}{$-0.011$}%
\rbnset{E5.mean.hi}{$+0.000$}%
\rbnset{E5n.mean}{$-0.0057$}%
\rbnset{E5n.ci}{[$-0.0331$, $+0.0218$]}%
\rbnset{E0c.bf16exp.levels}{1021}%
\rbnset{E0c.fp8e4m3exp.levels}{56}%
\rbnset{E0c.mxfp8.levels}{120}%
\rbnset{E0c.nvfp4.levels}{7}%
\rbnset{E0c.uint8.levels}{255}%
\rbnset{E3b.dg.mean}{$+0.0205$}%
\rbnset{E3b.dg.250}{$+0.0553$}%
\rbnset{E3b.d.250span}{8259}%
\rbnset{E3b.g.250span}{33.1}%
\rbnset{E3a.ii.mean}{$-0.0380$}%
\rbnset{E3a.ii.ci}{[$-0.0831$, $+0.0071$]}%
\rbnset{E3a.ii.250}{$+0.0003$}%
\rbnset{E3a.i.250}{$+0.0014$}%
\rbnset{E3a25.vsFWMP}{$-0.0020$}%
\rbnset{E3a25.blimp}{$-1.81$}%
\rbnset{E3a25.blimp.ci}{[$-2.40$, $-1.19$]}%
\rbnset{E3a25.at1000}{$+0.0125$}%
\rbnset{E3a25.at2000}{$+0.0167$}%
\rbnset{E0b.u8.25.long}{0.147}%
\rbnset{E0b.allzero.max}{1.8}%
\rbnset{E0b.allzero.maxB}{1.0}%
\rbnset{E0b.allzero.end}{0.2}%
\rbnset{E8a.vs1337}{$-0.0011$}%
\rbnset{E8a.vs1337.ci}{[$-0.0027$, $+0.0005$]}%
\rbnset{E8.ss.c4.ci}{[$-0.0025$, $+0.0038$]}%
\rbnset{E8c.held.wt}{$+0.0132$}%
\rbnset{E8p.held.wt}{$-0.0084$}%
\rbnset{E8.blimp.s7.3}{$-0.567$}%
\rbnset{E8.blimp.s7.3ci}{[$-1.179$, $+0.015$]}%
\rbnset{E8.blimp.s1337.3}{$-1.261$}%
\rbnset{E8.blimp.s1337.3ci}{[$-1.851$, $-0.664$]}%
\rbnset{E0d.twopass}{1.5}%
\rbnset{E6.maxabs}{0.0041}%
\rbnset{E6.max3090}{0.0003}%
\rbnset{E7.fwm}{$+0.0030$}%
\rbnset{E7.fwm.ci}{[$+0.0009$, $+0.0050$]}%
\rbnset{E7.mm}{$-0.0063$}%
\rbnset{I.machineA.runs}{4}%
\rbnset{I.machineA.gpuh}{167.3}%
\rbnset{I.machineB.runs}{94}%
\rbnset{I.machineB.gpuh}{286.7}%
\rbnset{S0.A250.diff}{$+0.232$}%
\rbnset{S0.A250.ci}{[$+0.224$, $+0.240$]}%
\rbnset{S0.B50.diff}{$+0.016$}%
\rbnset{S0.B50.ci}{[$+0.015$, $+0.017$]}%
\rbnset{S0.B125.diff}{$+0.233$}%
\rbnset{S0.B125.ci}{[$+0.227$, $+0.239$]}%
\rbnset{S0.nsteps}{38}%
\rbnset{S0.tokM}{5}%
\rbnset{S0.A250.gnratio}{46}%
\rbnset{S0.A250.dl1}{$+0.24$}%
\rbnset{S0.A250.dlN}{$+3.69$}%
\rbnset{S0.B125.dl1}{$+0.24$}%
\rbnset{S0.B125.dlN}{$+0.02$}%
\rbnset{S0.A250.h0}{32.6}%
\rbnset{S0.B125.h0}{11.4}%
\rbnset{S0.B50.h0}{4.8}%

\newcommand{\codeurl}{\href{https://github.com/Soren-Shangzhen/Pretraining-Transformers-with-Quantized-Softmax-in-Attention}{\mbox{https://}\allowbreak\mbox{github.com/Soren-Shangzhen/}\allowbreak\mbox{Pretraining-}\allowbreak\mbox{Transformers-}\allowbreak\mbox{with-}\allowbreak\mbox{Quantized-}\allowbreak\mbox{Softmax-}\allowbreak\mbox{in-}\allowbreak\mbox{Attention}}}
\makeatletter
\newcommand{\codefootnote}{\begingroup\renewcommand{\thefootnote}{}\long\def\@makefntext##1{\noindent ##1}\footnotetext{Code: \codeurl}\endgroup}
\makeatother

\title{Pretraining Transformers\\with Quantized Softmax in Attention\vspace{-15pt}}

\author{Shangzhen Zhu$^{1}$, Muyan Hu$^{1}$ \& Tomasz Kozlowski$^{1,2}$ \\
$^{1}$University of Illinois Urbana-Champaign \\
$^{2}$National Centre for Nuclear Research, Otwock, Poland \\
\texttt{\{szhu48,muyanhu2,txk\}@illinois.edu}}

\iclrfinalcopy
\begin{document}
\raggedbottom
\maketitle
\vspace{-15pt}
\codefootnote

\begin{abstract}
Low-precision Transformer systems increasingly quantize attention matrix multiplications, while softmax often remains
at higher precision. During pretraining, an approximate softmax changes the gradients that train the model as well as
its forward computation. We study this interaction with $K$-interval attention, which approximates the exponential
using $K{+}1$ grid values. We vary per-row grid calibration, interpolation versus hard rounding, and the placement of a
straight-through surrogate relative to normalization. We derive the corresponding backward rules, including
calibration derivatives, and compare these choices in pretraining experiments matched on model, data, and optimizer.
Detaching the row extrema leaves the forward computation unchanged but produces a delayed increase in validation
loss, under both min--max and fixed-window calibration. With hard rounding at $K{=}4$, min--max calibration and a pre-normalization surrogate incur a large loss gap;
changing either choice substantially reduces it. At 124M parameters and 2.5B training tokens, fixed-window calibration with interpolation ends \rbn{E1b.final3} nats
from softmax at $K{=}4$; with hard rounding it ends $+0.019$ nats at $K{=}4$ under a post-normalization surrogate and
$+0.004$ nats at $K{=}16$ under a pre-normalization surrogate.
\end{abstract}

\section{Introduction}
\label{sec:intro}

Low-precision Transformer training has moved the linear projections and, increasingly, the attention
matrix products to narrow formats, while the exponential and its row reduction are kept in FP16/FP32 by
design \citep{smoothquant2023,mxfp8_2025,fp8dpa2025,sagebwd2026,fullstackfp4_2026}. Whether a low-precision reconstruction of the exponential can be used
during pretraining is left open by several works
\citep{exaq2024,sageattention2025,mxfp8_2025,baps2026}. FlashAttention-4 \citep{zadouri2026fa4} addresses the
exponential bottleneck on Blackwell GPUs by emulating part of the exponentials in software, with a result that is
indistinguishable from the hardware one after rounding to BF16, so the attention it computes matches exact softmax at
that precision. The approximations we study are coarse enough to change the attention probabilities.

The question differs in kind from the inference-time one. At inference an approximate operator perturbs a
fixed function; during training it defines the learning rule, because the model sees the approximate
forward and is updated by whatever gradient the implementation attaches to it. For softmax,
detaching the row maximum is harmless: shift invariance makes its gradient cancel exactly, so subtracting
the maximum is treated as a numerical convenience rather than part of the function. The same habit carried
over to a quantized grid is not harmless, because there the row extrema also set the grid (\S\ref{sec:res-detach});
quantization-aware training frameworks also keep their observer statistics out of the gradient computation
\citep{pytorch_fakequant}.

We study a family simple enough that every forward and backward term is closed-form (Figure~\ref{fig:schematic},
Table~\ref{tab:labels}). Three design axes define an operator: \emph{calibration}, row-wise min--max calibration
(MinMax) over each row's own extrema or a fixed window of $\tau$ nats anchored at the row maximum with zero weight below
it (\FW); \emph{reconstruction} of the exponential on $K$ intervals from $K{+}1$ tabulated grid values, by
piecewise-linear interpolation (LERP) or by rounding each score to the nearest grid center (\CH); and \emph{surrogate
placement}, with the straight-through surrogate applied to the unnormalized attention weights before normalization
(Weight-STE) or to the normalized probabilities (Prob-STE). We
call the family $K$-interval attention.
This paper analyses pretraining with an approximate softmax; the forward operator and its backward rule are both
objects of study. Training has to be studied because trained weights
depend on the training operator (\S\ref{sec:res-downstream}), so the operator used at deployment must also be used in
training.

\begin{itemize}[leftmargin=1.2em,itemsep=1pt,topsep=2pt]
\item \textbf{The same forward can fail solely because the calibration backward is incomplete.} Detaching
  the row extrema leaves the forward unchanged but drops two Jacobian terms and violates the zero-sum
  identity that shift invariance implies. Training then tracks softmax for 25--30M tokens, diverges, and ends
  0.65--3.07 nats above the matched run with full calibration gradients (the backward that keeps the
  derivatives through the row extrema; replicated on two further seeds). Restoring the zero-sum identity alone
  does not repair it. The same failure occurs on a fixed-width grid, where only the row maximum is detached
  (\S\ref{sec:res-detach}; Figure~\ref{fig:detach} and Table~\ref{tab:calgrad}).
\item \textbf{For hard reconstruction, calibration policy and surrogate placement strongly affect training
  quality and stability.} MinMax$\times$Weight-STE at $K{=}4$ ends $+0.39$ nats above softmax at 250M
  tokens and $+0.89$ at 2.5B. A post-normalization surrogate or \FW each removes most of the deficit, and
  the two together add little more. At $K{=}4$ the effects of calibration, of surrogate placement and of their interaction keep their
  direction at 1B parameters and on all five seeds, and the surrogate effect shrinks as $K$ increases
  (\S\ref{sec:res-2x2}; Table~\ref{tab:cross} and Figure~\ref{fig:2x2}). Switching the surrogate at a saved checkpoint,
  from Weight-STE to Prob-STE or the reverse, moves the run toward the trajectory of the new rule (\S\ref{sec:res-geom}).
\item \textbf{Coarse deterministic reconstruction need not prevent near-baseline performance.} LERP at the tested $K\in\{4,16,32\}$ ends within $0.005$ nats of softmax at 2.5B tokens,
  \FW--LERP at $K{=}4$ at \rbn{E1b.final3}, and \FW--Weight at $K{=}16$ at $+0.004$. Downstream, every large-gap condition scores below softmax on every benchmark
  configuration, while conditions within 0.01 nats show small, task-dependent differences of either sign
  (\S\ref{sec:res-k}, \S\ref{sec:res-downstream}; Figures~\ref{fig:k} and~\ref{fig:heat}).
\end{itemize}

\section{Related work}
\label{sec:related}

\textbf{Low-precision Transformers and the softmax path.} Weight and activation quantization and FP8/FP4
training \citep{gptq2023,smoothquant2023,fp8lm2023,mxfp8_2025,fullstackfp4_2026} quantize linear layers,
GEMM operands and data formats; the attention matmuls are quantized during training by FP8 dot-product
attention, SageBwd and Full-Stack FP4 \citep{fp8dpa2025,sagebwd2026,fullstackfp4_2026}, while the
exponential and its normalization stay in high precision: NVIDIA's MXFP8 recipe leaves reducing their
precision ``to future work'', and the NVFP4 recipe keeps softmax and both attention matrix products at
their original precision \citep{nvfp4_2025}. The nearest inference-side operators at the softmax itself subtract the row maximum and read the exponential
from a small lookup table over a window below it. EXAQ \citep{exaq2024} sets the window width from calibration
statistics of the softmax input and clamps inputs below the window to its edge (2--3-bit lookup); IndexSoftmax
\citep{intattention2026} uses a fixed window, zeroes inputs beyond it, and quantizes table entries and
probabilities to UINT8. EXAQ and BAPS \citep{baps2026} state that training is untested, and IndexSoftmax is
designed as a training-free drop-in replacement. \FW (\S\ref{sec:grad}) uses the same construction as IndexSoftmax, a fixed window anchored at the row maximum with
zero weight below it; the two differ in window width, number of levels, rounding and table precision
(Appendix~\ref{app:related}). We train IndexSoftmax and EXAQ as baselines in \S\ref{sec:res-k}. Our scope is the reconstruction of the exponential inside the softmax, the
derivative of its data-dependent row calibration, and where a surrogate sits relative to normalization. The row sum and
the division of $K$-interval attention are computed in fp32; two baselines with reduced-precision normalization are in
Appendices~\ref{app:rb-formats} and~\ref{app:rb-index}.

\textbf{Approximate and non-exponential softmax.} Softermax, I-BERT and ConSmax approximate the exponential or the row
reduction with static or learned scales and fine-tune or train small models \citep{softermax2021,ibert2021,consmax2024}; a
base-2 classifier softmax trains because its Jacobian changes only by a constant \citep{zhang2022base2}, whereas random (i.i.d.)\
softmax error above $10^{-6}$ breaks training \citep{zhang2023base2hp}; the error of our operators is a deterministic
function of the scores. Non-exponential attention trains with a stabilizer
\citep{wortsman2023relu,shen2023relu,rela2021,ramapuram2024sigmoid,katharopoulos2020}; LERP at $K{=}1$, which replaces
the exponential by a single linear segment over the row range, is the member of our family closest to it
(Appendix~\ref{app:related}).

\textbf{Surrogates and calibration gradients.} The straight-through estimator was named and analysed by
\citet{bengio2013ste}; binarized networks \citep{bnn2016} needed clipped surrogates and explicit weight
clipping because a hard forward is insensitive to latent magnitude, and \citet{yin2019ste} show that a
surrogate must match the hard forward at its extremes for the coarse gradient to correlate with the true
one. Straight-through Gumbel-softmax \citep{jang2017gumbel,maddison2017concrete} is the probability-level
STE with only one place to sit, because the relaxation is the normalization; our reconstruct-then-normalize
operator creates the weight-level/probability-level choice. PACT, LSQ and TQT
\citep{pact2018,lsq2020,tqt2020} learn clipping or step-size parameters with a gradient, and ITA
\citep{ita2023} learns the clipping range of an integer attention softmax; our calibration quantities are per-row
statistics of the scores, and dropping their derivative changes $\partial L/\partial s$ itself
(Appendix~\ref{app:related}). We are not aware of prior work that back-propagates through a per-row calibration
statistic in attention pretraining.

\section{$K$-interval attention and its gradients}
\label{sec:grad}

\begin{figure}[b]
\centering
\includegraphics[width=\textwidth]{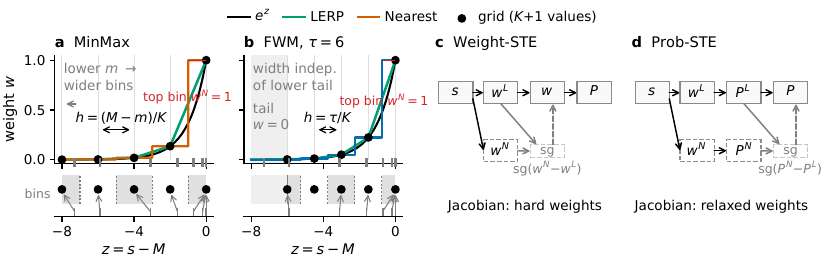}
\vspace{-8pt}
\caption{\textbf{The operator family} ($K{=}4$; three design axes). (a) MinMax calibrates the $K{+}1$
exponential grid values (dots) over the observed row range, $h=(M-m)/K$; a lower minimum widens every
bin at fixed $M$, $K$. (b) \FW anchors a fixed-width window at the row maximum, $h=\tau/K$, zero weight
below it (shaded tail). LERP (green) interpolates between adjacent grid values; \CH (step) rounds to the nearest grid
center; strips: \CH rounding bins. (c, d) Weight-STE applies the straight-through replacement to the unnormalized
weights, Prob-STE to the normalized probabilities; \emph{sg} (stop-gradient) receives both reconstructions (c) or
both normalized probabilities (d) and is added to the LERP branch, so the hard forward $P=P^N$ is the same.}
\label{fig:schematic}
\end{figure}

\begin{table}[b]
\vspace{-4pt}
\begin{minipage}[t]{0.535\textwidth}\vspace{0pt}\centering\scriptsize
\caption{Method identity (three design axes); edge cases in Appendix~\ref{app:ops}.}
\label{tab:labels}
\setlength{\tabcolsep}{2.5pt}\begin{tabular}{@{}llll@{}}
\toprule
Name & Calibration & Reconstr. & Backward \\
\midrule
softmax & --- & exp & autograd \\
LERP & MinMax & LERP & exact Jacobian \eqref{eq:full} \\
MinMax--Weight & MinMax & \CH & Weight-STE \eqref{eq:wste} \\
MinMax--Prob & MinMax & \CH & Prob-STE \eqref{eq:pste} \\
\FW--Weight & \FW, $\tau{=}6$ & \CH & Weight-STE \\
\FW--Prob & \FW, $\tau{=}6$ & \CH & Prob-STE \\
\FW--LERP & \FW, $\tau{=}6$ & LERP & exact Jacobian \\
\bottomrule
\end{tabular}
\end{minipage}\hfill
\begin{minipage}[t]{0.445\textwidth}\vspace{0pt}\centering\scriptsize
\caption{Notation.}
\label{tab:notation}
\setlength{\tabcolsep}{2.5pt}\begin{tabular}{@{}lp{0.80\linewidth}@{}}
\toprule
Symbol & Meaning \\
\midrule
$s_j$; $M,m$ & score of key $j$; row maximum, minimum \\
$z_j$; $q$ & $s_j-M\le0$; quantizer of $z$, $K{+}1$ levels \\
$h$; $\tau$ & interval width; \FW window width \\
$w_j$ & unnormalized attention weight (activation) \\
$w^L$; $w^N$ & interpolated; rounded $w$ \\
$W$; $P_j$ & $\sum_l w_l$; $w_j/W$ \\
$g_j$; $\gamma_j$ & $\partial L/\partial P_j$; $\partial L/\partial w_j$; $\langle g\rangle_P=\sum_iP_ig_i$ \\
$b_r$; $\Delta_j$ & grid points; slope of $w^L_j$ \\
$\sg(x)$ & stop-gradient ($x$ forward, 0 backward) \\
$\dNLL$ & NLL minus softmax (nats/token) \\
\bottomrule
\end{tabular}
\end{minipage}
\vspace{-2pt}
\end{table}

\textbf{Quantization view.} Every operator here quantizes the row-max-shifted score before
normalization: $z_j=s_j-M\le0$, $w_j=q(z_j)$, $P_j=w_j/\sum_lw_l$, where $w_j$ is the unnormalized attention weight (an
activation, not a parameter). $q$ has $K{+}1$ levels, an index of $\lceil\log_2(K{+}1)\rceil$ bits, and three design axes:
calibration (where the grid is placed), reconstruction (\CH returns one level; LERP interpolates between two) and, for
\CH, the surrogate that replaces $q'$. Under MinMax the grid is recalibrated for every row from the row maximum $M$ and
minimum $m$. Under \FW it is static in the shifted coordinate: the step $\tau/K$ and all $K{+}1$ values are constants, and
the only per-row quantity is $M$. All operators are computed in fp32.

\textbf{Calibration.} For one causal row of fp32 scores $s=(s_1,\dots,s_n)$, $S=QK^\top/\sqrt{d_h}$, and output
$O=PV$ (notation: Table~\ref{tab:notation}), \emph{MinMax} places the grid on the observed row range:
$h=(M-m)/K$, $b_r=(1-r/K)(m-M)$ for $r=0,\dots,K$, key $j$ in interval $r_j$ at local position
$t_j\in[0,1]$; all $K{+}1$ grid values $e^{b_r}$ move with the row extrema, and $h$ is set by the row minimum, so
the learned negative tail sets the resolution near the row maximum, where the high-weight keys are. \emph{Fixed Window to Max} (\FW) anchors a fixed-width window $[M-\tau,M]$ at
the row maximum: $h=\tau/K$, $b_r=-\tau+rh$, $\tilde z_j=\mathrm{clip}(z_j,-\tau,0)$, and, by explicit
implementation choice, $w_j=0$ for $z_j<-\tau$ (zero tail); it needs only $M$, whose key is always in the window, and keeps the gradient through $M$. $\tau=6$ was fixed before any \FW training run by a multi-stage inference-time
screen on frozen softmax-trained weights (Appendix~\ref{app:protocol}); no other $\tau$ was trained.
Edge cases (degenerate rows, ties, index switches, the window boundary) are specified in Appendix~\ref{app:ops};
the formulas below assume a nondegenerate row, a locally fixed interval index and unique extrema.

\textbf{Reconstruction.} LERP interpolates between adjacent exponential grid values,
$w^L_j=(1-t_j)e^{b_{r_j}}+t_je^{b_{r_j+1}}$, a piecewise-linear reconstruction of the exponential whose
unnormalized weights are continuous rather than grid values. \CH rounds each score to the nearest grid center in score coordinates (not the nearest
exponential value), $\ell_j=r_j+\ind[t_j\ge\tfrac12]$, $w^N_j=e^{b_{\ell_j}}$: grid-centered rounding
bins with boundaries halfway between grid points (Figure~\ref{fig:schematic}a,b), so distinct scores in one bin
get the same unnormalized weight. \CH is ``hard'' in the sense that it selects one grid value per key; attention
itself is not one-hot. The index $\ell_j$ is piecewise constant with zero derivative almost everywhere. Under
MinMax the grid values move with the extrema, so the hard map still has extremum-mediated derivatives where the
index is locally constant (Appendix~\ref{app:grad}); these carry no sensitivity to the index selection, which the
surrogate below supplies. Under \FW the hard unnormalized weights are locally constant, and the
zero tail is discontinuous at $z=-\tau$.

\textbf{The full calibration gradient and the zero-sum identity.} With $g_j=\partial L/\partial P_j$ the
normalization Jacobian gives $\gamma_j\equiv\partial L/\partial w_j=(g_j-\sum_iP_ig_i)/W$, and inside an
interval the LERP slope is the secant $\Delta_j=(e^{b_{r_j+1}}-e^{b_{r_j}})/h$. Because the MinMax grid depends
on $M$ and $m$, so does $w^L_j$; with $\rho_j=(r_j+t_j)/K$ and
$X_j=\tfrac1K[(1-t_j)r_je^{b_{r_j}}+t_j(r_j{+}1)e^{b_{r_j+1}}]$,
\begin{equation}
\frac{\partial w^L_j}{\partial M}=-w^L_j-\Delta_j\rho_j+X_j,\qquad
\frac{\partial w^L_j}{\partial m}=-\Delta_j(1-\rho_j)+w^L_j-X_j,
\label{eq:calgrad}
\end{equation}
\begin{equation}
\frac{\partial L}{\partial s_i}=\gamma_i\Delta_i
+\ind[i{=}\arg\max]\sum_j\gamma_j\frac{\partial w_j}{\partial M}
+\ind[i{=}\arg\min]\sum_j\gamma_j\frac{\partial w_j}{\partial m}.
\label{eq:full}
\end{equation}
The three per-key terms sum to zero, hence $\sum_i\partial L/\partial s_i=0$, the gradient identity that
shift invariance $F(s+c\mathbf1)=F(s)$ implies; under \FW the identity holds with
$\partial w_j/\partial M=-\Delta_j$ and no $m$ term. We call \eqref{eq:full} the \emph{full calibration
gradient}. For the differentiable LERP operator it is the exact Jacobian, calibration included; for the
straight-through rules below it is the set of calibration derivatives that the surrogate carries, not the
derivative of the hard forward. The \emph{detach} variant keeps only $\gamma_i\Delta_i$ and violates the identity
(\S\ref{sec:res-detach}). For a hard forward the index path's derivative is zero almost everywhere, so we adopt a
LERP surrogate, and the question becomes which calibration derivatives the surrogate carries.

\textbf{Two places to put the surrogate.} Weight-STE applies the straight-through replacement to the unnormalized
attention weights before sum normalization, Prob-STE to the normalized attention probabilities; $\sg(x)$ returns
$x$ in the forward pass and has zero derivative in the backward pass:
\begin{align}
\text{Weight-STE:}\ \ w = w^L+\sg(w^N-w^L),\ P=\tfrac{w}{\sum_l w_l},
&\qquad \gamma^{W}_j = \frac{g_j-\langle g\rangle_{P^N}}{W^N}, \label{eq:wste}\\
\text{Prob-STE:}\ \ P = P^L+\sg(P^N-P^L),\ P^L=\tfrac{w^L}{W^L},
&\qquad \gamma^{P}_j = \frac{g_j-\langle g\rangle_{P^L}}{W^L}. \label{eq:pste}
\end{align}
Both rules define the same mathematical hard forward $P=P^N$
(the two floating-point implementations differ by rounding, measured on frozen checkpoints in
Appendix~\ref{app:fwdaudits}). Both use the same LERP reconstruction derivatives $\Delta$, $\partial w/\partial M$,
$\partial w/\partial m$ of \eqref{eq:full}, because both stop the gradient on the difference branch. They differ in
the point at which the normalization Jacobian is evaluated, so the centering distribution in the numerator and
the normalization factor can both differ. Prob-STE's backward is the Jacobian of
$s\mapsto\mathrm{normalize}(w^L(s))$ contracted with the hard-forward upstream gradient. Weight-STE combines the
normalization Jacobian at the hard point ($W^N,P^N$) with the LERP reconstruction derivatives and generally
differs from that Jacobian. The two rules coincide when $w^N=w^L$. Their difference $\gamma^{W}-\gamma^{P}$ depends on the within-interval positions, on $g$ and on the
reconstructed unnormalized weights, not on $h$ alone, but a wider interval permits a larger mismatch between hard and
interpolated reconstructions. Under MinMax the width grows with the learned span; under \FW $h=\tau/K$. The surrogate
effect should therefore shrink as $K$ increases, faster under \FW than under MinMax, a qualitative expectation that
\S\ref{sec:res-2x2} tests.

\section{Experimental protocol}
\label{sec:protocol}

\begin{table}[t]
\centering\footnotesize
\caption{Suites A--D and the question each answers.}
\label{tab:suites}
\begin{tabular}{@{}llrrlrr@{}}
\toprule
Suite & Purpose & Model & Tokens & Corpus & Seeds & Runs \\
\midrule
A long-horizon & $K$, horizon, downstream & 124M & 2.5B & FineWebEdu-3B & 1 & 15 \\
B factorial & calibration $\times$ surrogate $\times K$ & 124M & 250M & FineWebEdu-3B & 1 & 10 \\
C scale probe & ordering at $8\times$ parameters & 1B & 100M & WikiText-103 & 1 & 12 \\
D seeds & $5{\times}5$ grid, seed spread & 124M & 100M & WikiText-103 & 5 & 130 \\
\bottomrule
\end{tabular}

\smallskip
\begin{minipage}{\linewidth}\scriptsize Every suite matches model, data and optimizer protocol (AdamW, cosine
decay, 1024-token sequences, 131,072 tokens per step, one learning-rate schedule for all operators) and varies the operator; suite D
also varies the seed. Training stacks: Tables~\ref{tab:dstack} and~\ref{tab:hardware}.\end{minipage}
\vspace{-6pt}
\end{table}

\textbf{Training.} GPT-2-style models (124M: 12 layers, $d{=}768$; 1B: 32 layers, $d{=}1536$). Everything outside
attention runs under bf16 autocast; the attention operator runs in fp32 with TF32 matmuls enabled in training
and disabled at evaluation. Per-suite hardware and software stacks: Appendix~\ref{app:protocol}.

\textbf{Suites} (Table~\ref{tab:suites}). Suite A trains for 2.5B tokens under its own cosine schedule, $10\times$
the budget of suite B, so it tests whether the gaps of the factorial persist at a much larger training budget; it is also
the only suite trained long enough for downstream evaluation to be meaningful. Suite B is a balanced $2{\times}2{\times}2$ design (calibration $\times$ surrogate $\times$
$K\in\{4,16\}$) that estimates the main effects and interactions (the comparison between the two surrogates carries a forward-rounding confound;
Appendix~\ref{app:fwdaudits}).
Suite C asks whether the ordering reappears at $8\times$ the parameters in an early regime (0.1
tokens/parameter) on another corpus (a probe, not a scale experiment). Suite D holds the full
$5{\times}5$ grid fixed and varies the seed, the only suite that measures seed-to-seed spread.

\textbf{Evaluation.} Every checkpoint is evaluated on the full validation split of its corpus (976 blocks
FineWebEdu, 243 WikiText-103; 1024-token blocks, batch 1, fp32 attention kernel, TF32 off) through the run's
own frozen training code, since a shared implementation or a larger batch perturbs single
\CH blocks by up to $10^{-2}$ (Appendix~\ref{app:protocol}). All comparisons are paired differences $\dNLL$ in nats per token (GPT-2 BPE), positive when the condition is
worse than softmax; $\delta$ nats is a perplexity ratio $e^{\delta}$ (0.01 nats $\approx$ 1\%, $+0.89$ nats
$\approx2.4\times$).

\textbf{Statistics.} A contrast (a paired difference between two conditions, or a difference of such differences) is
computed per block and averaged; single-seed suites report 95\%
circular moving-block bootstrap intervals over validation blocks (block 16, 4000 resamples), covering
evaluation-set sampling only. Suite D pairs each condition with softmax on the same seed and reports
the mean of five paired differences, their seed SD and a 95\% $t$ interval, covering seed-to-seed
variation. A single-seed contrast is \emph{separated} at a checkpoint when its CI lower bound exceeds $+0.01$ nats,
a pre-declared caution margin, not a significance or equivalence threshold. On five seeds the seed-to-seed spread of a
condition's $\dNLL$ exceeds its paired evaluation SE several-fold (\S\ref{sec:res-2x2}), so single-seed differences
below about 0.01 nats are not interpreted, and single-seed intervals never stand in for seed variation.
Trajectories, geometry and cross-condition associations are observations; the interventions of
\S\ref{sec:res-detach} (calibration quantities taken off the graph) and \S\ref{sec:res-geom} (surrogate switching and step
freezing from saved checkpoints) are the causal evidence.

\textbf{Geometry probe.} At every checkpoint we recompute each layer's fp32 scores on 8 validation blocks and record
the row span $M-m$, the native width $h$ and the total variation TV between native and softmax probabilities, a
measure of the operator's distortion (Appendix~\ref{app:protocol-eval}).

\section{Results}
\label{sec:results}

\subsection{Calibration-gradient intervention: same forward, delayed failure}
\label{sec:res-detach}

\begin{figure}[t]
\centering
\includegraphics[width=\textwidth]{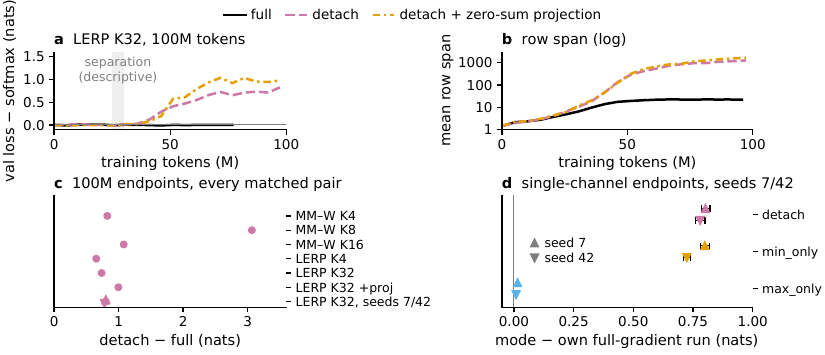}
\vspace{-12pt}
\caption{\textbf{Same forward, different calibration backward} (124M, 100M tokens, WikiText-103). (a) LERP $K{=}32$,
seed 1337: logged validation loss minus softmax under full (black), detach (purple dashed) and detach with the zero-sum
projection (orange dash-dot); grey band: observed separation. (b) Logged mean row span (log). (c) 100M
endpoints, detach minus full, every matched pair (MM--W: MinMax--Weight; +proj: with the zero-sum projection; CIs: Tables~\ref{tab:detach}, \ref{tab:detachrep}). (d) Single-channel ablation, LERP $K{=}32$: max\_only and min\_only keep only the maximum- or the minimum-dependent calibration gradient; each run minus its own same-seed full run, seeds 7
($\triangle$) and 42 ($\triangledown$), 95\% block-bootstrap CIs over validation blocks (evaluation sampling only; Table~\ref{tab:e1}).}
\label{fig:detach}
\end{figure}

Before suites A--D, the operator was trained with a backward that detached the row extrema: the forward code
path is unchanged apart from the \texttt{.detach()} calls, and the backward keeps $\gamma_i\Delta_i$ but drops the
extremum terms of \eqref{eq:full}. LERP $K{=}32$ is the cleanest case because its forward is differentiable. Its matched
run with full calibration gradients shows no detectable difference from softmax in this evaluation ($\dNLL=-0.0008$,
95\% CI $[-0.004,+0.002]$); the same forward with the detached backward ends $+0.737$ nats higher
(Figure~\ref{fig:detach}a). The failure is delayed: the run tracks softmax through 25M tokens and separates near 30M,
where its row span also leaves the full-gradient run and then grows by more than an order of magnitude
(Figure~\ref{fig:detach}b). The contrast replicates on seeds 7 and 42 (Table~\ref{tab:detachrep}), and \CH fails at every
$K$ against its matched run (Figure~\ref{fig:detach}c, Table~\ref{tab:detach}).

\textbf{Restoring zero-sum is not enough.} Projecting each detached row onto the zero-sum subspace removes the
common-mode component exactly, yet the projected run ends worse than plain detach (Figure~\ref{fig:detach}a,
Table~\ref{tab:detach}).

\textbf{Which extremum's gradient matters.} Retaining only the maximum-dependent calibration gradient recovers almost
the whole full--detach gap on both seeds; retaining only the minimum-dependent gradient recovers almost none of it
(Figure~\ref{fig:detach}d, Table~\ref{tab:e1}). The maximum-only backward is not strictly zero-sum yet ends 0.015 and
0.009 nats above full on the two seeds, while the projected run above satisfies zero-sum and fails, so zero-sum alone does not
explain the outcomes. Fixed-upstream decompositions at 20M and 40M agree (Table~\ref{tab:graddecomp}).

\begin{table}[t]
\centering\scriptsize
\caption{The calibration gradient across calibration types (124M parameters; $\dNLL$ of the run with the listed quantity
off the graph minus the run with it on the graph, nats).}
\label{tab:calgrad}
\setlength{\tabcolsep}{2pt}\begin{tabular}{@{}lllllp{2.35cm}@{}}
\toprule
Calibration & Off the graph & Operator & $\dNLL$, 100M & $\dNLL$, 250M & Span, off vs on (100M) \\
\midrule
per-row MinMax & $M$, $m$ & LERP $K{=}32$ & $+0.74$, $+0.80$, $+0.78$$^{a}$ & --- & 861--1193 vs 21--22 \\
per-row MinMax & $M$ ($m$ kept)$^{b}$ & LERP $K{=}32$ & $+0.80$, $+0.72$ & --- & 1489--1739 vs 21 \\
static (\FW) & $M$ & \FW--LERP $K{=}4$ & $+0.52$, $+0.83$ & $+1.71$ & 660--785 vs 28 \\
static (\FW) & $M$ & \FW--LERP $K{=}16$ & $+0.09$, $+0.06$ & --- & 56--57 vs 25 \\
static (\FW) & $M$ & \FW--Weight $K{=}4$ & $+1.08$, $+1.14$ & --- & 762--810 vs 19 \\
tensor-level (EXAQ) & $\sigma$, minimum & 3-bit, Prob-STE & $+0.0205$ [$+0.0002$, $+0.0408$]$^{c}$ & $+0.055$ & 114--118 vs 25--26 \\
per-head EMA & width (buffer) & \CH $K{=}4$ & $+0.216$ (W), $+0.206$ (P)$^{d}$ & $+0.307$, $+0.232$ & 51--54 (W), 677--800 (P) vs 25--26$^{e}$ \\
\bottomrule
\end{tabular}

\smallskip
\begin{minipage}{\linewidth}\scriptsize 100M: one value per seed; 250M: single runs. Rows 1--2: Appendix~\ref{app:detach};
other rows: Appendix~\ref{app:rb}. EMA: exponential moving average. $^{a}$Seed 1337 (Table~\ref{tab:detach}) and seeds 7, 42 (Table~\ref{tab:detachrep}).
$^{b}$The single-channel run that keeps only the gradient through $m$ (Figure~\ref{fig:detach}d, Table~\ref{tab:e1}).
$^{c}$Three-seed mean with 95\% $t$ interval. $^{d}$Difference from softmax under Weight-STE (W) and Prob-STE (P); the EMA width has no on-graph counterpart.
$^{e}$Softmax.\end{minipage}
\vspace{-8pt}
\end{table}

\textbf{Calibration beyond per-row MinMax} (Table~\ref{tab:calgrad}). Under \FW the step is constant, so
detaching the row maximum removes only the arg-max term of Eq.~\eqref{eq:full}, the derivative through the grid origin. Training still fails, by \rbn{E11.K4.lo} to \rbn{E11.K4.hi}
nats at $K{=}4$ and \rbn{E11.K16.lo} to \rbn{E11.K16.hi} at $K{=}16$. EXAQ with its tensor-level statistics detached,
and a per-head window whose width is an exponential moving average (EMA) of the row span, held in a buffer, show the
same direction (Appendices~\ref{app:rb-fwmlerp}, \ref{app:rb-ema},
\ref{app:rb-exaq}).

\subsection{Calibration $\times$ surrogate interaction and seed robustness}
\label{sec:res-2x2}

\begin{figure}[b]
\centering
\includegraphics[width=\textwidth]{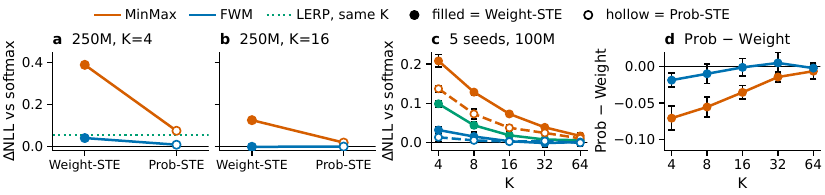}
\vspace{-10pt}
\caption{\textbf{Calibration $\times$ surrogate interaction for \CH.} (a, b) 124M @ 250M at
$K{=}4$ and $K{=}16$: final $\dNLL$ vs softmax by surrogate placement, one line per calibration, 95\%
block-bootstrap CIs (evaluation sampling only); dotted: LERP reference, in (a) only. (c) Five-seed mean $\dNLL$ vs $K$, all five families (suite D), 95\% $t$ intervals over seeds. (d) The surrogate effect Prob $-$ Weight against $K$ under each calibration,
paired by seed, 95\% $t$ intervals. Filled markers/solid lines:
Weight-STE; hollow/dashed: Prob-STE.}
\label{fig:2x2}
\end{figure}

Holding \CH reconstruction and $K{=}4$ fixed, calibration and surrogate placement interact strongly
(Table~\ref{tab:cross}, Figure~\ref{fig:2x2}a). MinMax--Weight ends $+0.39$ nats above softmax at 250M tokens and
$+0.89$ at 2.5B. Changing either factor, moving the surrogate to the probability level or replacing MinMax with \FW,
removes most of the deficit; changing both adds little more, so the interaction is large and positive. The calibration
effect, the surrogate effect and their interaction keep their sign at 1B parameters and on every seed of suite D.

The probability-level surrogate was tested under three calibrations. It removes most of the deficit under
per-row MinMax and a smaller amount under \FW. Under a per-head EMA window at $K{=}4$ its benefit is small
(\rbn{E4.WmP.mean} nats \rbn{E4.WmP.ci} at 100M tokens), and both Weight-STE and Prob-STE end
\rbn{E4.vsSoftmax.lo}--\rbn{E4.vsSoftmax.hi} nats above softmax (Appendix~\ref{app:rb-ema}).

A single-seed control that keeps \FW's fixed step but extends the grid into the tail recovers most of the
MinMax$\to$\FW gap: the improvement comes with the fixed step (Appendix~\ref{app:taildecomp}). A
strict-forward retraining at $K{=}4$ (seed 7, bit-exact hard forward) keeps the direction of both surrogate contrasts and
of the interaction, and the size of the MinMax contrast (Table~\ref{tab:strict}, Appendix~\ref{app:strict}).

\begin{table}[t]
\centering\scriptsize
\caption{Matched factorial contrasts at $K{=}4$ intervals across suites A--D (nats).}
\label{tab:cross}
\adjustbox{max width=\textwidth}{\begin{tabular}{lrrrl}
\toprule
Contrast, $K{=}4$ (nats) & B: 124M@250M & A: 124M@2.5B & C: 1B@100M & D: 5 seeds, mean [95\% $t$-CI] \\
\midrule
Prob $-$ Weight under MinMax, K4 & -0.316 & -0.829 & -0.074 & -0.071 [-0.087, -0.054] \\
Prob $-$ Weight under FWM, K4 & -0.032 & -0.023 & -0.022 & -0.019 [-0.028, -0.009] \\
FWM $-$ MinMax under Weight-STE, K4 & -0.350 & -0.848 & -0.159 & -0.176 [-0.193, -0.158] \\
FWM $-$ MinMax under Prob-STE, K4 & -0.067 & -0.042 & -0.108 & -0.124 [-0.138, -0.109] \\
Interaction, K4 & +0.284 & +0.805 & +0.051 & +0.052 [0.030, 0.075] \\
\bottomrule
\end{tabular}}

\smallskip
\begin{minipage}{\linewidth}\scriptsize Final checkpoints; corpora as in Table~\ref{tab:suites}. Single-seed CI
half-widths $\le0.0091$; suite D is paired by seed. Interaction $=(\mathrm{FWM}_P-\mathrm{FWM}_W)-(\mathrm{MM}_P-\mathrm{MM}_W)$, with MM for MinMax and subscripts $P$, $W$ for Prob-STE and Weight-STE.
Intervals and condition-vs-softmax values: Table~\ref{tab:crossfull}.\end{minipage}
\vspace{-8pt}
\end{table}

\textbf{The surrogate effect shrinks as $K$ increases.} In the 250M factorial the surrogate effect under \FW is about $30\times$
smaller at $K{=}16$ than at $K{=}4$ and only just detectable, whereas under MinMax it stays large (Figure~\ref{fig:2x2}b,
Table~\ref{tab:crossfull}); five seeds show the same decay (Figure~\ref{fig:2x2}d). MinMax--Weight $K{=}4$ is the only
condition with an optimization abnormality (45\% of steps clipped at 250M); MinMax--Weight $K{=}16$ shows none yet ends
$+0.124$ nats behind softmax.

\textbf{Seed robustness.} Across five seeds the ordering of conditions is reproducible (Kendall's $W$
0.91; Figure~\ref{fig:2x2}c). Averaged over $K$, \FW reduces NLL by 0.067 nats relative to MinMax and Prob-STE by
0.021 relative to Weight-STE, with a $+0.031$ interaction: the two changes substitute for each other. All three effects
shrink as $K$ increases, and the \FW cells at $K\ge16$ lie within seed noise of softmax. In these historical runs the training
stack was associated with the operator family (Appendix~\ref{app:protocol}).

\subsection{$K$ scaling and training horizon}
\label{sec:res-k}

\begin{figure}[b]
\centering
\includegraphics[width=\textwidth]{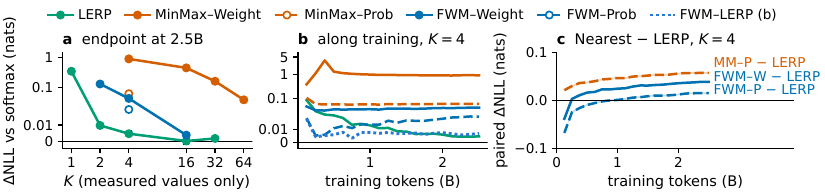}
\vspace{-10pt}
\caption{\textbf{124M @ 2.5B.} (a) Final $\dNLL$ vs softmax by $K$ and family at the measured $K$ only
(whiskers: 95\% block-bootstrap CI, evaluation sampling only). (b) Gap along training for the five $K{=}4$
conditions, and \FW--LERP $K{=}4$ (dotted; Appendix~\ref{app:rb-fwmlerp}). (c) Nearest minus LERP at matched $K{=}4$ (MM: MinMax; W, P: Weight-STE, Prob-STE); the 95\% CI (evaluation sampling only) is narrower than the
line width.
(a, b): symlog vertical scale, linear within $\pm0.02$ nats; (c): linear.}
\label{fig:k}
\end{figure}

\textbf{MinMax--LERP improves with training.} MinMax--LERP $K{=}4$ goes from $+0.086$ at 125M to $+0.005$
at 2.5B (Figure~\ref{fig:k}b); $K{=}16$ and 32 stay close to softmax from 250M on. The slow start comes from MinMax
calibration: \FW--LERP is already near softmax early in training. \FW--LERP $K{=}4$, which needs no straight-through
estimator, is at \rbn{E1b.at125} at 125M and \rbn{E1b.at250} at 250M and ends at \rbn{E1b.final} \rbn{E1b.final.ci}. Its
100M three-seed mean is \rbn{E1.K4.mean} \rbn{E1.K4.ci} (Appendix~\ref{app:rb-fwmlerp}).

\textbf{Reduced-precision and published baselines.} With $e^z$ rounded to BF16, FP8-E4M3, an
MXFP8-style or NVFP4-style block format, or UINT8 (the exponential's own derivative as backward), and with a full bf16
softmax chain, the three-seed mean at 100M tokens is within noise of softmax and the 250M endpoints are within
\rbn{E5.250.maxabs} nats (Appendix~\ref{app:rb-formats}). These rounding rules, applied in fp32, have a gradient that
does not involve the grid, so our conclusions concern calibration and the backward rule rather than bit width.
Trained with a Prob-STE backward of our choice, IndexSoftmax (32 levels, fixed window) is within noise of softmax at 100M and 250M tokens and ends \rbn{E3a25.final3} at 2.5B with its UINT8
output normalization; EXAQ at 3 bits ends \rbn{E3b.g.mean3} to \rbn{E3b.d.mean3} at 100M
(Appendices~\ref{app:rb-exaq}, \ref{app:rb-index}).

\textbf{Larger $K$ reduces the \CH deficit, with diminishing returns.} At 2.5B MinMax--Weight falls from $+0.89$ at $K{=}4$
to $+0.04$ at $K{=}64$ (Table~\ref{tab:endpoints}); at 1B parameters (100M tokens) LERP, MinMax--Weight and \FW--Weight order the same way at
every $K$; on five
seeds the slope against $\log_2K$ is negative for every family, with the high-$K$ \FW cells within seed noise of softmax
(Table~\ref{tab:ktrend}).

\textbf{The ordering of \CH and MinMax--LERP depends on the training budget.} \FW--Weight $K{=}4$
minus MinMax--LERP $K{=}4$ is negative at 125M and positive at 2.5B, and so is \FW--Prob $K{=}4$ minus MinMax--LERP $K{=}4$
(Figure~\ref{fig:k}c); both are
negative at 250M training tokens and at 1B parameters (100M tokens): MinMax--LERP's gap keeps closing while the \CH gaps plateau, so the relative NLL of MinMax--LERP and \CH
changes with the training budget. Under \FW the ordering does not change with the budget: \FW--LERP $K{=}4$ is at or
below both \FW--\CH variants at 250M and at 2.5B (Appendix~\ref{app:rb-fwmlerp}).

\subsection{Learned geometry: observations, interventions and a hypothesis}
\label{sec:res-geom}

Among the conditions of suites A--D with full calibration gradients, MinMax--Weight is the only one whose score span keeps
growing (Figure~\ref{fig:geom} in Appendix~\ref{app:ext}). At 2.5B its $K{=}4$ row span is about seven times softmax's,
its native width reaches $h{=}53$ nats and the total variation between native and softmax probabilities is 0.86: the
model sharpens scores that the operator cannot resolve. Moving the surrogate to the
probability level with the same hard forward keeps the span at softmax's level (MinMax--Prob $K{=}4$), as does \FW, which
changes calibration, tail and backward together; the span growth is specific to the surrogate--calibration combination.

Three observations limit the interpretation. MinMax--Weight $K{=}64$'s span more than doubles over training while its
gap stays between $+0.02$ and $+0.04$ nats and $h$ stays moderate (Figure~\ref{fig:geom}c). At 250M the
MinMax--Weight/Prob gap is already open at 50M tokens while the two geometries nearly coincide
(Figure~\ref{fig:onset}). MinMax--Weight $K{=}16$ has a gap without any gradient abnormality. Across the five-seed grid
the condition-mean $\dNLL$ has Spearman correlation $+0.95$ with native width and $-0.31$ with score span; among these
designed conditions the width association is the stronger one (Figure~\ref{fig:widthgap}).

\textbf{Interventions from saved checkpoints.} We branched the 250M-token MinMax $K{=}4$ runs at 50M and
125M tokens and continued under either surrogate (Appendix~\ref{app:rb-interv}). Switching Weight-STE to Prob-STE at
50M removes \rbn{E2.50.closes}\% of the endpoint gap between the unswitched branches and brings the span to the Prob-STE
level; the reverse switch reproduces \rbn{E2.50.opens}\% of it. From 125M the two switches remove \rbn{E2.125.closes}\%
and reproduce \rbn{E2.125.opens}\%. On three seeds at 100M tokens the effect has the same sign on every seed (\rbn{E2d.WP.mean} nats
\rbn{E2d.WP.ci}; \rbn{E2d.PW.mean} \rbn{E2d.PW.ci}), and each switched branch ends within \rbn{E2d.target.max} nats of the
branch that used its new rule throughout. The surrogate rule is therefore a cause of the MinMax--Weight failure.

\textbf{The width as mediator and a feedback hypothesis.} Freezing the step (at each layer's mean native width at the
branch point) under Weight-STE removes \rbn{E2.50.freeze}\% of the gap from 50M and \rbn{E2.125.freeze}\% from 125M. The
frozen-step operator also changes the forward's tail, so these shares reflect the width together with the tail. We
hypothesize a range--resolution feedback under MinMax--Weight: a wider interval degrades the surrogate gradient, which
lets the score span grow and widens the interval further. The interventions support two of its links, that the
surrogate rule drives the span and the gap and that the interval width carries part of the gap; QK-normalization and a
learning-rate sweep would test the loop as a whole.

\subsection{Downstream transfer: graded, not binary}
\label{sec:res-downstream}

Four predictions, decision rules and three pre-specified groups (large-gap, intermediate-gap and near-baseline, by 2.5B $\dNLL$)
were fixed before any downstream run, and every condition is evaluated with its native forward
(Appendix~\ref{app:full}). On WikiText-103, PTB and C4 the ordering is preserved (Spearman $\ge0.95$), and the
near-baseline group shows small held-out differences of both signs: 10 of its 15 paired intervals exclude zero and 7 lie
within $\pm0.01$ nats (Table~\ref{tab:nearci}). Two softmax seeds differ by \rbn{E8.ss.wt} and \rbn{E8.ss.ptb} nats
on WikiText-103 and PTB (Appendix~\ref{app:rb-seed}), so held-out differences of this size are not attributable to the
operator. Substituting exact softmax at evaluation widens the gap in 37 of 42
cells; for LERP $K{=}2$ on the WikiText-103 test set it goes from $+0.005$ to $+1.814$ nats (Table~\ref{tab:subst}): the trained weights
depend on the training operator. The IndexSoftmax model trained for 2.5B tokens loses \rbn{E3a25.swap.lo}--\rbn{E3a25.swap.hi}
nats when exact softmax or the float-normalized lookup replaces its operator at evaluation (Appendix~\ref{app:rb-index}). The substitution was run on the corpora and probes, not on the benchmark battery.

\begin{figure}[t]
\centering
\includegraphics[width=\textwidth]{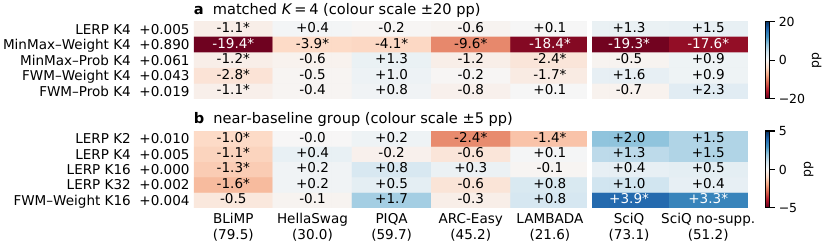}
\vspace{-8pt}
\caption{\textbf{Downstream accuracy differences from softmax} (pp, native forward), 2.5B endpoints; row labels:
training-domain $\dNLL$; column labels: softmax's accuracy. (a) The matched $K{=}4$ conditions ($\pm20$ pp scale); (b)
the near-baseline group ($\pm5$ pp). BLiMP: macro-accuracy; HellaSwag/PIQA/ARC-Easy: length-normalized accuracy; LAMBADA and
SciQ: accuracy. Star: BH $q<0.05$ within the column's 14-comparison family; all 14 conditions in Appendix~\ref{app:full}.}
\label{fig:heat}
\end{figure}

On six benchmarks evaluated in seven configurations (Figure~\ref{fig:heat}, Table~\ref{tab:stagec}), large corpus-NLL
degradation transfers: all 35 large-gap comparisons (five conditions $\times$ seven configurations) have negative point
estimates and 27 reach $q<0.05$ after Benjamini--Hochberg (BH) correction, and MinMax--Weight $K{=}4$ loses 3.9--19.4
percentage points (pp) of accuracy, less at larger $K$. Of the near-baseline group's 35 comparisons, 8 reach $q<0.05$, six negative (LERP $K{=}2/4/16/32$ on BLiMP; LERP $K{=}2$ on ARC-Easy and
LAMBADA) and two positive (\FW--Weight $K{=}16$ on both SciQ configurations; Table~\ref{tab:stagec}). On a second seed of
softmax and LERP $K{=}16$, the validation gap reproduces (\rbn{E8.val.s7} against \rbn{E8.val.s1337}) and the BLiMP
difference keeps its sign at about half the size (\rbn{E8.blimp.s7} pp \rbn{E8.blimp.s7.ci} against \rbn{E8.blimp.s1337}
pp \rbn{E8.blimp.s1337.ci}), while the two softmax runs differ by \rbn{E8.ss.lambada} pp on LAMBADA
(Appendix~\ref{app:rb-seed}). \FW--Prob $K{=}4$
has the smaller training-domain gap than \FW--Weight $K{=}4$ but the larger copying gap at $L{=}128$ (single seed,
synthetic probe; Table~\ref{tab:probec}).

\section{Discussion and limitations}
\label{sec:discussion}

The same hard forward yields different models under different backward rules, and an incomplete Jacobian fails
where the complete one does not. \S\ref{sec:res-geom} shows by intervention that the surrogate rule causes the
MinMax--Weight failure and that the interval width carries part of the effect.

A reconstruction rule whose backward is the exact derivative of its forward is not thereby a better training rule. A
bin-mean reconstruction, which sets each bin's weight to the mean of $\exp$ over its members so that bin mass is
conserved and no surrogate is needed, ends $+0.14$ nats above softmax at \FW $K{=}4$, and replacing only the backward
of strict \FW--Prob by this rule reproduces the whole gap (Appendix~\ref{app:binmean}). Its score gradient replaces each
key's upstream gradient by the bin average and drops the within-bin contrast that the LERP surrogate retains.

\textbf{Deployment and cost.} A model trained with an approximate operator must use the same operator at
inference (\S\ref{sec:res-downstream}). On FlashAttention-4's roofline, removing the exponential
leaves the GPU bound unchanged because the matrix multiply is co-limiting, and a single-pass lookup kernel would compute a
different function from the trained operator (up to \rbn{E6.online.max} nats; Appendices~\ref{app:rb-cost},
\ref{app:rb-evalset}). A coarse exponential is cheaper on integer and lookup hardware: \FW needs no exponential at run
time, and IntAttention, an integer attention pipeline whose IndexSoftmax uses the same fixed-window construction, runs up
to $2.0\times$ faster than an INT8 attention pipeline that keeps a floating-point softmax and up to $3.7\times$ faster than
FP16 attention on an \mbox{RK3588S2} (Armv8) CPU \citep{intattention2026}. The GPU bound is our estimate from
FlashAttention-4's published inputs; the Armv8 figures are their measurements.

\textbf{Limitations.} \emph{Seeds and budget.} Suites A--C use one seed, as do the 250M- and 2.5B-token runs of
Appendix~\ref{app:rb}, except softmax and LERP $K{=}16$, which have two seeds at 2.5B; suite D's seed spread applies to its own
setting, and its $t$ intervals assume near-normal seed effects. The tail-policy comparison is single-seed (historical
STE arithmetic, not equal cost), the channel ablation two seeds of one setting, and the strict-forward retraining and the bin-mean
control of Appendix~\ref{app:binmean} one seed each at 100M tokens; the reduced-precision baselines have three seeds (Appendix~\ref{app:rb-formats}). Each suite uses a
single learning-rate schedule.

\emph{Forward implementation and training stack.} The two surrogates share the hard forward mathematically but not
bitwise in the historical implementations: over 12 audited 124M $K{=}4$ endpoints the mean validation NLL differs by at
most $3.4{\times}10^{-4}$ nats (Table~\ref{tab:strictaudit}), and a bit-exact retraining reproduces the direction of both
contrasts at one seed and horizon (Appendix~\ref{app:strict}). In suite D the training stack was partly associated with
the operator family (Appendix~\ref{app:protocol}).

\emph{Window width, scale and horizon.} $\tau=6$ came from an inference-time screen on frozen weights whose final
stage compared $\tau\in\{5,6,7\}$ on the WikiText-103 validation split later used for suites C and D; training used this
value throughout. The 1B suite is an early-regime probe on another corpus, with Prob-STE only at $K{=}4$, and the runs
reach at most 1B parameters and 2.5B tokens, due to the limit of our compute budget. Detectability and margin status remain separate labels (\S\ref{sec:protocol}).

\section{Conclusion}
\label{sec:conclusion}

Pretraining with quantized softmax depends on both the forward approximation and its backward rule. Detaching the row
extrema leaves the forward computation unchanged but causes delayed training failure, on both a per-row calibrated grid and a fixed-width grid; projecting the resulting
gradients onto the zero-sum subspace does not recover the result with full calibration gradients. Under hard rounding
at $K{=}4$, MinMax calibration combined with Weight-STE incurs a large loss gap. Replacing MinMax calibration with \FW
or moving the surrogate to the probability level reduces this gap, with effects consistent in direction across the
tested horizons, scales and seeds. Selected hard and interpolated configurations approach softmax validation loss. The
remaining performance differences depend on the operator, $K$, training horizon and evaluation metric. Approximate softmax operators for pretraining should therefore be
evaluated through their calibration, reconstruction and backward rules together.

\label{m:end-of-main-text}

\subsection*{AI use statement}
Generative AI assistants (code-capable large language models) were used throughout this project, and we
describe their role by activity. \emph{Numerical results.} Every training run,
evaluation, statistic, figure and table comes from the training and analysis code and the source data
described in Appendix~\ref{app:assets}; no number in the paper was generated by an AI model. \emph{Reasoning
and design.} AI assistants were used in discussions of research directions and of what the results
support, in designing analyses (contrast definitions, bootstrap and pairing schemes, pre-declarations for
the downstream evaluation), in drafting and revising analysis, evaluation, figure- and table-generation
scripts and shell commands, in diagnosing experiments (including the gradient-consistency diagnostics
that identified the detached-extremum defect), in choosing follow-up experiments, and in drafting,
editing and restructuring the manuscript, its related-work survey and its responses to reviewers.
\emph{Mathematical claims.} AI assistants were also used to derive and check the mathematical claims of
\S\ref{sec:grad} and Appendix~\ref{app:grad}: the closed-form calibration derivatives, the zero-sum identity, the
Weight-STE and Prob-STE backward rules and the derivative of the hard map. Every derivation was checked line by line by
the authors, and each closed form is verified numerically against automatic differentiation of the frozen training code
(Appendix~\ref{app:grad}).
\emph{Author verification.} The authors set the research question, approved the experiments,
executed or supervised the training and evaluation jobs, and reviewed the manuscript and its AI-assisted
edits before submission. The degenerate-row behaviour is verified against the frozen code by a script that ships with the
analysis code, and every number in the paper is regenerated from the stored result files by the figure and table
scripts. Literature comparisons rest on the cited papers'
stated methods and claims; we do not claim that every cited paper was read in full by every author. We take responsibility for the final
content of this work, including text, claims and artifacts produced with the aid of generative AI.

\subsection*{Ethics statement}
This work studies numerical attention operators on public text corpora (FineWebEdu, WikiText-103) and public
benchmarks; it involves no human subjects, no personal data and no deployment. Its potential downstream use is
more efficient training hardware, which carries the general dual-use considerations of any efficiency advance
and no specific additional risk that we can identify.

\subsection*{Reproducibility statement}
Every run is identified by a frozen manifest carrying operator axes, seed, checkpoint paths and the SHA-256 of
the training code root it is evaluated with (Appendix~\ref{app:assets}); all checkpoints are evaluated through
their own frozen code at batch size 1 (\S\ref{sec:protocol}, Appendix~\ref{app:protocol}). The closed-form
gradients of \S\ref{sec:grad} are derived in Appendix~\ref{app:grad} and verified against automatic
differentiation by a script that ships with the analysis code. Figures and tables are generated from the
per-checkpoint result files by two scripts; the $\tau$ selection record, the frozen-checkpoint forward
discrepancy measurement and the detach intervention are documented in Appendices~\ref{app:protocol}
and~\ref{app:detach}. Code, run manifests and per-block evaluation results for all experiments, including those of
Appendices~\ref{app:rb} and~\ref{app:binmean}, are available at \codeurl.

\bibliography{references}
\bibliographystyle{iclr2027_conference}

\clearpage
\flushbottom
\appendix
\renewcommand{\thefigure}{S\arabic{figure}}\renewcommand{\theHfigure}{S\arabic{figure}}
\renewcommand{\thetable}{S\arabic{table}}\renewcommand{\theHtable}{S\arabic{table}}
\setcounter{figure}{0}\setcounter{table}{0}
\section{Operator definitions and edge cases}
\label{app:ops}

All operators act on one causal row of fp32 scores $s=(s_1,\dots,s_n)$, $S=QK^\top/\sqrt{d_h}$, computed
inside a disabled-autocast region (fp32 tensors; TF32 tensor-core matrix products are enabled during training and
disabled at evaluation); everything outside the attention kernel runs under bf16 autocast.
Invalid (future) positions get $w=0$. With $M=\max_j s_j$, $m=\min_j s_j$, $z_j=s_j-M$:

\textbf{Softmax.} $w^E_j=e^{z_j}$, $P^E=\mathrm{softmax}(s)$; explicit fp32 stable softmax.

\textbf{MinMax grid.} $z_{\min}=m-M$, $h=(M-m)/K$, $b_r=z_{\min}+rh$ for $r=0,\dots,K$;
$r_j=\mathrm{clip}(\lfloor(z_j-z_{\min})/h\rfloor,0,K{-}1)$, $t_j=\mathrm{clip}((z_j-b_{r_j})/h,0,1)$.

\textbf{\FW} (Fixed Window to Max). $h=\tau/K$,
$b_r=-\tau+rh$, $\tilde z_j=\mathrm{clip}(z_j,-\tau,0)$; $w_j=0$ if $z_j<-\tau$; $\tau=6$. The row minimum
is computed only as a diagnostic.

\textbf{LERP.} $w^L_j=(1-t_j)e^{b_{r_j}}+t_je^{b_{r_j+1}}$: a continuous, interpolated weight, not one of
the $K{+}1$ grid values. For $K{=}1$ a single interval interpolates between $e^{z_{\min}}$ and 1.

\textbf{\CH.} $\ell_j=r_j+\ind[t_j\ge\tfrac12]$, $w^N_j=e^{b_{\ell_j}}$: the nearest grid
\emph{center} in $z$, not the nearest exponential value (nearest-center rounding). Under \FW the hard operator can
additionally produce a zero tail.

\textbf{Normalization.} $P=w/\sum_lw_l$ after reconstruction, so $\sum_jP_j=1$ in exact arithmetic for every operator.

\paragraph{Edge cases and conventions (as implemented).} The conventions below are those of the attention code used by all runs; in a numerical check all values are
finite and row sums equal 1 within $10^{-6}$. Three conditions are kept apart: the nondegeneracy condition $h>0$ under which the paper's formulas are
stated; the implementation's regular branch ($M-m>\varepsilon$); and its degenerate branch ($M-m\le\varepsilon$,
$\varepsilon=10^{-12}$). On the MinMax path the extrema are taken over the causal-valid keys and a safe width $h_{\mathrm{safe}}=1$
(degenerate) or $h$ (regular) is formed before any division, so no $0/0$ is formed. Table~\ref{tab:edge} lists the
conventions; all identities in the paper (secant slope, calibration derivatives, zero-sum) are stated for a nondegenerate
row with $h>0$, a locally fixed interval index and the tie convention of the table; no epsilon or span clamp other than
$\varepsilon$ exists.

\begin{table}[htb]
\centering\footnotesize
\caption{Edge cases: forward value and backward convention as implemented.}
\label{tab:edge}
\begin{tabular}{@{}p{3.5cm}p{4.85cm}p{4.75cm}@{}}
\toprule
Condition & Forward & Backward \\
\midrule
Degenerate MinMax row ($M-m\le\varepsilon$): single valid key or equal valid scores & weights set to 1 (\texttt{torch.where}), invalid keys 0, $P$ uniform (softmax's forward) & zero gradient through \texttt{where}, extremum gradients cancel; finite \\
Very small positive span ($10^{-6}>\varepsilon$) & regular branch, tiny $h$, normal reconstruction & all gradients finite \\
\FW: single key or equal scores & $z_j=0$ for every valid key, all at the top grid value, $P$ uniform; no degenerate branch & regular formulas \\
Tied maxima/minima, positive span & unique-extremum formulas & \texttt{amax}/\texttt{amin} split the extremum gradient evenly among ties \\
Grid-boundary index switch & $r_j$ is a floor, piecewise constant & zero derivative a.e.; LERP weight continuous, secant identities on either side \\
\CH midpoint tie $t_j=\tfrac12$ & rounds up; hard weight discontinuous & \eqref{eq:hardjac} applies away from these points \\
\FW cutoff $z_j<-\tau$ & masked to $w_j=0$ after clamping; jump at $z=-\tau$ & zero gradient (mask and clamp); statements apply to in-window keys \\
\bottomrule
\end{tabular}
\end{table}

Every run of suites A--D includes the full calibration gradient defined by its calibration rule, and runs with a
detached calibration quantity are labeled as such; Table~\ref{tab:labels} therefore lists only
the three design axes. \FW--LERP is trained in Appendix~\ref{app:rb-fwmlerp}.

Under MinMax the interval width follows the learned row span, $h=\mathrm{span}/K$; under \FW it is the constant
$h=\tau/K=6/K$: at $K{=}4$, $h=1.5$ nats under \FW against a quarter of the row span under MinMax (6--8 nats at
softmax-like spans of 25--32, far more once the span expands).

\section{Gradient derivations}
\label{app:grad}

Throughout, $g_j=\partial L/\partial P_j$ is the upstream gradient at the attention probabilities of one
row, always evaluated at the forward (for \CH: hard) probabilities. The value path $O=PV$ uses the
forward $P$: $\partial L/\partial V=P^\top\partial L/\partial O$.

\paragraph{Normalization Jacobian.} For $P=w/W$, $\partial P_i/\partial w_j=(\delta_{ij}-P_i)/W$, so
$\gamma_j\equiv\partial L/\partial w_j=\frac1W\big(g_j-\sum_iP_ig_i\big)$.

\paragraph{Softmax.} $\partial L/\partial s_j=P^E_j(g_j-\sum_iP^E_ig_i)$. The implementation subtracts $M$
before exponentiating; autograd routes a gradient through $M$ that cancels exactly by shift invariance.

\paragraph{LERP own-score slope.} With the grid held fixed,
$\Delta_j=\partial w^L_j/\partial z_j=(e^{b_{r_j+1}}-e^{b_{r_j}})/h$: the floor defining $r_j$ is locally
constant (zero derivative almost everywhere) and the clamp on $t$ acts as the identity in the interior of an
interval. Compared with the exact derivative $e^{z_j}$, $\Delta_j$ is constant within an interval.

\paragraph{Full calibration gradient (MinMax).} Since $z_{\min}=m-M$ and $h=(M-m)/K$, the grid values
$b_{r_j},b_{r_j+1}$ and the width $h$ all depend on $M$ and $m$. Writing $\rho_j=(r_j+t_j)/K\in[0,1]$ for
the normalized position of key $j$ in $[z_{\min},0]$ and
$X_j=\tfrac1K[(1-t_j)r_je^{b_{r_j}}+t_j(r_j{+}1)e^{b_{r_j+1}}]$, differentiating $w^L_j$ through $z_j$,
$b_{r_j}$, $b_{r_j+1}$ and $h$ gives \eqref{eq:calgrad}. PyTorch's \texttt{amax}/\texttt{amin} split the
extremum gradient evenly among ties. Adding the three per-key partials gives
$\Delta_j+\partial w_j/\partial M+\partial w_j/\partial m=0$ for every $j$, hence $\sum_i\partial
L/\partial s_i=0$.

\paragraph{Detach.} $\partial L/\partial s_i|_{\mathrm{detach}}=\gamma_i\Delta_i$. Since
$\sum_i\gamma_i\Delta_i\ne0$ in general, each update carries a spurious common-mode component along a direction that
cannot change the loss, and the relative gradients between keys are also wrong. The projection control of \S\ref{sec:res-detach} keeps the detached terms
and projects each causal row onto the zero-sum subspace, $g'_{ij}=g_{ij}-\tfrac1{i+1}\sum_{k\le i}g_{ik}$, removing the
common-mode component only.

\paragraph{Weight-STE.} $w=w^L+\sg(w^N-w^L)$, $P=w/\sum_lw_l$. Forward $w=w^N$, $P=P^N$. Backward: the
normalization Jacobian is evaluated at the hard point and the scalar derivatives (including the calibration
terms) are those of LERP,
\[
\begin{aligned}
\frac{\partial L}{\partial s_i}&=\gamma^{W}_i\Delta_i
+\ind[i{=}\arg\max]\sum_j\gamma^{W}_j\frac{\partial w^L_j}{\partial M}
+\ind[i{=}\arg\min]\sum_j\gamma^{W}_j\frac{\partial w^L_j}{\partial m},\\
\gamma^{W}_j&=\frac{g_j-\sum_iP^N_ig_i}{W^N}.
\end{aligned}
\]

\paragraph{Prob-STE.} $P^N=\sg(w^N/W^N)$, $P^L=w^L/W^L$, $P=P^L+\sg(P^N-P^L)$. Forward $P=P^N$. Backward is
the full Jacobian of normalized LERP at the LERP point: the same expression with
$\gamma^{P}_j=(g_j-\sum_iP^L_ig_i)/W^L$.

\paragraph{\FW.} $w_j$ depends on the scores only through $\tilde z_j=\mathrm{clip}(s_j-M,-\tau,0)$.
For in-window keys $\partial w^L_j/\partial s_j=\Delta_j$ and $\partial w^L_j/\partial M=-\Delta_j$; tail
keys have $w_j=0$ and zero gradient through both the mask and the clamp; the row minimum plays no role. Thus
$\partial L/\partial s_i=\gamma_i\Delta_i-\ind[i{=}\arg\max]\sum_{j\in\mathrm{window}}\gamma_j\Delta_j$
with $\gamma=\gamma^{W}$ or $\gamma^{P}$ (tail excluded from $W^L,P^L$), and the row gradient again sums to
zero.

\paragraph{What differs between the surrogates.}\enlargethispage{-3pt} Both share the hard forward, $g$, the value-path gradient and
$\Delta$, $\partial w/\partial M$, $\partial w/\partial m$; they differ only in the point at which the normalization
Jacobian is evaluated, $\gamma^{W}=(g-\langle g\rangle_{P^N})/W^N$ versus $\gamma^{P}=(g-\langle g\rangle_{P^L})/W^L$:
\[
\gamma^{W}_j-\gamma^{P}_j=(g_j-\langle g\rangle_{P^N})\Big(\frac1{W^N}-\frac1{W^L}\Big)+\frac{\langle g\rangle_{P^L}-\langle g\rangle_{P^N}}{W^L}
\]
vanishes when $w^N=w^L$ and otherwise depends on the within-interval positions of the scores (through $w^N-w^L$), on
the reconstructed weights and probabilities and on $g$, so it is not proportional to $h$ and need not be monotone in
$h$. The main text uses only the qualitative statement that a wider interval permits a larger $|w^N-w^L|$; the empirical
decay of the surrogate effect with $K$ is a measured result, not a consequence of these formulas. Under \FW $h=\tau/K$
regardless of the learned tail, and tail keys contribute to neither $W$.

\paragraph{Derivative of the full hard map.} Under MinMax the grid value at level $\ell$ is
$b_\ell=(1-\ell/K)(m-M)$, so with the index $\ell_j$ held fixed (away from index switches and midpoint
ties) and unique extrema, $\partial b_{\ell_j}/\partial m=c_j$ and $\partial b_{\ell_j}/\partial M=-c_j$
with $c_j=1-\ell_j/K$, and
\begin{equation}
\partial w^N_j/\partial s_i=c_j\,w^N_j\big(\ind[i{=}\arg\min s]-\ind[i{=}\arg\max s]\big),
\label{eq:hardjac}
\end{equation}
generally nonzero (a two-key row is exactly the two-element softmax). This derivative is \emph{not} what any trained
operator back-propagates: in both STE implementations the hard branch enters only inside the stop-gradient (as
$w^N-w^L$ or $P^N-P^L$), so its dependence on $M$ and $m$ is cut and the calibration derivatives that reach $s$ are those
of $w^L$. Under \FW $b_\ell=-\tau+\ell h$ is data-independent and the hard weights are locally constant when index and
tail mask are fixed.

\paragraph{Verification.} A numerical check confirms \eqref{eq:hardjac} in
float64 on random rows away from boundaries (max deviation $2.8{\times}10^{-17}$ from autograd,
$8.8{\times}10^{-11}$ from central finite differences), the two-key softmax identity ($10^{-16}$) and the zero Jacobian of
the \FW hard map. A second check compares the closed forms above with autograd of the training code on random
float64 rows for MinMax and \FW $\times$ \{LERP, Weight-STE, Prob-STE\}: all
agree to $<10^{-9}$, and every autograd row gradient sums to zero. The gradient-consistency diagnostics of the detached
backward are in Table~\ref{tab:graddiag}.
\section{Protocol details}
\label{app:protocol}

\subsection{Training, hardware and conditions}
\label{app:protocol-train}

\paragraph{Training.} Within a suite every run shares the model, data and optimizer protocol; differences in training stack and in
gradient checkpointing are given in Tables~\ref{tab:dstack} and~\ref{tab:hardware} and below. 124M: 12 layers, 12 heads, $d{=}768$, 1024-token
context. 1B: 32 layers, 24 heads, $d{=}1536$ ($\approx$983M parameters). AdamW, $\beta_1{=}0.9$, $\beta_2{=}0.95$, weight
decay 0.1, gradient clip 1.0, linear warmup then cosine decay to $0.1\times$ peak, micro-batch 1 with gradient accumulation
to 131,072 tokens per step. Peak LR $6{\times}10^{-4}$ (124M) and $3{\times}10^{-4}$ (1B). Warmup 25M tokens (250M and 2.5B
suites), 10M (100M suites). Final steps: 1908 (250,085,376 tokens), 19,074 (2,500,067,328), 763 (100,007,936). Two 2.5B
runs (MinMax--Prob $K{=}4$, \FW--Prob $K{=}4$) were interrupted and resumed at 500.0M and 625.1M tokens; the
trajectories show no discontinuity.

\paragraph{Numerics and hardware.} Every training script enables TF32, so $QK^\top$ and $PV$ are TF32 tensor-core products during training while the operator itself (calibration,
reconstruction, normalization, surrogate) runs on fp32 tensors inside a disabled-autocast region; everything outside
attention runs under bf16 autocast; evaluation and all diagnostics use the same split with TF32 off.
Table~\ref{tab:hardware} lists the training and evaluation hardware and software of each suite. GPU hours are
approximate and count training only. Gradient checkpointing: six suite-D runs on seed 1337 (softmax, LERP $K{=}4/8/16/32$, MinMax--Weight
$K{=}32$) were trained with it enabled and the other 124 without; step-0 gradient equivalence with and without it was checked
on the inputs of Appendix~\ref{app:detach} only. Table~\ref{tab:dstack} splits suite D by operator family and training
stack: all 50 Prob-STE runs and the \FW--Weight $K{=}32/64$ runs were trained on RTX~5090, the other families on
the RTX~3090, so training stack is associated with operator family in suite D and the historical surrogate
contrasts also reflect training-stack differences; same-GPU evaluation controls evaluation-stack differences only. The
strict-forward runs (Appendix~\ref{app:strict}) give a matched comparison within one training stack at seed 7; their
differences from the historical runs (Table~\ref{tab:strict}) combine training stack and forward implementation and do
not isolate a hardware effect. The experiments of Appendix~\ref{app:rb} add \rbn{I.runs} training runs and
\rbn{I.gpuh} training GPU-hours on RTX~5090 (stack S1).

\begin{table}[htb]
\begin{minipage}[t]{0.48\textwidth}\centering\scriptsize
\caption{Suite D: training runs per operator family and training GPU. Software: stack S3 for the RTX~3090 runs
and S2 for the RTX~5090 runs (Table~\ref{tab:hardware}).}
\label{tab:dstack}
\adjustbox{max width=\linewidth}{\begin{tabular}{lrrr}
\toprule
Family & RTX 3090 & RTX 5090 & runs \\
\midrule
softmax & 5 & 0 & 5 \\
LERP & 25 & 0 & 25 \\
MinMax--Weight & 25 & 0 & 25 \\
MinMax--Prob & 0 & 25 & 25 \\
FWM--Weight & 15 & 10 & 25 \\
FWM--Prob & 0 & 25 & 25 \\
\midrule all & 70 & 60 & 130 \\
\bottomrule
\end{tabular}
}
\end{minipage}\hfill
\begin{minipage}[t]{0.50\textwidth}\centering\scriptsize
\caption{Secondary frozen-checkpoint audit (historical expressions only): Prob-STE frozen code minus the shared Weight-STE-expression evaluator, $\Delta$ per block on the full validation split (976 blocks for suites A and B, 243 for C); mean $\Delta$ over blocks and the largest single-block $|\Delta|$. Neither side is a direct hard forward (Table~\ref{tab:strictaudit}); this audit alone covers suite C and $K{=}16$.}
\label{tab:rounding}
\begin{tabular}{lrr}
\toprule
Run (suite: condition) & mean $\Delta$ & max $|\Delta|$ \\
\midrule
B: MinMax--Prob K4 & $3.28\mathrm{e}{-5}$ & $2.14\mathrm{e}{-2}$ \\
B: MinMax--Prob K16 & $-3.37\mathrm{e}{-5}$ & $8.90\mathrm{e}{-3}$ \\
B: FWM--Prob K4 & $1.08\mathrm{e}{-4}$ & $9.90\mathrm{e}{-3}$ \\
B: FWM--Prob K16 & $0$ & $0$ \\
A: MinMax--Prob K4 & $-9.91\mathrm{e}{-5}$ & $1.38\mathrm{e}{-2}$ \\
A: FWM--Prob K4 & $2.33\mathrm{e}{-5}$ & $1.41\mathrm{e}{-2}$ \\
C: MinMax--Prob K4 & $3.48\mathrm{e}{-4}$ & $1.36\mathrm{e}{-2}$ \\
C: FWM--Prob K4 & $-3.58\mathrm{e}{-4}$ & $7.98\mathrm{e}{-3}$ \\
\bottomrule
\end{tabular}
\end{minipage}
\end{table}

\begin{table}[htb]
\centering\scriptsize
\caption{Training and evaluation hardware per suite and group. Stacks: S1 $=$ RTX~5090, torch 2.13.0+cu132,
py3.12; S2 $=$ RTX~5090, PyTorch version not recorded; S3 $=$ RTX~3090, torch 2.5.1+cu121, py3.11. GPU-h: approximate
single-GPU hours, training only.}
\label{tab:hardware}
\adjustbox{max width=\textwidth}{\begin{tabular}{>{\raggedright\arraybackslash}p{5.2cm}rp{5.0cm}>{\raggedright\arraybackslash}p{2.4cm}r}
\toprule
Suite / group & Runs & Training stack (one GPU per run) & Evaluation GPU & GPU-h \\
\midrule
A (124M @ 2.5B) & 15 & S1 (13 runs); S2 (2 Prob-STE runs) & RTX 5090 & 763 \\
B (124M @ 250M) & 10 & S1 (5 runs); S2 (5 runs) & RTX 5090 & 56 \\
C (1B @ 100M) & 12 & S1 (10 runs); S2 (2 Prob-STE runs) & RTX 5090 & 245 \\
D (124M @ 100M, 5 seeds) & 130 & S3 (70 runs); S2 (60 runs); family split: Table~\ref{tab:dstack} & RTX 3090 & 498 \\
\midrule Historical detach record (\S\ref{sec:res-detach}) & 16 & S3 & RTX 3090 & --- \\
Detach replication, seeds 7/42 (App.~\ref{app:detach}) & 2 & S3 & RTX 3090 & 11 \\
Single-channel ablation, seeds 7/42 (App.~\ref{app:detach}) & 4 & S3 & RTX 3090 & 18 \\
Strict-forward retraining (App.~\ref{app:strict}) & 4 & S1 & RTX 5090 and RTX~3090 & 11 \\
Tail-policy (App.~\ref{app:taildecomp}) & 2 & S3 & RTX 3090 & 16 \\
Reduced-precision exponential baselines (App.~\ref{app:lowprec}) & 2 & S3 & RTX 3090 & 4.5 \\
Bin-mean control (App.~\ref{app:binmean}) & 5 & S3 & RTX 3090 & 31 \\
\midrule Appendix~\ref{app:rb} (a 2-GPU and an 8-GPU RTX 5090 machine) & 98 & S1 (Python 3.12.11, CUDA 13.2) & RTX 5090 & 454.1 \\
\bottomrule
\end{tabular}}
\end{table}

\paragraph{Conditions per suite.} A (124M @ 2.5B, 15 runs): softmax; LERP $K{=}1/2/4/16/32$;
MinMax--Weight $K{=}4/16/32/64$; MinMax--Prob $K{=}4$; \FW--Weight $K{=}2/4/16$; \FW--Prob $K{=}4$; 20 checkpoints
per run. B (124M @ 250M, 10 runs): softmax; LERP $K{=}4$; \{MinMax, \FW\}$\times$\{Weight-STE, Prob-STE\}$\times\{K{=}4,16\}$;
10 checkpoints per run. C (1B @ 100M, 12 runs): softmax; \{LERP, MinMax--Weight, \FW--Weight\}$\times\{K{=}4,16,32\}$;
MinMax--Prob $K{=}4$; \FW--Prob $K{=}4$; 5 checkpoints per run. D (124M @ 100M, 130 runs): softmax and
\{LERP, MinMax--Weight, MinMax--Prob, \FW--Weight, \FW--Prob\}$\times\{K{=}4,8,16,32,64\}$ on seeds
7/42/253/1337/2026; 5 checkpoints per run. Suites A--D total 167 runs and 1,106 evaluated checkpoints (Table~\ref{tab:assets}); reproduction tolerances of the
endpoint evaluations are given under ``Common limits'' in Appendix~\ref{app:fwdaudits}.

\paragraph{Corpora.} FineWebEdu-3B and WikiText-103, GPT-2 BPE. Validation: 976 non-overlapping
1024-token blocks (FineWebEdu) and 243 (WikiText-103); a block is $x=\mathrm{data}[kT{:}kT{+}T]$,
$y=\mathrm{data}[kT{+}1{:}kT{+}T{+}1]$.

\subsection{Forward implementation: three comparisons kept apart}
\label{app:fwdaudits}

\paragraph{Identity statements.} (i) \emph{Mathematical}: Weight-STE and Prob-STE define the same hard forward $P^N$.
(ii) \emph{Implementation}: a single shared operator implementation, which evaluates the Weight-STE expression
$w^L+(w^N-w^L)$, reproduces the frozen softmax, LERP $K{=}1$ and Weight-STE code bit for bit on the parity blocks, but not
the Prob-STE code, which evaluates $P^L+(P^H-P^L)$ in fp32 with $P^H$ itself normalized from the Weight-STE expression.
Neither historical surrogate evaluates $P^N$ bit-exactly on GPU at trained score scales, and \CH models amplify such
differences through rounding decisions in later layers, so the Prob-STE runs were trained on a forward that differs from
the Weight-STE forward at the rounding level: a confound of the surrogate contrast that frozen-code evaluation reports
but does not remove. The detach/full modes instead share one forward code path differing only by \texttt{.detach()} on
the row extrema, so their forwards are identical by construction. Batch 1 vs batch 8 differs by up to
$8.1{\times}10^{-3}$ on single blocks, which is why every number uses batch 1. The three comparisons below concern
floating-point implementations of one mathematical forward and change no number in the tables of suites A--D; each states its object, scope,
result and own limit, and the shared limits follow once.

\paragraph{(i) Historical Prob-STE code vs the shared Weight-STE expression (secondary audit).} \emph{Object}: the two
historical expressions against each other, not against a direct hard forward. \emph{Scope}: all 25 \CH runs of suites
A--C, evaluated twice on their full validation split (976 or 243 blocks, batch 1, same numerics) with the frozen training
code and with the shared implementation; the only comparison covering suite C and $K{=}16$. \emph{Result}: the 17
Weight-STE runs read exactly zero on every block, so the shared implementation matches the Weight-STE code; for the 8
Prob-STE runs (Table~\ref{tab:rounding}) the difference in mean NLL is at most $1.1{\times}10^{-4}$ nats on the 976-block
FineWebEdu splits and $3.6{\times}10^{-4}$ on the 243-block WikiText split, with inconsistent sign (four positive, three
negative, one exactly zero) and per-block differences that reach $2.1{\times}10^{-2}$ but cancel in the mean: a
frozen-checkpoint forward discrepancy, much smaller than the MinMax $K{=}4$ surrogate contrasts ($-0.32$ to $-0.83$
nats) and the pooled five-seed surrogate effect ($-0.021$), comparable to the smallest contrasts discussed (about
$0.001$, \FW at $K{=}16$). \emph{Limit}: it does not bound the deviation of either surrogate from $P^N$, and the
single-block maximum is an observed value, not a sensitivity bound.

\paragraph{(ii) Historical native forward vs a direct hard forward on frozen endpoints.}
\label{app:strict}
\emph{Object}: each historical implementation against a direct hard forward on the same trained weights. In the historical
code the hard value is never computed directly (Weight-STE evaluates $w^H=w^L+\sg(w^N-w^L)$ and
$P^H=\mathrm{normalize}(w^H)$; Prob-STE evaluates $P=P^L+\sg(P^H-P^L)$ with that same nested $P^H$; $P^H=P^N$ in exact
arithmetic but not bitwise), so a strict implementation must replace both the weight-level and the probability-level
expression. \emph{Scope}: 12 frozen $K{=}4$ endpoints (suites A, B and D seed 7; MinMax and \FW; Weight-STE and Prob-STE;
124M), each evaluated twice on one RTX 3090 with its own frozen code root, native and with only the surrogate combination
rewritten to $\sg(x^N)+(x^L-\sg(x^L))$ at both levels, whose forward was verified bit-identical to the directly computed
$w^N/\sum w^N$ on random rows and on the real scores of every layer. \emph{Result} (Table~\ref{tab:strictaudit}):
replacing the historical forward by the direct hard forward changes mean validation NLL by at most $3.4{\times}10^{-4}$
nats (largest observed $3.32{\times}10^{-4}$), while single blocks differ by up to $0.026$ nats, so the mean bound is not
a per-block bound. \emph{Limit}: an evaluation-time effect at these frozen checkpoints on this stack, not extending to
suite C, $K{=}16$ or unevaluated checkpoints; the two CIs that exclude zero are not a finding, nor the ten that include
zero bitwise equivalence. The tables of suites A--D keep the native evaluations.

\begin{table}[htb]
\centering\scriptsize
\caption{Frozen-endpoint audit: historical native forward minus direct hard forward, 12 $K{=}4$ endpoints (124M; suites A
and B on their 976-block FineWebEdu split, suite D seed 7 on the 243-block WikiText-103 split), both forwards on one RTX
3090 with the endpoint's own frozen code root. $\Delta$ mean NLL in nats with 95\% circular block-bootstrap CI (block 16,
4000 resamples); last column: largest single-block $|\Delta|$. Largest $|\Delta|$ mean over the 12 endpoints:
$3.32{\times}10^{-4}$.}
\label{tab:strictaudit}
\adjustbox{max width=\textwidth}{\begin{tabular}{lllrlll}
\toprule
Surrogate & Suite & Calib. & blocks & $\Delta$ mean NLL & 95\% CI & max $|\Delta|$ per block \\
\midrule
Weight-STE & A: 124M @ 2.5B & FWM & 976 & $3.04\mathrm{e}{-5}$ & [$-6.36\mathrm{e}{-5}$, $1.25\mathrm{e}{-4}$] & $8.42\mathrm{e}{-3}$ \\
Weight-STE & A: 124M @ 2.5B & MinMax & 976 & $-3.32\mathrm{e}{-4}$ & [$-6.40\mathrm{e}{-4}$, $-2.46\mathrm{e}{-5}$] & $2.61\mathrm{e}{-2}$ \\
Weight-STE & B: 124M @ 250M & FWM & 976 & $1.00\mathrm{e}{-4}$ & [$5.60\mathrm{e}{-6}$, $1.93\mathrm{e}{-4}$] & $8.68\mathrm{e}{-3}$ \\
Weight-STE & B: 124M @ 250M & MinMax & 976 & $-8.29\mathrm{e}{-5}$ & [$-4.56\mathrm{e}{-4}$, $2.91\mathrm{e}{-4}$] & $2.26\mathrm{e}{-2}$ \\
Weight-STE & D: 124M @ 100M, seed 7 & FWM & 243 & $1.03\mathrm{e}{-5}$ & [$-1.38\mathrm{e}{-4}$, $1.68\mathrm{e}{-4}$] & $6.47\mathrm{e}{-3}$ \\
Weight-STE & D: 124M @ 100M, seed 7 & MinMax & 243 & $-2.12\mathrm{e}{-4}$ & [$-6.54\mathrm{e}{-4}$, $2.53\mathrm{e}{-4}$] & $8.70\mathrm{e}{-3}$ \\
\midrule
Prob-STE & A: 124M @ 2.5B & FWM & 976 & $-8.41\mathrm{e}{-5}$ & [$-2.69\mathrm{e}{-4}$, $1.02\mathrm{e}{-4}$] & $1.12\mathrm{e}{-2}$ \\
Prob-STE & A: 124M @ 2.5B & MinMax & 976 & $-4.58\mathrm{e}{-5}$ & [$-3.00\mathrm{e}{-4}$, $2.17\mathrm{e}{-4}$] & $1.43\mathrm{e}{-2}$ \\
Prob-STE & B: 124M @ 250M & FWM & 976 & $-1.34\mathrm{e}{-4}$ & [$-3.38\mathrm{e}{-4}$, $5.73\mathrm{e}{-5}$] & $1.13\mathrm{e}{-2}$ \\
Prob-STE & B: 124M @ 250M & MinMax & 976 & $-4.13\mathrm{e}{-5}$ & [$-3.83\mathrm{e}{-4}$, $2.91\mathrm{e}{-4}$] & $2.39\mathrm{e}{-2}$ \\
Prob-STE & D: 124M @ 100M, seed 7 & FWM & 243 & $-1.09\mathrm{e}{-4}$ & [$-3.78\mathrm{e}{-4}$, $1.69\mathrm{e}{-4}$] & $8.83\mathrm{e}{-3}$ \\
Prob-STE & D: 124M @ 100M, seed 7 & MinMax & 243 & $2.46\mathrm{e}{-4}$ & [$-1.53\mathrm{e}{-4}$, $6.24\mathrm{e}{-4}$] & $1.09\mathrm{e}{-2}$ \\
\bottomrule
\end{tabular}}
\end{table}

\paragraph{(iii) Retraining with a strict hard forward.} \emph{Object}: whether the surrogate contrast depends
on the rounding of the historical Prob-STE forward. \emph{Scope}: the four $K{=}4$ cells of suite D retrained at seed 7
(124M, WikiText-103, 100M tokens) with a forward that returns $P^N$ bit for bit, $P=\sg(P^N)+\big(P^L-\sg(P^L)\big)$ for
Prob-STE and $w=\sg(w^N)+\big(w^L-\sg(w^L)\big)$, $P=w/\sum_lw_l$, for Weight-STE, from one shared hard reconstruction ($P^N$
detached because the MinMax hard branch still carries a gradient through the extrema; parentheses fix the evaluation
order); the identity was verified against the direct $P^N$ on synthetic, boundary and real seed-7 scores on CPU and GPU,
and the two placements give bit-identical forwards at equal parameters. Both surrogates were retrained because the
historical Weight-STE forward is itself not bit-identical to $w^N$ on GPU at trained score scales (the subtraction
leaves the Sterbenz range as $h$ grows under MinMax; under \FW a 1-ulp CPU/GPU difference in $e^{-4.5}$). Training ran on
an RTX 5090 with torch 2.13, unlike the RTX 3090 / torch 2.5.1 of the seed-7 controls; endpoints were evaluated on both
GPUs. \emph{Result} (Table~\ref{tab:strict}): the direction of both contrasts and the MinMax magnitude are unchanged; the
larger \FW value comes mostly from the new Prob-STE run ($-0.012$ against the historical one). \emph{Limit}: one pair of
runs cannot separate rounding, environment and training nondeterminism (identical-code replays on the same hardware
diverge within 50 steps), so the change is not attributed to the forward; the historical five-seed SD for that cell
(0.013) is background only; the small single-seed advantage of the new \FW--Prob run over softmax is not a general
improvement; the four runs are not pooled with the five-seed statistics, and the training-time effect in suites
A--C remains unmeasured.

\begin{table}[htb]
\centering\scriptsize
\caption{Strict-forward replication at $K{=}4$ (hard Nearest, 124M, WikiText-103, 100M tokens, seed 7; $\Delta$NLL in nats, paired per validation block, 95\% circular block-bootstrap CIs over 243 blocks, evaluation sampling only). Interaction follows Table~\ref{tab:cross}, $(\mathrm{FWM}_P-\mathrm{FWM}_W)-(\mathrm{MM}_P-\mathrm{MM}_W)$. Groups: strict pairs on the training GPU; the same pairs on a secondary evaluator; comparisons against the historical seed-7 runs on frozen weights; historical background, which is not a same-stack control.}
\label{tab:strict}
\adjustbox{max width=\textwidth}{\begin{tabular}{llrl}
\toprule
Contrast (nats) & Evaluator & $\Delta$ & 95\% CI \\
\midrule
\multicolumn{4}{@{}l}{\emph{Strict pairs: the four strict-forward runs, strict-hard forward, evaluated on the training GPU}} \\
MinMax: Prob $-$ Weight (strict pair) & RTX 5090 & -0.0770 & [-0.0812, -0.0729] \\
FWM: Prob $-$ Weight (strict pair) & RTX 5090 & -0.0390 & [-0.0422, -0.0359] \\
Interaction $(\mathrm{FWM}_P-\mathrm{FWM}_W)-(\mathrm{MM}_P-\mathrm{MM}_W)$ & RTX 5090 & +0.0381 & [+0.0331, +0.0425] \\
\multicolumn{4}{@{}l}{\emph{Secondary evaluator: the same strict pairs re-evaluated on the RTX 3090}} \\
MinMax: Prob $-$ Weight (strict pair) & RTX 3090 & -0.0772 & [-0.0816, -0.0731] \\
FWM: Prob $-$ Weight (strict pair) & RTX 3090 & -0.0388 & [-0.0423, -0.0356] \\
Interaction, as above & RTX 3090 & +0.0384 & [+0.0331, +0.0433] \\
\midrule \multicolumn{4}{@{}l}{\emph{Historical comparisons: new strict runs vs the historical seed-7 runs (frozen weights, different training stack)}} \\
MinMax: new Prob $-$ historical Prob, strict forward on both & RTX 5090 & +0.0008 & [-0.0024, +0.0040] \\
FWM: new Prob $-$ historical Prob, strict forward on both & RTX 5090 & -0.0121 & [-0.0154, -0.0090] \\
MinMax: new strict Weight $-$ historical Weight & RTX 3090 & -0.0017 & [-0.0047, +0.0012] \\
FWM: new strict Weight $-$ historical Weight & RTX 3090 & +0.0030 & [+0.0001, +0.0059] \\
MinMax: historical Prob, native $-$ strict forward (frozen weights) & RTX 5090 & -0.0001 & [-0.0005, +0.0003] \\
FWM: historical Prob, native $-$ strict forward (frozen weights) & RTX 5090 & +0.0000 & [-0.0003, +0.0003] \\
\midrule \multicolumn{4}{@{}l}{\emph{Background: historical seed-7 runs only (not a same-stack control)}} \\
MinMax: historical Prob $-$ historical Weight, seed 7 & RTX 3090 & -0.0793 & [-0.0834, -0.0752] \\
FWM: historical Prob $-$ historical Weight, seed 7 & RTX 3090 & -0.0240 & [-0.0264, -0.0216] \\
\bottomrule
\end{tabular}}
\end{table}

\paragraph{Common limits.} Neither (i) nor (ii) bounds accumulated training-time effects, and (iii), the only training-time
evidence, covers neither suites A--C nor other seeds. Every CI above covers evaluation sampling only. Reproduction
tolerances refer to specific checks: re-evaluating each suite's endpoints against its earlier independent evaluation
agreed within $4.5{\times}10^{-5}$ nats except MinMax--Weight $K{=}4$ ($5.4{\times}10^{-4}$ at 2.5B, $3.9{\times}10^{-4}$ at
1B). Across stacks they do not hold: in the
measured RTX 3090 / RTX 5090 comparisons (different GPUs and torch versions) mean NLL of the same checkpoint differed by up
to about $5{\times}10^{-4}$ nats with a maximum per-block difference of 0.0125, an
observed range, not a bound.
Same-GPU pairing keeps such offsets out of a contrast but does not make the native$-$strict difference itself invariant
across GPUs (for suite D MinMax--Prob its sign differs between the two GPUs); contrasts of about $0.001$ nats are therefore
read only within one evaluation stack.

\subsection{Selection of $\tau$, statistics and evaluation}
\label{app:protocol-eval}

\paragraph{Selection of $\tau$ (multi-stage).} $\tau=6$ was selected before any \FW training run by an inference-time
screen on frozen softmax-trained weights: the suite-A softmax endpoint, operator
swapped at inference, bf16 forward. The split was the WikiText-103 validation file on which suites C and D are
later evaluated, so WikiText-103 validation participated in the
selection; suites A and B are evaluated on a different corpus. Stages: (1) floor tail ($z\leftarrow-\tau$), 32 blocks,
$K\in\{4,16,32\}$, $\tau\in\{8,12,16,20,24,32\}$, $\tau{=}8$ best; (2a) floor tail, $\tau\in\{4,5,6,7,8,10,12\}$, $\tau{=}8$
best and collapse below 8; (2b) zero tail ($w{=}0$), same candidates, $\tau{=}7$ at $K{=}4$ and $\tau{=}6$ at $K{=}16,32$;
(3) zero tail, all 243 blocks, $K\in\{4,8,16,32\}$, $\tau\in\{5,6,7\}$. Stage-3 NLL (softmax 4.3119; 10,000-resample block
bootstrap CIs recorded per condition):

\begin{center}\scriptsize
\begin{tabular}{rrrrrl}
\toprule
$K$ & MinMax & $\tau{=}5$ & $\tau{=}6$ & $\tau{=}7$ & $\tau{=}6$ $-$ softmax [95\% CI] \\
\midrule
4 & 5.8590 & 4.3655 & \textbf{4.3613} & 4.3670 & $+0.0494$ [$+0.0467$, $+0.0522$] \\
8 & 5.3821 & 4.3311 & \textbf{4.3244} & 4.3285 & $+0.0125$ [$+0.0113$, $+0.0139$] \\
16 & 4.7333 & 4.3224 & \textbf{4.3132} & 4.3150 & $+0.0013$ [$+0.0007$, $+0.0020$] \\
32 & 4.3765 & 4.3200 & \textbf{4.3111} & 4.3115 & $-0.0008$ [$-0.0014$, $-0.0001$] \\
\bottomrule
\end{tabular}
\end{center}
The procedure is reproducible and inference-time only: it does not establish that $\tau=6$ is optimal for training, and
the wider candidates were eliminated on a 32-block subset. Two inference-time studies on other models report windows of
the same scale (IntAttention \citep{intattention2026}: $c=6.6$, stable for $c\in[5.5,7.7]$; EXAQ \citep{exaq2024}: optimal
clipping 3.3--8.0 nats); neither is a training result.

\paragraph{Contrasts and intervals.} A contrast $\sum_ca_c\,\mathrm{NLL}_c$ is computed per block and
averaged; $2{\times}2$ main effects average over the other factor with weights $\pm\tfrac12$ and the
interaction is $(\mathrm{FWM}_P-\mathrm{FWM}_W)-(\mathrm{MM}_P-\mathrm{MM}_W)$. Single-seed CIs: circular moving-block
bootstrap, block length 16, 4000 resamples, 95\% percentile. Five-seed: paired differences by seed, 95\%
$t$ interval, exact enumeration of all $2^5$ sign assignments for the permutation $p$ (floor
$2/32$), BH $q$ across the 25 conditions.

\paragraph{Geometry probe.} Forward pre-hooks on each attention layer recompute the fp32 scores exactly as the operator
does on the first 8 validation blocks (rows $i\ge1$; averaged over rows, reported per layer and as a layer mean): span
$M-m$; native width $h$; 0.1\% quantile of $z$; per-token query/key norms; softmax entropy on the model's own scores;
fraction of entries with $z<-6$; top-bucket mass (softmax mass on keys in the operator's top level or top interval);
effective buckets $\exp H(q)$ with $q_b\propto\sum_{j\in b}P^E_j$; $\mathrm{TV}=\tfrac12\sum_j|P^{\mathrm{native}}_j-P^E_j|$.

\paragraph{Checkpoints of suite D.} Suite D has 646 evaluated checkpoints rather than 650, because the
seed-1337 softmax run has only its endpoint checkpoint.

\paragraph{Downstream (2.5B suite only).} Pre-declared condition groups, predictions P1--P4, a $\pm0.01$-nat
practical-equivalence margin for NLL-type metrics only, and paired-bootstrap decision rules were fixed before any
downstream number was produced. Each NLL-type comparison carries two independent labels, \emph{detectability} (the
paired 95\% CI excludes zero / includes zero) and \emph{margin status} (the CI lies wholly inside the pre-declared
$\pm0.01$-nat margin / does not); ``within the pre-declared margin'' is used only for the first case of the second
label, ``no detectable difference'' only when the CI includes zero, never as a synonym for equivalence; no margin is
declared for accuracy metrics, so accuracy results are never called equivalent; split-half fluctuation is a descriptive
scale, not a threshold. Layer-1 $q$ values are BH within the 14 condition-vs-softmax comparisons of each corpus and
forward mode. \emph{Layer 1}: WikiText-103 test (276 blocks), PTB test (102), C4 validation (302 documents,
document-cluster bootstrap); split-half fluctuation 0.057/0.072/0.131 nats. \emph{Layer 2}: three synthetic probes from
random token sequences with a fixed generator seed: A, induction (a bigram $(A,B)$ inserted, $A$ recurs at distance
$d\in\{64,\dots,896\}$, predict $B$; 400 families $\times$ 5 distances $=$ 2,000 induction samples, each scored with a
matched control sequence without the earlier occurrence, 4,000 scored sequences in all); B, key--value retrieval (16--64
pairs, near/far queries, 2,400 samples); C, prefix copying (a random segment $S$ of length $L\in\{32,64,128\}$, a
separator, then $S$ again, the second copy scored per token; 300 samples per $L$, 900 in total); scores are per-token
target NLL and top-1 accuracy, and only probe C had adequate resolution at 124M. \emph{Layer 3}: lm-evaluation-harness
0.4.13 \citep{gao2023lmeval}, zero-shot, batch size 1, bf16 autocast outside the fp32 attention kernel, fp32 logits;
BLiMP (task \texttt{blimp}, validation, a fixed 200-pair subset of each of the 67 paradigms selected by the harness
\texttt{limit} argument with seed 20260920, 13,400 pairs), HellaSwag (validation, 10,042), PIQA (validation, 1,838),
ARC-Easy (test, 2,376), LAMBADA (\texttt{lambada\_openai}, test, 5,153), SciQ (test, 1,000) with the support passage and
a no-support variant (task YAML with the passage removed); comparisons paired per item on identical items in identical
order (paired bootstrap on per-item differences, 4000 resamples; permutation $p$, 20,000; BH within each configuration
$\times$ metric family of 14 comparisons); nothing is combined into a single score. Task selection: the
softmax 2.5B endpoint alone was first scored on nine benchmarks in ten configurations (the seven above plus WinoGrande,
ARC-Challenge and OpenBookQA); those three, on which softmax scored below chance (4.7--5.2 SE),
were not carried further, a choice made on softmax-only results and before any approximate condition was scored; the seven remaining
configurations were then fixed before the first approximate condition was scored, and no task or metric was added,
dropped or re-weighted afterwards.
\section{Seed robustness, horizon and score geometry: extended results}
\label{app:ext}

\subsection{Full cross-suite contrast table}
\begin{table}[htb]
\centering\scriptsize
\caption{Every matched contrast of Table~\ref{tab:cross} for suites A--D with its interval, plus the main effects, the $K{=}16$ terms of the 250M factorial and the condition-vs-softmax rows (final checkpoint, nats); the $K$-interaction rows exist only in the 250M factorial. Single-seed suites: 95\% circular block-bootstrap CI over validation blocks (evaluation sampling only); suite D: mean over five seeds, paired by seed, 95\% $t$-CI. Main effects average over the other factor; interaction $=(\mathrm{FWM}_P-\mathrm{FWM}_W)-(\mathrm{MM}_P-\mathrm{MM}_W)$.}
\label{tab:crossfull}
\adjustbox{max width=\textwidth}{\setlength{\tabcolsep}{3pt}\begin{tabular}{lllll}
\toprule
Contrast (nats) & B: 124M@250M & A: 124M@2.5B & C: 1B@100M & D: 124M@100M, 5 seeds \\
& FineWebEdu [block CI] & FineWebEdu [block CI] & WikiText [block CI] & WikiText, mean [$t$-CI] \\
\midrule
MinMax--Weight K4 $-$ softmax & +0.389 [0.385, 0.393] & +0.890 [0.881, 0.899] & +0.220 [0.214, 0.226] & +0.208 [0.191, 0.224] \\
MinMax--Prob K4 $-$ softmax & +0.074 [0.072, 0.075] & +0.061 [0.060, 0.063] & +0.146 [0.142, 0.151] & +0.137 [0.129, 0.145] \\
FWM--Weight K4 $-$ softmax & +0.039 [0.038, 0.040] & +0.043 [0.041, 0.045] & +0.061 [0.057, 0.065] & +0.032 [0.023, 0.040] \\
FWM--Prob K4 $-$ softmax & +0.007 [0.006, 0.008] & +0.019 [0.018, 0.021] & +0.038 [0.035, 0.041] & +0.013 [0.002, 0.024] \\
LERP K4 $-$ softmax & +0.052 [0.050, 0.053] & +0.005 [0.003, 0.006] & +0.096 [0.092, 0.100] & +0.098 [0.091, 0.106] \\
\midrule Prob $-$ Weight, MinMax, K4 & -0.316 [-0.319, -0.312] & -0.829 [-0.837, -0.820] & -0.074 [-0.078, -0.070] & -0.071 [-0.087, -0.054] \\
\bottomrule
\end{tabular}}
\end{table}

\subsection{Five-seed suite}
\begin{figure}[htb]
\centering
\includegraphics[width=\textwidth]{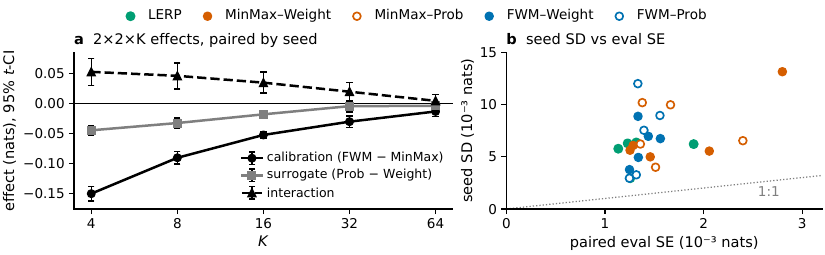}
\caption{\textbf{124M @ 100M on five seeds, WikiText-103.} (a) Calibration, surrogate and interaction
effects by $K$, paired by seed, 95\% $t$ intervals. (b) Seed SD of each condition's $\dNLL$ against the
paired evaluation SE of the same contrast; every point lies above the 1:1 line.}
\label{fig:seeds}
\end{figure}

\begin{table}[htb]
\begin{minipage}[t]{0.50\textwidth}\centering\scriptsize
\caption{Five-seed $2{\times}2{\times}K$ effects in nats: mean over five paired seed-level observations (seed SD); the 95\% $t$ interval is mean $\pm1.24\,\mathrm{SD}$ (Figure~\ref{fig:seeds}a), and every $K\le16$ interval excludes zero.}
\label{tab:seedfact}
\adjustbox{max width=\linewidth}{\begin{tabular}{lrrr}
\toprule
$K$ & calibration: FWM $-$ MinMax & surrogate: Prob $-$ Weight & interaction \\
\midrule
$K{=}4$ & -0.1497 (0.0094) & -0.0446 (0.0062) & +0.0523 (0.0180) \\
$K{=}8$ & -0.0902 (0.0086) & -0.0326 (0.0064) & +0.0457 (0.0173) \\
$K{=}16$ & -0.0524 (0.0046) & -0.0182 (0.0052) & +0.0344 (0.0139) \\
$K{=}32$ & -0.0303 (0.0077) & -0.0046 (0.0074) & +0.0194 (0.0121) \\
$K{=}64$ & -0.0134 (0.0069) & -0.0041 (0.0058) & +0.0042 (0.0085) \\
pooled & -0.0672 (0.0034) & -0.0208 (0.0027) & +0.0312 (0.0047) \\
\bottomrule
\end{tabular}}
\end{minipage}\hfill
\begin{minipage}[t]{0.46\textwidth}\centering\scriptsize\setlength{\tabcolsep}{5pt}%
\caption{Five-seed $K$ trend: slope of $\dNLL$ vs softmax in nats per doubling of $K$ (least squares against $\log_2K$ over $K\in\{4,\dots,64\}$, fitted per seed); mean over seeds, 95\% $t$-CI, seed SD.}
\label{tab:ktrend}
\begin{tabular}{lrlr}
\toprule
Family & slope & 95\% $t$-CI & seed SD \\
\midrule
LERP & -0.0222 & [-0.024, -0.020] & 0.0014 \\
MinMax--Weight & -0.0472 & [-0.051, -0.043] & 0.0033 \\
MinMax--Prob & -0.0301 & [-0.034, -0.027] & 0.0028 \\
FWM--Weight & -0.0078 & [-0.010, -0.005] & 0.0020 \\
FWM--Prob & -0.0030 & [-0.007, +0.001] & 0.0029 \\
\bottomrule
\end{tabular}

\end{minipage}
\end{table}

\paragraph{Five-seed details.} Figure~\ref{fig:seeds} and Tables~\ref{tab:seedfact}--\ref{tab:ktrend} give the effects and slopes; the endpoints are Table~\ref{tab:seedend} in Appendix~\ref{app:full}. softmax's own endpoint ranges 4.137--4.153 across seeds. Pairwise Spearman between
seeds is 0.83--0.92. The per-family slopes of $\dNLL$ against $\log_2K$ are in Table~\ref{tab:ktrend}; under \FW the gap
is already small at $K{=}4$ ($+0.032$ Weight-STE, $+0.013$ Prob-STE). The cost of approximation appears between 40M and 60M tokens for the conditions that degrade; at 20M no condition is
degraded (early-checkpoint contrasts pair the four seeds whose softmax run has intermediate checkpoints, the seed-1337
softmax run having only its endpoint; the 100M endpoints pair all five).

\FloatBarrier
\subsection{Horizon and $K$ details at 2.5B}
\begin{figure}[htb]
\centering
\includegraphics[width=3.4in]{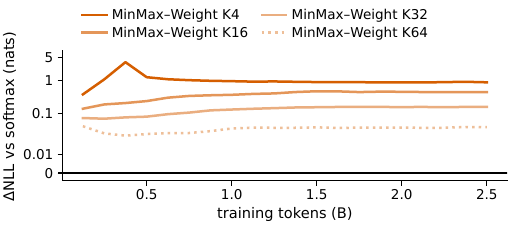}
\caption{\textbf{MinMax--Weight $K$ sweep along training at 2.5B}, $\dNLL$ vs softmax (symlog).}
\label{fig:ksweep}
\end{figure}
LERP $K{=}1$ (a single interval) is still improving at 2.5B ($+0.35$). \FW--Weight $K{=}4$ pays early ($+0.043$ at 25M,
the worst $K{=}4$ condition then), because a fixed 1.5-nat grid is coarse relative to early row spans of 3--4, and then
holds flat. MinMax--Weight $K{=}4$ passes through a collapse ($+3.66$ nats at 375M tokens, single-step gradient norms of
$10^4$, 98\% of steps clipped over the last 1.5B tokens) and partially recovers to its $+0.89$ plateau
(Figure~\ref{fig:ksweep}); no Prob-STE or \FW run clips abnormally.

\subsection{Tail policy in the MinMax $\to$ \FW change}
\label{app:taildecomp}
\paragraph{Definition and experiment.} Replacing MinMax by \FW changes the grid step near the row maximum (adaptive
$(M-m)/K$ to $\tau/K$), the dependence of the calibration and its backward on $m$, and the treatment of keys below the
window (kept on the grid versus set to zero). The diagnostic operator \emph{resolved} keeps \FW's step $\tau/K$ and grid
origin $-\tau$ but continues the grid below $-\tau$ without bound (negative grid indices), so every key is reconstructed
on the same fine grid and none is truncated; $K{=}4$ fixes the step within the window, not the number of usable levels.
Two contrasts follow: MinMax$\to$resolved, an untruncated fixed-step calibration contrast (step, its adaptivity, the
$m$-dependence of the backward and the number of usable levels change at once), and resolved$\to$\FW, a tail-policy
contrast at a fixed grid rule. They sum to the paper's MinMax$\to$\FW change along this path by construction, which
establishes neither a unique mechanism nor a causal mediation share. Two runs (resolved $\times$ Weight-STE and
$\times$ Prob-STE; Nearest, $K{=}4$, $\tau{=}6$, 124M @ 100M, WikiText-103, seed 1337, RTX 3090 / torch 2.5.1, historical
STE arithmetic rather than the strict forward of Appendix~\ref{app:strict}) were trained with the \FW code, to which a tail-policy option was added; in-window unnormalized weights are bit-identical to \FW's (probabilities are
not, since tail keys enter the denominator), the closed-form backward matches autograd to $4.4{\times}10^{-16}$, and
evaluation follows the paper's protocol.

\paragraph{Result.} Table~\ref{tab:taildecomp} gives the contrasts against the suite-D seed-1337 runs. The untruncated
fixed-step operator recovers most of the MinMax--\FW gap: it ends slightly below \FW under Weight-STE ($-0.012$ nats;
path ratio 1.08, interval excluding 1) and at a similar loss under Prob-STE ($+0.001$, within the 0.01-nat single-seed
margin; ratio 0.99). Zeroing the tail is therefore not
required for the observed improvement in this setting. Native probability mass on tail keys is $0.008$--$0.010$ under
resolved, $0$ under \FW and $0.018$--$0.023$ under MinMax, against $0.007$--$0.014$ softmax mass on the same keys;
resolved uses 14.5 and 19.1 distinct nonzero grid levels per row against at most 5 for \FW (five nonzero levels plus the
zero tail) and MinMax, so the comparison is not at equal deployment cost.

\paragraph{Limits.} One seed (intervals cover evaluation sampling only; the calibration terms exceed the suite-D seed SD
of 0.003--0.013 by an order of magnitude, the tail-policy terms do not), one $K$ and $\tau$, and no isolation of
resolution alone or of why the fixed-step grid carries the effect; the result does not show that truncation is
unnecessary elsewhere or equivalent at equal cost.

\begin{table}[htb]
\centering\scriptsize
\caption{Tail policy in the MinMax $\to$ \FW change (hard Nearest, $K{=}4$, $\tau{=}6$, 124M @ 100M, WikiText-103, seed 1337). Paired per validation block against the suite-D seed-1337 runs; 95\% circular block-bootstrap CIs (evaluation sampling only); $^{\circ}$: $|\Delta|<0.01$ nats, below the single-seed interpretation margin. The path ratio $(\mathrm{MinMax}-\mathrm{resolved})/(\mathrm{MinMax}-\mathrm{FWM})$ is descriptive and path-specific; it exceeds 1 when the tail-policy term has the opposite sign.}
\label{tab:taildecomp}
\adjustbox{max width=\textwidth}{\begin{tabular}{llrl}
\toprule
Surrogate & Contrast (nats) & $\Delta$ & 95\% CI \\
\midrule
Weight-STE & MinMax $-$ softmax & +0.1979 & [+0.1926, +0.2031] \\
Weight-STE & FWM $-$ softmax & +0.0403 & [+0.0366, +0.0438] \\
Weight-STE & resolved $-$ softmax & +0.0282 & [+0.0246, +0.0315] \\
Weight-STE & MinMax $-$ resolved (untruncated fixed-step calibration contrast) & +0.1697 & [+0.1645, +0.1748] \\
Weight-STE & resolved $-$ FWM (tail-policy contrast, fixed grid rule) & -0.0122 & [-0.0149, -0.0096] \\
Weight-STE & path ratio (descriptive) & 1.077 & [1.061, 1.095] \\ \midrule
Prob-STE & MinMax $-$ softmax & +0.1399 & [+0.1349, +0.1449] \\
Prob-STE & FWM $-$ softmax & +0.0130 & [+0.0095, +0.0159] \\
Prob-STE & resolved $-$ softmax & +0.0137 & [+0.0090, +0.0177] \\
Prob-STE & MinMax $-$ resolved (untruncated fixed-step calibration contrast) & +0.1262 & [+0.1220, +0.1309] \\
Prob-STE & resolved $-$ FWM (tail-policy contrast, fixed grid rule) & +0.0007 & [-0.0020, +0.0032] $^{\circ}$ \\
Prob-STE & path ratio (descriptive) & 0.995 & [0.975, 1.016] \\
\bottomrule
\end{tabular}}
\end{table}

\subsection{Reduced-precision exponential baselines}
\label{app:lowprec}

Appendix~\ref{app:rb-formats} extends these baselines to three seeds and five formats.

\paragraph{Purpose and protocol.} These two runs ask whether lowering the precision of the exponential,
with no $K$-interval grid, also pretrains from scratch. They use the suite-D protocol at seed 7 (124M, WikiText-103, 100M
tokens, no gradient checkpointing) on the RTX 3090 with torch 2.5.1+cu121, the stack of the seed-7 softmax control,
with the code of the five-seed LERP and MinMax--Weight runs, to which the two exponential modes were added.
Inside the attention path everything is fp32 as in every other run and the only change is
how $e^{z}$ is produced; there is no learned or data-dependent quantization scale beyond the row-max shift. The row maximum stays on
the autograd path in both runs: a detach field left in their saved configurations from older code does not detach it in
this code path.

\paragraph{Definitions.} \emph{BF16-exp}: $z$ is cast to bfloat16, the exponential is evaluated with a bfloat16 output
(input rounded before exponentiation, not only the output afterwards) and the weights are returned to fp32 for
normalization; the backward is native autograd through the casts, the reduced-precision gradient of this composite map
rather than a hand-written surrogate. \emph{FP8-rounded exp (FP32 exp backward)}: $w^{\mathrm{exp}}=e^{z}$ is computed in
fp32, rounded to \texttt{torch.float8\_e4m3fn} and returned to fp32,
$w=w^{\mathrm{fp8}}+\big(w^{\mathrm{exp}}-\sg(w^{\mathrm{exp}})\big)$, so the forward is the E4M3 value bit for bit (verified
against an independent evaluation on CPU and GPU) while the rounding step carries an identity surrogate and the
exponential keeps its fp32 derivative; the normalization Jacobian is evaluated at the rounded weights. This is a
weight-level straight-through placement, not a literal ``no-STE'' baseline, with the fp32 exponential rather than the
LERP reconstruction as surrogate. In E4M3 the exponential weights $w\in(0,1]$ can take 56 nonzero values (49 normal
values with exponents $-6$ to $0$ and 7 subnormal multiples of $2^{-9}$); round-to-nearest maps $w\le2^{-10}$ to zero,
i.e.\ $z$ below about $-6.93$ underflows, a rounding tail rather than \FW's explicit $\tau=6$ truncation (the exact
boundary is not claimed bitwise). The counts 56 and 5 refer to reconstructed weight values, not to distinct normalized
probabilities. In both modes the row maximum stays on the autograd path, so the implemented score gradient has the
row-zero-sum property in exact arithmetic; the measured residual $|\sum_j\partial L/\partial s_j|/\sum_j|\partial
L/\partial s_j|$ is at most $2.3{\times}10^{-7}$ on random rows at three score scales on CPU and GPU.

\paragraph{Endpoints.} Table~\ref{tab:lowprec} gives the 100M endpoints against the same-seed softmax, all evaluated on
one RTX 3090 (243 blocks, batch 1, TF32 off), with 95\% circular moving-block bootstrap intervals (block 16, 4000
resamples) over validation blocks; each row is one run. The new runs end $-0.0034$ [$-0.0058$, $-0.0009$] (BF16-exp) and
$+0.0028$ [$+0.0002$, $+0.0055$] (FP8-rounded exp) nats from softmax: small differences that the evaluation intervals
detect, both below 0.01 nats in magnitude. The training stack differs across rows and same-GPU evaluation does not remove
that, so the historical and strict \FW--Prob rows are background (historical \FW--Prob $K{=}4$ five-seed mean $+0.0132$,
SD $0.0089$, a seed-level spread, not an evaluation CI); the same-seed LERP and MinMax rows of the table are NLL gaps at the same scale, not all of them instabilities.

\begin{table}[htb]
\centering\scriptsize
\caption{Reduced-precision exponential baselines and same-seed references at 100M tokens (124M, WikiText-103, seed 7):
$\Delta$NLL against the seed-7 softmax run, paired per validation block, all evaluated on one RTX 3090 (243 blocks); 95\%
circular block-bootstrap CI over validation blocks (evaluation sampling only).
Historical \FW--Prob $K{=}4$ five-seed mean $+0.0132$ (SD $0.0089$) is seed-level background.}
\label{tab:lowprec}
\adjustbox{max width=\textwidth}{\begin{tabular}{p{5.2cm}p{5.3cm}rl}
\toprule
Operator (seed 7, 124M @ 100M) & Training GPU; software stack & $\Delta$NLL & 95\% eval.\ CI \\
\midrule
BF16-exp (new) & RTX 3090; torch 2.5.1+cu121 & -0.0034 & [-0.0058, -0.0009] \\
FP8-rounded exp, FP32 exp backward (new) & RTX 3090; torch 2.5.1+cu121 & +0.0028 & [+0.0002, +0.0055] \\
LERP $K{=}32$ & RTX 3090; torch 2.5.1+cu121 & +0.0047 & [+0.0024, +0.0069] \\
LERP $K{=}4$ & RTX 3090; torch 2.5.1+cu121 & +0.0942 & [+0.0913, +0.0969] \\
MinMax--Prob $K{=}4$ (historical) & RTX 5090; PyTorch version not recorded & +0.1452 & [+0.1412, +0.1493] \\
MinMax--Weight $K{=}4$ (historical) & RTX 3090; torch 2.5.1+cu121 & +0.2245 & [+0.2190, +0.2302] \\
FWM--Prob $K{=}4$ (historical) & RTX 5090; PyTorch version not recorded & +0.0083 & [+0.0057, +0.0107] \\
FWM--Prob $K{=}4$ (strict forward, App.~\ref{app:strict}) & RTX 5090; torch 2.13.0+cu132 & -0.0036 & [-0.0070, -0.0003] \\
\bottomrule
\end{tabular}}
\end{table}

\begin{figure}[htb]
\centering
\includegraphics[width=\textwidth]{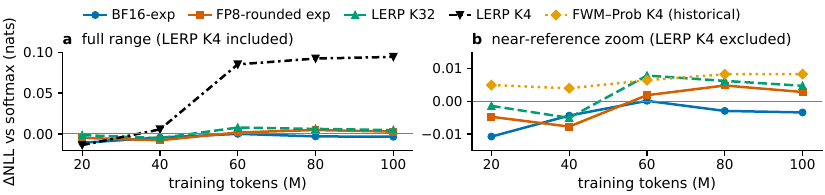}
\caption{\textbf{Reduced-precision exponential baselines} (suite-D protocol, 124M, WikiText-103, seed 7): validation
NLL minus the same-seed softmax run at the five evaluated checkpoints, all evaluated on one RTX 3090. (a) Full range
including LERP $K{=}4$; (b) near-reference trajectories, LERP $K{=}4$ excluded, with the historical \FW--Prob $K{=}4$ run
(different training stack). One run per curve; endpoint evaluation intervals in Table~\ref{tab:lowprec}.}
\label{fig:lowprec}
\end{figure}

\paragraph{Trajectories and stability.} Figure~\ref{fig:lowprec} shows the differences from the same-seed softmax run at
the five evaluated checkpoints (checkpoint values, not a continuous bound): both new runs stay within $\pm0.011$ nats of
softmax throughout, while LERP $K{=}4$ opens its gap between 40M and 60M. The training log (one point per 5M tokens)
shows no nonfinite values or excess loss increases (largest train-loss increase between consecutive logs 0.035 for
BF16-exp, 0.034 for FP8-rounded exp and 0.034 for softmax), which does not exclude excursions between logs. Final
pre-clip gradient norms are 0.497 / 0.482 / 0.498 and the 100M score geometry (layer mean over 8 validation blocks) is
close to softmax: row span 25.9 / 24.9 / 26.2, softmax entropy 3.75 / 3.75 / 3.74, query and key norms within 3\%
(BF16-exp / FP8-rounded exp / softmax); LERP $K{=}4$ has span 13.2 at the same point.

\paragraph{Native FP8 backward diagnostic.} With PyTorch's native backward for the cast, the gradient itself is cast to
E4M3. On the seed-7 initialization and the first training update (128 micro-batches, training precision, unscaled), this
flushed the $Q$/$K$ projection parameter gradients to exactly zero (L2 norm 0, all gradients finite); with the fp32
exponential backward used for training they are ordinary ($0.0485$). This is a
diagnostic of that unscaled implementation at initialization; it does not show that FP8 training is impossible or that
fp32 exponential gradients are necessary in general (gradient scaling and other backward designs were not tested).

\paragraph{Reading and limits.} Like \FW and unlike MinMax, both baselines use a fixed quantization/reconstruction rule in
row-max-shifted coordinates, independent of the learned row span, and their rounding does not coarsen as the span grows
(floating-point formats are not equally spaced grids; BF16-exp additionally rounds the input $z$), although the realized
mismatch still depends on the score distribution. This is consistent with the range--resolution reading of
\S\ref{sec:res-geom} but not a controlled test of it: format, grid density, tail treatment and backward rule all differ
from the $K$-interval operators. The controls show that the tested reductions in exponential precision remain close to
the fp32-softmax reference at this horizon; they establish neither equivalence (no equivalence test was run; the
downstream $\pm0.01$-nat margin is not applied here), nor a cross-seed ranking of BF16-exp, FP8-rounded exp, \FW--Prob
$K{=}4$ and softmax against training randomness (softmax's own five-seed endpoint SD is 0.0062), nor a cost-matched
comparison with five-level reconstruction (representable values are not a proxy for compute, storage width or speed);
one seed, 100M tokens, no longer horizon or coarser format (FP4, 4-bit lookup). The calibration and surrogate
interventions elsewhere provide the controlled evidence for the large failures within the studied operator family.

\subsection{Score geometry along training at 2.5B}
\begin{figure}[htb]
\centering
\includegraphics[width=\textwidth]{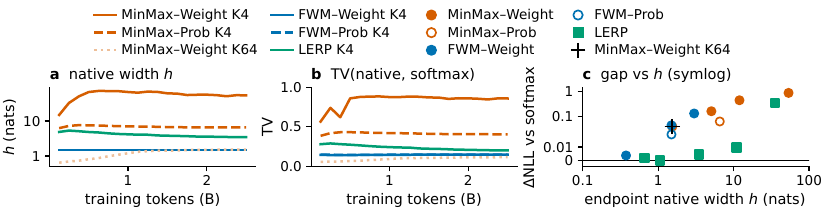}
\vspace{-10pt}
\caption{\textbf{Geometry and resolution at 2.5B.} Geometry statistics are layer means over eight validation blocks on each model's own scores. (a) Native interval width $h$. (b) Total variation between native
and softmax probabilities. (c) Endpoint NLL gap versus native width for all 14 approximate conditions. The cross marks MinMax--Weight $K{=}64$ ($h=1.54$, $\dNLL=0.038$): large score span, moderate interval width. Vertical axis symlog, linear for $|\dNLL|\le0.02$.}
\label{fig:geom}
\end{figure}

\begin{figure}[htb]
\centering
\includegraphics[width=0.72\textwidth]{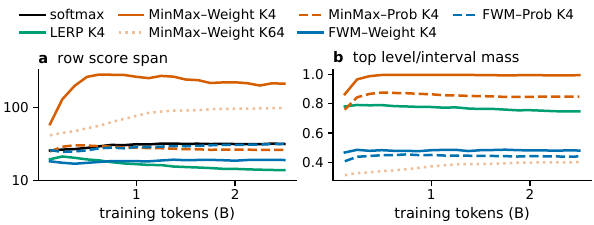}
\caption{\textbf{Layer-mean score geometry along training at 2.5B}, $K{=}4$ conditions, softmax and
MinMax--Weight $K{=}64$ (dotted), on each model's own scores over 8 validation blocks. (a) Row span; (b) softmax mass
on the keys in the operator's top set, which is the top rounding bucket for \CH and the top interval for LERP (these
sets differ, so the panel compares each operator with its own top set). The native interval width $h$ and the per-row TV
of the same six approximate conditions and checkpoints are Figure~\ref{fig:geom}a,b.}
\label{fig:geomfull}
\end{figure}

\begin{figure}[htb]
\centering
\includegraphics[width=\textwidth]{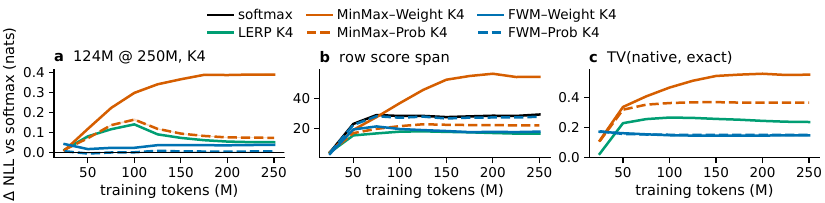}
\caption{\textbf{124M @ 250M, $K{=}4$ conditions.} (a) $\dNLL$ vs softmax with 95\% CI at 25M resolution.
(b) Row span and (c) TV along training. At 50M the MinMax--Weight/MinMax--Prob NLL gap (0.047) is open while
their geometry is nearly identical.}
\label{fig:onset}
\end{figure}

\begin{figure}[htb]
\centering
\includegraphics[width=\textwidth]{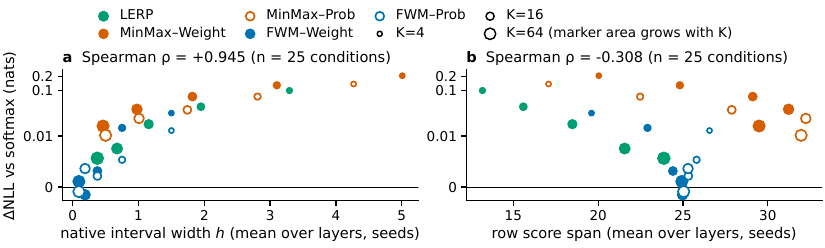}
\caption{\textbf{Five-seed condition means: geometry vs endpoint gap.} (a) Native interval width $h$ and
(b) row score span, each averaged over layers and seeds at the 100M endpoint, against the condition's
mean $\dNLL$ vs softmax (symlog axis). Spearman correlations are across the 25 approximate conditions of a
designed grid: cross-condition associations, not causal estimates.}
\label{fig:widthgap}
\end{figure}

\paragraph{Additional geometry observations.} At 2.5B (Figure~\ref{fig:geomfull}) MinMax--Weight $K{=}4$ peaks at span 287 and $K{=}16$ reaches 191;
the largest logit dispersion reported for Gemma 7B is about 33 \citep{velickovic2024softmax}, so softmax's 25--32 is in
that range while MinMax--Weight's and the detached runs' spans are not. The softmax mass in MinMax--Weight $K{=}4$'s top
rounding bucket reaches 99.5\%: the operator maps almost all mass to one grid value, which is not the model concentrating
probability on one key. \FW is structurally the calibration-side counterpart of the explicit clip that binarized networks
needed \citep{bnn2016}, an analogy not promoted to a shared mechanism. Within-run Spearman between span and the NLL gap
over the 10 checkpoints of the 250M MinMax--Weight run is about 0.9. Each link of the range--resolution feedback
hypothesis has a literature counterpart (\citealp{velickovic2024softmax,eureka2024}); the tail-policy comparison of
Appendix~\ref{app:taildecomp} separates one part of the compound calibration change; freezing the step and a surrogate swap from saved
checkpoints are reported in \S\ref{sec:res-geom} and Appendix~\ref{app:rb-interv}; QK-normalization \citep{vit22b2023}
and a learning-rate sweep \citep{wortsman2024proxies} would test the loop as a whole. Geometric diagnostics moved before the loss in the
detach failure but after it in the full-gradient MinMax--Weight runs (Figure~\ref{fig:onset}), so they are not a universal early
warning; across the five-seed conditions the endpoint gap tracks $h$ and span only as a cross-condition association
(Figure~\ref{fig:widthgap}). All
associations in this appendix are descriptive.

\section{Full endpoint results and downstream evaluation}
\label{app:full}

\begin{table}[htb]
\centering\scriptsize
\caption{All single-seed endpoints (final checkpoint, full validation, 95\% circular block-bootstrap CI over validation blocks). NLL is the mean over 1024-token blocks of the per-token NLL under GPT-2 BPE (nats/token); the corresponding perplexity $\exp(\text{mean NLL})$ is \emph{token-level} and not comparable to word-level WikiText-103 perplexities in the literature.}
\label{tab:endpoints}
\adjustbox{max width=\textwidth}{\begin{tabular}{lrrl}
\toprule
Condition & NLL & $\Delta$ vs softmax & 95\% block CI \\
\midrule
\multicolumn{4}{l}{\emph{124M @ 250M, FineWebEdu-3B, 976 blocks}} \\
softmax & 3.8883 & --- & --- \\
LERP K4 & 3.9401 & +0.0518 & [+0.050, +0.053] \\
MinMax--Weight K4 & 4.2774 & +0.3892 & [+0.385, +0.393] \\
MinMax--Weight K16 & 4.0127 & +0.1244 & [+0.123, +0.126] \\
MinMax--Prob K4 & 3.9618 & +0.0735 & [+0.072, +0.075] \\
MinMax--Prob K16 & 3.9054 & +0.0171 & [+0.016, +0.018] \\
FWM--Weight K4 & 3.9271 & +0.0388 & [+0.038, +0.040] \\
FWM--Weight K16 & 3.8851 & -0.0032 & [-0.004, -0.002] \\
FWM--Prob K4 & 3.8951 & +0.0068 & [+0.006, +0.008] \\
FWM--Prob K16 & 3.8862 & -0.0020 & [-0.003, -0.001] \\
\midrule
\multicolumn{4}{l}{\emph{1B @ 100M, WikiText-103, 243 blocks}} \\
softmax & 4.0904 & --- & --- \\
LERP K4 & 4.1867 & +0.0963 & [+0.092, +0.100] \\
LERP K16 & 4.1322 & +0.0417 & [+0.039, +0.044] \\
LERP K32 & 4.1101 & +0.0197 & [+0.017, +0.022] \\
MinMax--Weight K4 & 4.3107 & +0.2202 & [+0.214, +0.226] \\
MinMax--Weight K16 & 4.1653 & +0.0748 & [+0.072, +0.077] \\
MinMax--Weight K32 & 4.1357 & +0.0453 & [+0.043, +0.048] \\
MinMax--Prob K4 & 4.2368 & +0.1464 & [+0.142, +0.151] \\
FWM--Weight K4 & 4.1512 & +0.0608 & [+0.057, +0.065] \\
FWM--Weight K16 & 4.1090 & +0.0186 & [+0.016, +0.021] \\
FWM--Weight K32 & 4.1067 & +0.0163 & [+0.014, +0.019] \\
FWM--Prob K4 & 4.1288 & +0.0383 & [+0.035, +0.041] \\
\bottomrule
\end{tabular}\hfill
\begin{tabular}{lrrl}
\toprule
Condition & NLL & $\Delta$ vs softmax & 95\% block CI \\
\midrule
\multicolumn{4}{l}{\emph{124M @ 2.5B, FineWebEdu-3B, 976 blocks}} \\
softmax & 3.1928 & --- & --- \\
LERP K1 & 3.5386 & +0.3458 & [+0.340, +0.351] \\
LERP K2 & 3.2025 & +0.0097 & [+0.008, +0.011] \\
LERP K4 & 3.1975 & +0.0047 & [+0.003, +0.006] \\
LERP K16 & 3.1931 & +0.0003 & [-0.001, +0.002] \\
LERP K32 & 3.1947 & +0.0019 & [+0.000, +0.003] \\
MinMax--Weight K4 & 4.0831 & +0.8903 & [+0.881, +0.899] \\
MinMax--Weight K16 & 3.6397 & +0.4469 & [+0.442, +0.452] \\
MinMax--Weight K32 & 3.3527 & +0.1600 & [+0.158, +0.162] \\
MinMax--Weight K64 & 3.2311 & +0.0383 & [+0.037, +0.040] \\
MinMax--Prob K4 & 3.2542 & +0.0614 & [+0.060, +0.063] \\
FWM--Weight K2 & 3.3226 & +0.1298 & [+0.128, +0.132] \\
FWM--Weight K4 & 3.2355 & +0.0427 & [+0.041, +0.045] \\
FWM--Weight K16 & 3.1967 & +0.0039 & [+0.002, +0.005] \\
FWM--Prob K4 & 3.2121 & +0.0193 & [+0.018, +0.021] \\
\bottomrule
\end{tabular}}
\end{table}

\begin{table}[htb]
\centering\scriptsize
\caption{Five-seed endpoints (124M @ 100M, WikiText-103; mean NLL in nats/token, GPT-2 BPE; the token-level perplexity $\exp(\text{mean NLL})$ is not comparable to word-level perplexities). $\Delta$ is paired by seed; high-$K$ \FW cells are within seed noise of softmax and are not strictly ordered.}
\label{tab:seedend}
\adjustbox{max width=\textwidth}{\begin{tabular}{lrrrlrr}
\toprule
Condition & mean NLL & seed SD & $\Delta$ vs softmax & 95\% t-CI & seed SD of $\Delta$ & paired eval SE \\
\midrule
softmax & 4.1457 & 0.0062 & --- & --- & --- & --- \\
LERP K4 & 4.2441 & 0.0080 & +0.0985 & [+0.091, +0.106] & 0.0062 & 0.0019 \\
LERP K8 & 4.1894 & 0.0098 & +0.0437 & [+0.036, +0.052] & 0.0064 & 0.0013 \\
LERP K16 & 4.1639 & 0.0074 & +0.0182 & [+0.015, +0.022] & 0.0029 & 0.0013 \\
LERP K32 & 4.1532 & 0.0089 & +0.0076 & [-0.000, +0.015] & 0.0063 & 0.0012 \\
LERP K64 & 4.1513 & 0.0098 & +0.0057 & [-0.001, +0.013] & 0.0058 & 0.0011 \\
MinMax--Weight K4 & 4.3532 & 0.0097 & +0.2076 & [+0.191, +0.224] & 0.0131 & 0.0028 \\
\bottomrule
\end{tabular}}
\end{table}

\subsection{Downstream evaluation of the 2.5B suite}

\paragraph{Pre-declared condition groups.} Before any downstream run the fifteen 2.5B endpoints were assigned to three
groups by their 2.5B $\dNLL$ vs softmax, with the declared expectation in quotes: \emph{large-gap}, $\dNLL$ 0.13--0.89,
MinMax--Weight K4 (+0.890), MinMax--Weight K16 (+0.447), LERP K1 (+0.346), MinMax--Weight K32 (+0.160), \FW--Weight K2
(+0.130), ``should be clearly detectable''; \emph{intermediate-gap}, 0.02--0.06, MinMax--Prob K4 (+0.061), \FW--Weight K4
(+0.043), MinMax--Weight K64 (+0.038), \FW--Prob K4 (+0.019), ``possibly detectable, task dependent'';
\emph{near-baseline}, $<0.01$, LERP K2 (+0.010), LERP K4 (+0.005), \FW--Weight K16 (+0.004), LERP K32 (+0.002), LERP
K16 (+0.0003), ``expected not detectable''. The lists are the pre-declaration; the ranges are descriptive and three members sit at their rounded edges.
They fix the denominators 35, 28 and 35 of \S\ref{sec:res-downstream}.

\begin{figure}[htb]
\centering
\includegraphics[width=0.62\textwidth]{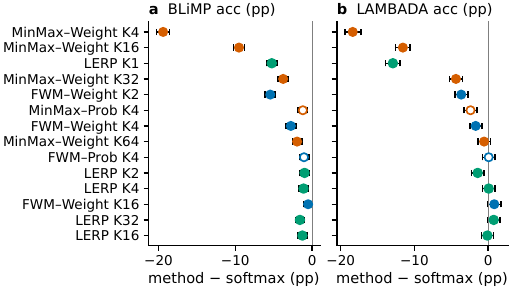}
\caption{\textbf{Downstream accuracy of the fourteen approximate 2.5B endpoints}, method minus softmax in
percentage points with 95\% paired bootstrap CIs: (a) BLiMP macro accuracy, (b) LAMBADA accuracy. Rows are
ordered by 2.5B $\dNLL$; the $\dNLL$ values on the training domain and the three held-out corpora are in
Tables~\ref{tab:endpoints} and~\ref{tab:subst}.}
\label{fig:downstream}
\end{figure}

\paragraph{Held-out corpora (layer 1).} On WikiText-103, PTB and C4 the Spearman correlation with the in-domain ordering
is 0.996, 0.952 and 0.996; MinMax--Weight $K{=}4$ goes from $+0.89$ in domain to $+1.22$ on WikiText-103 and $+1.12$ on
PTB, unchanged on C4. Table~\ref{tab:nearci} gives the paired 95\% intervals for the five near-baseline conditions on the
three corpora (counts under P2 below): the differences are below 0.06 nats and of inconsistent sign (the four LERP
conditions \emph{below} softmax on PTB, \FW--Weight $K{=}16$ above it on PTB and C4), small but detectable held-out
differences in both directions, neither ``no detectable difference'' nor equivalence. Table~\ref{tab:subst} separates the native-forward and softmax-forward
evaluations of the same weights.

\begin{table}[htb]
\centering\scriptsize
\caption{Layer 1, near-baseline group: paired $\dNLL$ vs softmax on the held-out corpora (native forward) with 95\% CIs (circular moving-block bootstrap, block 16, for WikiText-103 and PTB; document-cluster bootstrap for C4; 4000 resamples), BH $q$ within each corpus's 14-comparison family, and the two independent labels of Appendix~\ref{app:protocol}.}
\label{tab:nearci}
\adjustbox{max width=\textwidth}{\begin{tabular}{llrlrcc}
\toprule
Condition & Corpus & $\dNLL$ & 95\% CI & BH $q$ & CI excl.\ 0 & CI inside $\pm0.01$ \\
\midrule
LERP K2 & WikiText-103 test & +0.0052 & [-0.0005, +0.0106] & 0.087 & no & no \\
LERP K2 & PTB test & -0.0162 & [-0.0235, -0.0090] & <0.001 & yes & no \\
LERP K2 & C4 validation & +0.0131 & [+0.0096, +0.0165] & <0.001 & yes & no \\
LERP K4 & WikiText-103 test & +0.0023 & [-0.0020, +0.0068] & 0.318 & no & yes \\
LERP K4 & PTB test & -0.0177 & [-0.0238, -0.0123] & <0.001 & yes & no \\
LERP K4 & C4 validation & +0.0045 & [+0.0012, +0.0078] & 0.007 & yes & yes \\
LERP K16 & WikiText-103 test & -0.0084 & [-0.0130, -0.0036] & <0.001 & yes & no \\
LERP K16 & PTB test & -0.0578 & [-0.0639, -0.0519] & <0.001 & yes & no \\
LERP K16 & C4 validation & -0.0007 & [-0.0036, +0.0022] & 0.656 & no & yes \\
LERP K32 & WikiText-103 test & -0.0059 & [-0.0099, -0.0021] & 0.004 & yes & yes \\
LERP K32 & PTB test & -0.0393 & [-0.0460, -0.0330] & <0.001 & yes & no \\
LERP K32 & C4 validation & +0.0011 & [-0.0015, +0.0038] & 0.444 & no & yes \\
FWM--Weight K16 & WikiText-103 test & -0.0021 & [-0.0070, +0.0026] & 0.391 & no & yes \\
FWM--Weight K16 & PTB test & +0.0227 & [+0.0167, +0.0286] & <0.001 & yes & no \\
FWM--Weight K16 & C4 validation & +0.0050 & [+0.0018, +0.0082] & 0.002 & yes & yes \\
\bottomrule
\end{tabular}}
\end{table}

\begin{table}[htb]
\centering\scriptsize
\caption{Layer 1: $\dNLL$ vs softmax on three held-out corpora with the model's native forward, with softmax substituted at evaluation on the same weights, and their difference (softmax-fwd $-$ native). The softmax reference in each column is the softmax model evaluated in the same mode. In 37 of the 42 approximate-condition cells the substituted gap exceeds the native one.}
\label{tab:subst}
\adjustbox{max width=\textwidth}{\begin{tabular}{lrrrrrrrrr}
\toprule
& \multicolumn{3}{c}{WikiText-103 test} & \multicolumn{3}{c}{PTB test} & \multicolumn{3}{c}{C4 validation} \\
\cmidrule(lr){2-4}\cmidrule(lr){5-7}\cmidrule(lr){8-10}
Condition & native & softmax-fwd & diff & native & softmax-fwd & diff & native & softmax-fwd & diff \\
\midrule
MinMax--Weight K4 & +1.215 & +3.549 & +2.334 & +1.122 & +3.309 & +2.187 & +0.865 & +2.942 & +2.077 \\
MinMax--Weight K16 & +0.588 & +1.232 & +0.644 & +0.526 & +0.905 & +0.380 & +0.435 & +0.971 & +0.536 \\
LERP K1 & +0.652 & +4.947 & +4.295 & +0.702 & +4.735 & +4.034 & +0.390 & +4.637 & +4.247 \\
MinMax--Weight K32 & +0.209 & +0.242 & +0.033 & +0.176 & +0.202 & +0.026 & +0.156 & +0.210 & +0.054 \\
FWM--Weight K2 & +0.180 & +0.288 & +0.108 & +0.135 & +0.236 & +0.101 & +0.128 & +0.245 & +0.117 \\
MinMax--Prob K4 & +0.097 & +0.208 & +0.111 & +0.091 & +0.173 & +0.082 & +0.067 & +0.143 & +0.077 \\
FWM--Weight K4 & +0.066 & +0.084 & +0.019 & +0.049 & +0.057 & +0.008 & +0.044 & +0.046 & +0.002 \\
MinMax--Weight K64 & +0.045 & +0.057 & +0.013 & +0.068 & +0.063 & -0.005 & +0.038 & +0.048 & +0.010 \\
FWM--Prob K4 & +0.027 & +0.025 & -0.001 & -0.020 & -0.026 & -0.006 & +0.021 & +0.015 & -0.007 \\
LERP K2 & +0.005 & +1.814 & +1.809 & -0.016 & +1.670 & +1.686 & +0.013 & +1.685 & +1.672 \\
LERP K4 & +0.002 & +0.174 & +0.172 & -0.018 & +0.169 & +0.187 & +0.004 & +0.168 & +0.163 \\
FWM--Weight K16 & -0.002 & +0.002 & +0.004 & +0.023 & +0.030 & +0.007 & +0.005 & +0.004 & -0.001 \\
LERP K32 & -0.006 & -0.005 & +0.001 & -0.039 & -0.036 & +0.003 & +0.001 & +0.003 & +0.002 \\
LERP K16 & -0.008 & -0.002 & +0.007 & -0.058 & -0.052 & +0.006 & -0.001 & +0.006 & +0.007 \\
\bottomrule
\end{tabular}}
\end{table}

\paragraph{Benchmarks (layer 3).} The seven configurations are six benchmarks, SciQ scored with and without its
support passage (Table~\ref{tab:stagec}, symbols as in Figure~\ref{fig:heat}; BLiMP and LAMBADA with CIs in
Figure~\ref{fig:downstream}). BLiMP is the macro average over 67
paradigms of 200 minimal pairs each, resampled by item. The intermediate-gap group separates from softmax only on BLiMP and LAMBADA (6 of 28 comparisons); \FW--Weight
$K{=}16$ does not separate on BLiMP ($-0.5$ pp, $q=0.10$); the mean near-baseline difference ($+0.3$ pp) is descriptive
only. No condition loses more of the SciQ support-passage benefit than softmax after correction.

\begin{table}[htb]
\centering\scriptsize
\caption{Layer 3, all 14 conditions: accuracy minus softmax in percentage points, paired per item and computed from the
per-item differences (not from rounded absolutes); the header row gives softmax's absolute accuracy. BLiMP is the
macro-average over 67 paradigms (200 pairs each); HellaSwag/PIQA/ARC-Easy use length-normalized accuracy (the harness metric \texttt{acc\_norm}); LAMBADA and
SciQ use accuracy (\texttt{acc}); SciQ-ns is without the support passage. Rows are ordered by 2.5B $\dNLL$ with rules between the large-gap,
intermediate-gap and near-baseline groups. $^{*}$: BH $q<0.05$ within the configuration$\times$metric family of 14
comparisons (the symbol of Figure~\ref{fig:heat}); $^{\circ}$: 95\% paired CI excludes zero but $q\ge0.05$.}
\label{tab:stagec}
\begin{tabular}{lrrrrrrr}
\toprule
Condition & BLiMP & HellaSwag & PIQA & ARC-E & LAMBADA & SciQ & SciQ-ns \\
\midrule
softmax (absolute, \%) & 79.5 & 30.0 & 59.7 & 45.2 & 21.6 & 73.1 & 51.2 \\ \midrule
MinMax--Weight K4 & -19.4$^{*}$ & -3.9$^{*}$ & -4.1$^{*}$ & -9.6$^{*}$ & -18.4$^{*}$ & -19.3$^{*}$ & -17.6$^{*}$ \\
MinMax--Weight K16 & -9.5$^{*}$ & -2.7$^{*}$ & -2.2$^{\circ}$ & -5.7$^{*}$ & -11.6$^{*}$ & -9.3$^{*}$ & -8.7$^{*}$ \\
LERP K1 & -5.2$^{*}$ & -1.9$^{*}$ & -0.2 & -3.9$^{*}$ & -12.9$^{*}$ & -8.5$^{*}$ & -5.2$^{*}$ \\
MinMax--Weight K32 & -3.8$^{*}$ & -1.6$^{*}$ & -0.4 & -3.5$^{*}$ & -4.4$^{*}$ & -2.2 & -1.4 \\
FWM--Weight K2 & -5.4$^{*}$ & -1.1$^{*}$ & -1.0 & -2.6$^{*}$ & -3.6$^{*}$ & -1.2 & -1.4 \\ \midrule
MinMax--Prob K4 & -1.2$^{*}$ & -0.6$^{\circ}$ & +1.3 & -1.2 & -2.4$^{*}$ & -0.5 & +0.9 \\
FWM--Weight K4 & -2.8$^{*}$ & -0.5 & +1.0 & -0.2 & -1.7$^{*}$ & +1.6 & +0.9 \\
MinMax--Weight K64 & -1.9$^{*}$ & -0.3 & +0.4 & -0.7 & -0.5 & +1.1 & +0.6 \\
FWM--Prob K4 & -1.1$^{*}$ & -0.4 & +0.8 & -0.8 & +0.1 & -0.7 & +2.3 \\ \midrule
LERP K2 & -1.0$^{*}$ & -0.0 & +0.2 & -2.4$^{*}$ & -1.4$^{*}$ & +2.0 & +1.5 \\
LERP K4 & -1.1$^{*}$ & +0.4 & -0.2 & -0.6 & +0.1 & +1.3 & +1.5 \\
FWM--Weight K16 & -0.5 & -0.1 & +1.7$^{\circ}$ & -0.3 & +0.8 & +3.9$^{*}$ & +3.3$^{*}$ \\
LERP K32 & -1.6$^{*}$ & +0.2 & +0.5 & -0.6 & +0.8 & +1.0 & +0.4 \\
LERP K16 & -1.3$^{*}$ & +0.2 & +0.8 & +0.3 & -0.1 & +0.4 & +0.5 \\
\bottomrule
\end{tabular}
\end{table}

\paragraph{Induction and retrieval probes (layer 2, probes A and B).} Table~\ref{tab:probeab} reports every evaluated
condition on probe A (400 families $\times$ 5 distances $=$ 2,000 induction samples, each with a matched control) and
probe B (2,400 samples), native forward, against the softmax reference: per-token target NLL, the induction gain
(control minus induction target NLL, probe A) and top-1 accuracy. The NLL columns carry paired bootstrap CIs. The five
distances of a probe-A family share one base sequence and one $(A,B)$ pair, so the probe-A intervals resample the 400
families, keeping every distance, the induction/control pairing and the method pairing inside a family; probe-B samples
are independent and are resampled directly. The reference model itself barely solves either probe (top-1
0.05\% on A, 0.7\% on B; its induction gain falls from 1.22 nats at distance 64 to 0.21 at 896, 17\%), so differences
between conditions are measured at a baseline with little of the ability the probes target: P4 (degradation growing
with distance) could not be tested, and the large, detectable target-NLL differences of both signs on A and B are
reported as probe outcomes rather than as evidence; the largest, MinMax--Prob $K{=}4$'s $-1.28$-nat target NLL and
$+1.47$-nat induction gain on probe A (top-1 from 0.05\% to 4\%), is not interpreted as long-range copying. This is a
limit of the reference model at 124M, not a null result and not equivalence between conditions.

\begin{table}[htb]
\centering\scriptsize
\caption{Probes A and B, all 14 conditions plus the softmax reference (124M @ 2.5B endpoints, native forward). Softmax
row: absolute values; other rows: difference from softmax. The target-NLL and induction-gain columns carry 95\%
paired-bootstrap CIs (4000 resamples; probe A resampled by family over the 400 families, probe B by sample); the top-1
columns are point estimates only, in accuracy units (fraction). Induction gain: target NLL without minus with the earlier
occurrence (probe A).}
\label{tab:probeab}
\adjustbox{max width=\textwidth}{\begin{tabular}{lrrrrr}
\toprule
 & \multicolumn{3}{c}{Probe A: induction ($n{=}2{,}000$ induction samples)} & \multicolumn{2}{c}{Probe B: retrieval ($n{=}2{,}400$)} \\
\cmidrule(lr){2-4}\cmidrule(lr){5-6}
Condition & target NLL & induction gain & top-1 & target NLL & top-1 \\
\midrule
softmax (absolute) & 11.160 & 0.865 & 0.0005 & 10.255 & 0.0071 \\
\midrule
MinMax--Weight K4 & +0.583 [+0.377, +0.775] & -0.515 [-0.589, -0.440] & -0.0005 & +1.708 [+1.622, +1.799] & -0.0071 \\
MinMax--Weight K16 & +0.321 [+0.134, +0.496] & -0.087 [-0.174, +0.003] & +0.0060 & +1.298 [+1.217, +1.383] & -0.0067 \\
LERP K1 & +1.212 [+1.019, +1.399] & -0.753 [-0.825, -0.680] & -0.0005 & +1.508 [+1.432, +1.587] & -0.0067 \\
MinMax--Weight K32 & +0.214 [+0.047, +0.378] & -0.170 [-0.237, -0.103] & +0.0005 & +1.716 [+1.638, +1.798] & -0.0058 \\
FWM--Weight K2 & -0.059 [-0.245, +0.119] & +0.168 [+0.084, +0.259] & +0.0045 & +0.811 [+0.738, +0.883] & -0.0050 \\
MinMax--Prob K4 & -1.281 [-1.496, -1.067] & +1.469 [+1.300, +1.638] & +0.0400 & +0.658 [+0.579, +0.736] & -0.0004 \\
FWM--Weight K4 & -0.070 [-0.232, +0.098] & +0.116 [+0.040, +0.197] & +0.0020 & +0.816 [+0.743, +0.892] & -0.0058 \\
MinMax--Weight K64 & +1.204 [+1.007, +1.394] & -0.601 [-0.675, -0.526] & -0.0005 & +0.628 [+0.554, +0.697] & -0.0021 \\
FWM--Prob K4 & -0.559 [-0.725, -0.395] & +0.723 [+0.627, +0.822] & +0.0100 & +0.772 [+0.706, +0.840] & -0.0054 \\
LERP K2 & +0.234 [+0.052, +0.405] & -0.068 [-0.141, +0.008] & +0.0005 & +0.541 [+0.464, +0.621] & -0.0050 \\
LERP K4 & -0.219 [-0.390, -0.045] & +0.093 [+0.008, +0.184] & +0.0010 & +0.771 [+0.696, +0.848] & -0.0050 \\
FWM--Weight K16 & +0.737 [+0.579, +0.897] & -0.351 [-0.406, -0.292] & +0.0000 & -0.039 [-0.114, +0.035] & +0.0046 \\
LERP K32 & -0.170 [-0.330, -0.019] & +0.278 [+0.198, +0.360] & +0.0010 & +0.725 [+0.655, +0.794] & -0.0037 \\
LERP K16 & -0.310 [-0.485, -0.132] & +0.471 [+0.365, +0.581] & +0.0025 & +0.473 [+0.404, +0.539] & -0.0029 \\
\bottomrule
\end{tabular}}
\end{table}

\paragraph{Copying probe (layer 2, probe C).} Table~\ref{tab:probec} gives the per-token target-NLL difference from
softmax by copied length with softmax's absolute target NLL. The slope against $\log_2L$ is steeper under Prob-STE than
Weight-STE at matched calibration and $K{=}4$ (MinMax $0.94$ vs $0.11$, \FW $0.88$ vs $0.23$); under
MinMax Prob-STE is lower at every length, under \FW lower at $L{=}32,64$ and higher at $L{=}128$. The softmax-forward
substitution was run on the three probes for six conditions, not for the benchmark battery.

\begin{table}[htb]
\centering\scriptsize
\caption{Layer 2, probe C (verbatim sequence copying): softmax's absolute per-token target NLL, then each condition's per-token target-NLL difference from softmax by copied length $L$ and the slope of that difference against $\log_2L$ [95\% CI].}
\label{tab:probec}
\adjustbox{max width=\textwidth}{\begin{tabular}{lrrrl}
\toprule
Condition & $\Delta$ at L=32 & L=64 & L=128 & slope vs $\log_2 L$ [95\% CI] \\
\midrule
softmax, absolute target NLL & 1.723 & 1.576 & 1.776 & --- \\ \midrule
MinMax--Weight K4 & +7.13 & +7.06 & +7.34 & +0.106 [+0.064, +0.146] \\
MinMax--Weight K16 & +1.91 & +1.48 & +1.24 & -0.337 [-0.375, -0.300] \\
LERP K1 & +7.96 & +9.24 & +9.89 & +0.963 [+0.929, +0.997] \\
MinMax--Weight K32 & +0.50 & +0.40 & +0.26 & -0.119 [-0.145, -0.095] \\
FWM--Weight K2 & +3.21 & +3.91 & +4.67 & +0.727 [+0.684, +0.770] \\
MinMax--Prob K4 & +0.82 & +1.41 & +2.69 & +0.939 [+0.897, +0.980] \\
FWM--Weight K4 & +0.54 & +0.70 & +1.01 & +0.234 [+0.200, +0.268] \\
MinMax--Weight K64 & +0.24 & +0.42 & +0.36 & +0.057 [+0.028, +0.085] \\
FWM--Prob K4 & +0.18 & +0.48 & +1.94 & +0.877 [+0.847, +0.907] \\
LERP K2 & +0.37 & +0.30 & +0.70 & +0.163 [+0.134, +0.190] \\
LERP K4 & +0.07 & +0.05 & +0.47 & +0.200 [+0.174, +0.226] \\
FWM--Weight K16 & -0.00 & -0.05 & -0.07 & -0.033 [-0.056, -0.011] \\
LERP K32 & -0.08 & +0.05 & +0.27 & +0.178 [+0.154, +0.201] \\
LERP K16 & -0.15 & -0.01 & +0.22 & +0.185 [+0.161, +0.208] \\
\bottomrule
\end{tabular}}
\end{table}

\paragraph{Pre-declared predictions, outcome.} P1 (large-gap group detectable on layers 2 and 3): supported. P2
(near-baseline group not detectable): not supported on held-out LM NLL, where 10 of the 15 paired intervals exclude zero
(Table~\ref{tab:nearci}) although 7 lie inside the $\pm0.01$-nat margin, nor uniformly on the benchmarks, where 8 of the
group's 35 primary-accuracy comparisons reach $q<0.05$ (six negative, on BLiMP, ARC-Easy and LAMBADA; two positive, on
the two SciQ configurations; \S\ref{sec:res-downstream}). P3 (probe degradation exceeds LM degradation): supported on
probe C only (13 of 14 conditions). P4 (degradation grows with retrieval distance): not testable, because the reference
model's own induction gain decays to 17\% by distance 896. P5 (added after layer 1; softmax-forward reversal larger on probes than on LM):
holds in 5 of the 18 cells of the substitution on probes A/B/C for six conditions, so it fails as measured, and it is not
an effect-size ratio, because LM and probe metrics average over different token distributions, positions, tasks and
sample sizes. For the downstream metrics the
realized shrinkage from pairing was 0.27--1.32 (median 0.64), weaker than for corpus NLL ($s\approx0.20$), which is why
every interval is built from realized per-item differences.

\section{The detached-gradient record and its replication}
\label{app:detach}

\begin{figure}[htb]
\centering
\includegraphics[width=0.72\textwidth]{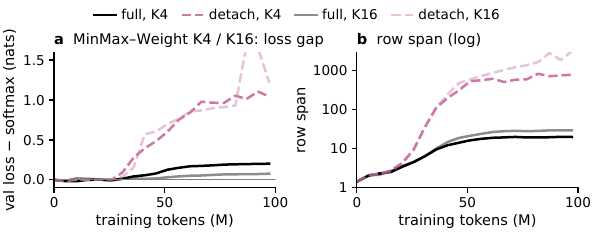}
\caption{\textbf{Detached vs full calibration gradient, \CH, seed 1337 (124M @ 100M, WikiText-103).}
MinMax--Weight $K{=}4$ (solid colour) and $K{=}16$ (faded), detached and full backward: (a) logged validation
loss minus the seed-1337 softmax run and (b) logged layer-mean row span along training. Endpoints with CIs:
Table~\ref{tab:detach}.}
\label{fig:detachextra}
\end{figure}

The first implementation computed the row maximum and the row minimum with \texttt{.detach()} applied to each; the
later implementation makes the detached and the full backward two modes of one forward code path, so forward values are
the same. The historical detached runs are 100M-token runs on seed 1337 (the softmax run, which is also the
seed-1337 softmax run of suite D; \CH $K{=}2/4/8/16$, LERP $K{=}4/32$) and six 5M-token confirmatory \CH runs ($K{=}2/4/8$ on seeds
1338 and 1339) with 5M softmax references on seeds 1337, 1338 and 1339, all WikiText-103, 124M, with the optimizer
settings of the later 100M runs. They enter no table of suites A--D. Table~\ref{tab:detach}
gives the standardized endpoints, including the full-gradient endpoints against softmax, and
Figure~\ref{fig:detachextra} the MinMax--Weight trajectories.

\begin{table}[htb]
\centering\scriptsize
\caption{Detached vs full calibration gradient at 100M tokens, paired on 243 WikiText-103 validation blocks (95\% block-bootstrap CIs). Softmax (seed 1337) NLL 4.1499. The full-gradient runs are the seed-1337 members of suite D. No full-gradient $K{=}2$ run exists.}
\label{tab:detach}
\adjustbox{max width=\textwidth}{\begin{tabular}{lllll}
\toprule
Condition (seed 1337, 100M) & historical backward & historical $-$ softmax & full-gradient $-$ softmax & historical $-$ full \\
\midrule
MinMax--Weight K2 & detach & +2.467 [+2.410, +2.525] & --- & --- \\
MinMax--Weight K4 & detach & +1.023 [+1.000, +1.045] & +0.1979 [+0.193, +0.203] & +0.825 [+0.806, +0.845] \\
MinMax--Weight K8 & detach & +3.200 [+3.141, +3.258] & +0.1331 [+0.130, +0.137] & +3.066 [+3.009, +3.125] \\
MinMax--Weight K16 & detach & +1.157 [+1.131, +1.185] & +0.0778 [+0.076, +0.080] & +1.079 [+1.053, +1.107] \\
LERP K4 & detach & +0.748 [+0.733, +0.764] & +0.0944 [+0.091, +0.098] & +0.654 [+0.639, +0.668] \\
LERP K32 & detach & +0.736 [+0.719, +0.752] & -0.0008 [-0.004, +0.002] & +0.737 [+0.721, +0.752] \\
LERP K32 & detach + zero-sum proj. & +0.996 [+0.977, +1.013] & -0.0008 [-0.004, +0.002] & +0.997 [+0.979, +1.015] \\
\bottomrule
\end{tabular}}
\end{table}

\begin{table}[htb]
\centering\scriptsize
\caption{Gradient-consistency diagnostics of the detached and the full backward.}
\label{tab:graddiag}
\begin{tabular}{@{}>{\raggedright\arraybackslash}p{3.2cm}>{\raggedright\arraybackslash}p{4.0cm}>{\raggedright\arraybackslash}p{5.8cm}@{}}
\toprule
Diagnostic & Question & Result \\
\midrule
forward shift invariance & $F(S+c\mathbf1)=F(S)$, softmax, LERP, \CH, $K{=}2\ldots32$ & Pass for all. \\
gradient shift residual & $\rho=|\sum_jg_{ij}|/\sum_j|g_{ij}|$ per row & softmax $4.7{\times}10^{-8}$; LERP full $2.2{\times}10^{-7}$; LERP detach median 0.026--0.410, max near 1. \\
all-ones directional finite difference (fp64) & autograd vs derivative along $s+c\mathbf1$ & Detach $K{=}32$: autograd $-2.0{\times}10^{-3}$ vs true 0. Full: $\approx0$. \\
interior finite difference & local LERP slopes away from extrema & softmax, detach and full all match. \\
zero-sum projection & removes exactly the common-mode gradient? & Forward bitwise equal to detach; projected $\rho_{\max}$ $1.7{\times}10^{-16}$; $\max|g_{\mathrm{proj}}-g_{\mathrm{full}}|=1.07{\times}10^{-3}$. \\
\bottomrule
\end{tabular}
\end{table}

\begin{table}[htb]
\centering\scriptsize
\caption{Final-checkpoint geometry (layer mean, 8 validation blocks), detached vs full.}
\begin{tabular}{llrrrrrrr}
\toprule
Condition & Backward & Span & $h$ & $\|q\|$ & Entropy & Top-level mass & Eff.\ levels & TV \\
\midrule
\CH K2 & detach & 226.2 & 113.1 & 58.0 & 3.97 & 0.679 & 1.81 & 0.439 \\
\CH K4 & full & 19.8 & 4.94 & 10.9 & 3.77 & 0.637 & 2.31 & 0.355 \\
\CH K4 & detach & 903.8 & 225.9 & 88.4 & 0.66 & 0.990 & 1.06 & 0.873 \\
\CH K8 & full & 25.5 & 3.18 & 11.3 & 3.86 & 0.506 & 3.57 & 0.259 \\
\CH K8 & detach & 1544.7 & 193.1 & 72.8 & 1.03 & 0.927 & 1.47 & 0.702 \\
\CH K16 & full & 29.6 & 1.85 & 12.2 & 3.78 & 0.386 & 6.70 & 0.160 \\
\CH K16 & detach & 5833.6 & 364.6 & 169.6 & 0.47 & 0.973 & 1.16 & 0.710 \\
LERP K4 & full & 13.1 & 3.28 & 11.4 & 3.53 & 0.685 & 2.24 & 0.190 \\
LERP K4 & detach & 69.0 & 17.3 & 30.6 & 1.87 & 0.903 & 1.44 & 0.513 \\
LERP K32 & full & 21.9 & 0.69 & 12.0 & 3.72 & 0.334 & 13.15 & 0.017 \\
LERP K32 & detach & 1192.6 & 37.3 & 130.1 & 1.12 & 0.825 & 2.35 & 0.281 \\
LERP K32 & detach + proj. & 1632.2 & 51.0 & 78.4 & 0.22 & 0.979 & 1.17 & 0.278 \\
softmax & native & 25.1 & --- & 12.2 & 3.79 & --- & --- & --- \\
\bottomrule
\end{tabular}
\end{table}

\paragraph{Relation to suites A--D.} The detached-gradient runs preceded suites A--D (5M-token sweeps,
Table~\ref{tab:early5m}, then 100M-token runs with delayed divergence). The diagnostics of Table~\ref{tab:graddiag}
identified the detached extrema as an incomplete Jacobian; the full backward was then added as a second mode with the
forward unchanged, the full-gradient runs trained from scratch were stable and became the seed-1337 members of suite D,
and the zero-sum projection was run afterwards. Every $K$-interval run of suites A--D uses the full calibration
gradient.

\paragraph{5M confirmatory endpoints.} Table~\ref{tab:early5m} reports the six 5M-token \CH runs with the detached
backward against the same-seed 5M softmax reference (paired per block, same block bootstrap as the 100M endpoints). All
six gaps are below $0.003$ nats: at this horizon the endpoints had not yet exposed the later failure, which is not
equivalence (five of six intervals exclude zero; cf.\ the short sweeps of \citealp{fp8dpa2025}), and these seeds are not
the seed-1337 trajectory; the delayed onset is carried by the seed-1337 100M runs and its cross-seed replication by the
LERP $K{=}32$ seed-7/42 runs.

\begin{table}[htb]
\centering\scriptsize
\caption{Historical 5M-token confirmatory runs, \CH $K$ with the detached backward, seeds 1338 and 1339 (124M,
WikiText-103), against the same-seed 5M softmax reference: endpoint NLL over 243 validation blocks at 5.0M tokens, paired
per block, 95\% circular block-bootstrap CI (evaluation sampling only).}
\label{tab:early5m}
\adjustbox{max width=\textwidth}{\setlength{\tabcolsep}{3pt}\begin{tabular}{llrrl}
\toprule
Seed & $K$ & NLL & softmax & $\Delta$ [95\% CI] \\
\midrule
1338 & K2 & 6.9084 & 6.9055 & +0.0029 [+0.0024, +0.0034] \\
1338 & K4 & 6.9082 & 6.9055 & +0.0027 [+0.0025, +0.0029] \\
1338 & K8 & 6.9065 & 6.9055 & +0.0010 [+0.0009, +0.0011] \\
1339 & K2 & 6.9263 & 6.9236 & +0.0027 [+0.0020, +0.0034] \\
1339 & K4 & 6.9242 & 6.9236 & +0.0006 [+0.0005, +0.0008] \\
1339 & K8 & 6.9236 & 6.9236 & +0.0000 [-0.0003, +0.0004] \\
\bottomrule
\end{tabular}}
\end{table}

\begin{table}[htb]
\centering\scriptsize
\caption{Detach replication, LERP $K{=}32$ at 124M @ 100M (WikiText-103): detach minus the matched same-seed full-gradient run at five checkpoints, and detach minus softmax at 100M; paired per block, 95\% circular block-bootstrap CIs (evaluation sampling only).}
\label{tab:detachrep}
\adjustbox{max width=\textwidth}{\setlength{\tabcolsep}{3pt}\begin{tabular}{lll}
\toprule
Ckpt & seed 7 & seed 42 \\
\midrule
20M & -0.002 [-0.006, +0.001] & -0.015 [-0.018, -0.013] \\
40M & +0.118 [+0.113, +0.123] & +0.074 [+0.069, +0.079] \\
60M & +0.672 [+0.653, +0.692] & +0.587 [+0.571, +0.604] \\
80M & +0.714 [+0.697, +0.731] & +0.781 [+0.764, +0.798] \\
100M & +0.802 [+0.785, +0.819] & +0.780 [+0.760, +0.799] \\
100M, detach $-$ softmax & +0.807 [+0.789, +0.823] & +0.788 [+0.768, +0.807] \\
\bottomrule
\end{tabular}}
\end{table}

\begin{figure}[htb]
\centering
\includegraphics[width=\textwidth]{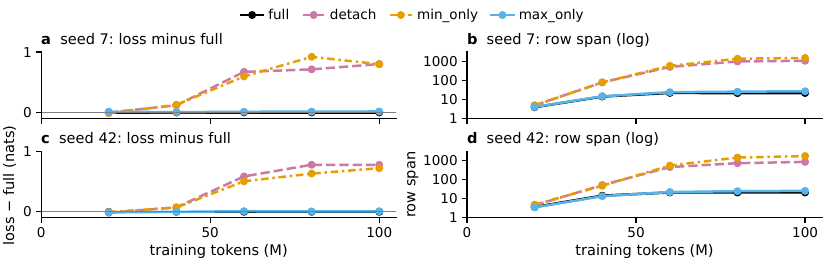}
\caption{\textbf{Single-channel ablation trajectories}, LERP $K{=}32$, seeds 7 (a, b) and 42 (c, d): new single-channel runs
paired with the existing same-seed full and detach controls on one GPU (the seed-1337 runs of Figure~\ref{fig:detach} were trained with an earlier version of
the code). (a, c) Validation loss minus the same-seed full-gradient run at the five evaluated checkpoints;
(b, d) layer-mean row span (log). The baseline is the same-seed full-gradient run, not softmax; the detach $-$ full
values at these checkpoints are tabulated in Table~\ref{tab:detachrep}.}
\label{fig:e1}
\end{figure}

\paragraph{Replication on two further seeds.} The headline condition, LERP $K{=}32$ with the detached backward,
was retrained at 124M @ 100M on WikiText-103 with seeds 7 and 42 (same code, data, schedule
and protocol as the full-gradient five-seed members; detach runs with gradient checkpointing, their full controls
without, step-0 check below), each paired against the existing same-seed full-gradient and softmax runs and evaluated
through its own frozen code (243 blocks, batch 1). Table~\ref{tab:detachrep} gives the paired contrasts at the five
evaluated checkpoints (circular block bootstrap, block 16, 4000 resamples); the detach and full trajectories of both seeds
appear, against the full-gradient baseline, in Figure~\ref{fig:e1}. Both seeds reproduce the endpoint (Table~\ref{tab:detachrep}) and the shape: at 20M no
degradation has appeared (seed 7 $-0.002$ [$-0.006$, $+0.001$]; seed 42 slightly \emph{below} full, $-0.015$ [$-0.018$,
$-0.013$], an interval that excludes zero), positive degradation is present by 40M ($+0.118$ and $+0.074$) and most of the
gap is open by 60M; the final logged gradient norms are 2.4 and 2.3 against about 0.5 for the full-gradient runs. The
seed-1337 value ($+0.737$) was obtained with an earlier version of the code and is consistent with the two new seeds;
the three seeds are not pooled. The other detached conditions of the record remain single-seed.

\paragraph{Single-channel ablation (two seeds).} In the maximum-only and minimum-only modes (max\_only and
min\_only in the tables and figures), only one extremum's gradient is retained in the implemented backward; the forward is unchanged. Four
runs (LERP $K{=}32$, 124M, WikiText-103, 100M tokens; seeds 7 and 42) were trained with the code of the five-seed LERP and MinMax--Weight runs, to which the two
modes were added, on
the same RTX 3090 / torch 2.5.1 as their same-seed full and detach controls and compared per validation block on that GPU
(243 blocks, circular moving-block bootstrap, block 16, 4000 resamples; Table~\ref{tab:e1}, Figure~\ref{fig:e1}); CIs
cover evaluation sampling only, and the two seeds are a replication, not a seed distribution. The detach controls were trained with gradient
checkpointing, the full controls and new runs without; step-0 gradients are bitwise identical with and without it on
the checked inputs. Adding the single-channel modes left the full and detach modes unchanged (bit-identical on CPU).

The two channels are not symmetric (Table~\ref{tab:e1}): the maximum-only run recovers almost the whole detach--full gap
on both seeds (0.98--0.99), ending 0.015 and 0.009 nats above full with a softmax-like span and query norm, whereas the
minimum-only run recovers almost none (0.003 and 0.072) and ends with a severely expanded span (layer means in the
hundreds to thousands, like detach) and a gradient-norm rise; against detach it shows no detected endpoint difference at
seed 7 and a small one at seed 42, and its larger span at a slightly lower loss shows that span and loss are not
strictly monotone. The maximum-only gap against full grows slowly from 40M to 100M at seed 7 but not at seed 42, so no
drift is claimed across seeds. Writing $w_j=f(z_j,h)$ with $z_j=s_j-M$ and $h=(M-m)/K$ on a locally smooth region,
and treating the key's own score and the calibration variables as independent inputs,
\[
\frac{\partial w_j}{\partial M}=-\frac{\partial f}{\partial z_j}+\frac1K\frac{\partial f}{\partial h},\qquad
\frac{\partial w_j}{\partial m}=-\frac1K\frac{\partial f}{\partial h},
\]
so the maximum channel carries both the score-origin and the grid-width sensitivity while the minimum channel carries
only the width term; retaining the maximum gradient therefore does not isolate a pure shift-anchor mechanism (index
switches and extremum selection are handled by the implementation; these local expressions are not claimed at every
boundary). Zero-sum residuals of the \emph{implemented} score gradient (median over rows, mean over layers; for detach,
minimum-only and maximum-only the modified backward, not the true derivative of the forward) at 100M are
$5.5{\times}10^{-9}$ (full), $0.250$ / $0.277$ (detach, seeds 7 / 42), $0.257$ / $0.233$ (minimum-only) and
$1.8{\times}10^{-3}$ on both seeds for maximum-only (about 65\% of rows above $10^{-3}$): strict per-row zero-sum is not
necessary for near-full performance here, although the maximum-only residual is still two orders below detach, so an
approximate zero-sum may matter; together with the projection run, the zero-sum constraint alone does not explain the rescue, and shift
invariance is not thereby shown to be irrelevant. The mean raw row minimum at seed 7 is $-19.7$ (full), $+2484$
(detach), $+1700$ (minimum-only) and $-27.1$ (maximum-only): a common shift of the raw scores is distinct from a change
of the relative tail $m-M$. Two seeds of one condition identify the dominant channel in this setting, not a complete
mechanism or a general necessity of the maximum channel.

\paragraph{Local gradient decomposition.} \emph{Definition.} The 20M and 40M checkpoints
of the two full-gradient LERP $K{=}32$ runs (seeds 7 and 42) were loaded without any optimizer step and the loss over the
first 8 WikiText-103 validation blocks was back-propagated block by block ($\mathrm{loss}_b/8$, gradients summed). In the
fixed-upstream analysis the upstream $u=\partial L/\partial P$ of every layer is recorded from the full model, and the
local vector--Jacobian product of that layer's operator ($P=w/\sum w$, $w=\mathrm{reconstruct}(s,\text{mode})$) is
evaluated for each mode at the same scores and the same $u$. With $g_{\mathrm{self}}$ the implemented local VJP of the
detach mode and $g_M$, $g_m$ the contributions of the two extremum channels, $g_{\mathrm{full}}=g_{\mathrm{self}}+g_M+g_m$,
$g_{\mathrm{max\_only}}=g_{\mathrm{self}}+g_M$ and $g_{\mathrm{min\_only}}=g_{\mathrm{self}}+g_m$ by construction (identity
residual 0 at every layer; the $M$ and $m$ terms fall entirely on the arg-max and arg-min keys). \emph{Implementation.}
Boundaries, ties and the clamp follow the implementation. This local VJP is distinct from the whole-model in-graph
backward, where a changed gradient in a later layer alters the upstream of earlier layers, so whole-model differences
are not additive by the local identity. The diagnostic used deterministic algorithms (fp32 attention, bf16 autocast
elsewhere, TF32 off). \emph{Result.}
Table~\ref{tab:graddecomp} reports norm ratios and relative errors $\|g_{\mathrm{mode}}-g_{\mathrm{full}}\|/\|g_{\mathrm{full}}\|$
per layer, averaged over the 12 layers. The maximum-dependent term dominates the omitted calibration correction (norm
ratio 0.36--0.52 against 0.008--0.009 for the minimum-dependent term; norms, not additive shares); preserving it leaves a
layer-averaged local relative gradient error below 1\%, consistent with the near-full training performance of the maximum-only run, while the minimum-only error is indistinguishable from detach. The detach error is concentrated on
the arg-max key (about 99.9--100\% of its squared error, rounded), which describes where detach departs from full, not
the full gradient itself. \emph{Limits.} Not a per-row or per-layer guarantee and not
a bound on whole-model score-gradient error; at 20M no clear loss degradation has appeared while at 40M an early
separation is present (detach $-$ full $+0.118$ at seed 7, $+0.074$ at seed 42), and reading checkpoints of the full
trajectory does not establish a temporal order along the failing trajectories; parameter-gradient norm differences are
not translated into update differences.

\begin{table}[htb]
\centering\scriptsize
\caption{Fixed-upstream decomposition of the score gradient at the 20M and 40M checkpoints of the full-gradient LERP
$K{=}32$ runs (seeds 7 and 42; first 8 WikiText-103 validation blocks). Left: norm of
each term relative to $\|g_{\mathrm{full}}\|$, per layer then averaged over 12 layers (norm ratios, not additive shares).
Right: relative error $\|g_{\mathrm{mode}}-g_{\mathrm{full}}\|/\|g_{\mathrm{full}}\|$ of the implemented local gradient,
per layer then averaged.}
\label{tab:graddecomp}
\adjustbox{max width=\textwidth}{\begin{tabular}{llrrrrrr}
\toprule
 & & \multicolumn{3}{c}{norm ratio to $g_{\mathrm{full}}$} & \multicolumn{3}{c}{relative error of the implemented gradient} \\
\cmidrule(lr){3-5}\cmidrule(lr){6-8}
Seed & ckpt & self & $M$ term & $m$ term & detach & min\_only & max\_only \\
\midrule
7 & 20M & 0.909 & 0.361 & 0.0083 & 0.361 & 0.361 & 0.0083 \\
 & 40M & 0.807 & 0.509 & 0.0092 & 0.509 & 0.509 & 0.0092 \\
42 & 20M & 0.897 & 0.384 & 0.0079 & 0.384 & 0.384 & 0.0079 \\
 & 40M & 0.792 & 0.522 & 0.0092 & 0.522 & 0.522 & 0.0092 \\
\bottomrule
\end{tabular}}
\end{table}

\begin{table}[htb]
\centering\scriptsize
\caption{Single-channel ablation, LERP $K{=}32$ (124M, WikiText-103, 100M tokens; seeds 7 and 42; every run and its same-seed controls evaluated on one RTX 3090). $\Delta$NLL in nats against the same-seed control, paired per validation block, 95\% circular moving-block bootstrap CIs (block 16, 4000 resamples) over 243 blocks: evaluation sampling only, not seed-to-seed variation. Span: layer-mean row span at 100M over 8 blocks. Gap recovery per seed $=(\mathrm{NLL}_{\mathrm{detach}}-\mathrm{NLL}_{x})/(\mathrm{NLL}_{\mathrm{detach}}-\mathrm{NLL}_{\mathrm{full}})$.}
\label{tab:e1}
\adjustbox{max width=\textwidth}{\begin{tabular}{llrllrr}
\toprule
Seed & Mode & NLL @100M & $-$ full & $-$ detach & span & gap recovery \\
\midrule
7 & full (control) & 4.1415 & --- & --- & 21.4 & 1.000 \\
 & detach (control) & 4.9437 & +0.8021 [+0.7849, +0.8202] & --- & 1055.6 & 0.000 \\
 & min\_only & 4.9414 & +0.7998 [+0.7817, +0.8188] & -0.0023 [-0.0074, +0.0031] & 1488.8 & 0.003 \\
 & max\_only & 4.1566 & +0.0151 [+0.0127, +0.0174] & -0.7870 [-0.8047, -0.7703] & 26.5 & 0.981 \\
\midrule
42 & full (control) & 4.1511 & --- & --- & 21.5 & 1.000 \\
 & detach (control) & 4.9311 & +0.7799 [+0.7605, +0.7993] & --- & 860.7 & 0.000 \\
 & min\_only & 4.8751 & +0.7240 [+0.7095, +0.7388] & -0.0559 [-0.0613, -0.0504] & 1738.6 & 0.072 \\
 & max\_only & 4.1602 & +0.0091 [+0.0070, +0.0114] & -0.7708 [-0.7889, -0.7523] & 25.9 & 0.988 \\
\bottomrule
\end{tabular}}
\end{table}

\newpage
\section{Assets, code and reproduction}
\label{app:assets}

\begin{table}[htb]
\centering\scriptsize
\caption{Training runs and evaluated checkpoints of suites A--D and of the further runs of Appendices C--F and~\ref{app:binmean}.}
\label{tab:assets}
\begin{tabular}{@{}>{\raggedright\arraybackslash}p{8.8cm}rr@{}}
\toprule
Suite / group & Runs & Checkpoints \\
\midrule
A: 124M @ 2.5B, FineWebEdu-3B, seed 1337 & 15 & 300 \\
B: 124M @ 250M, FineWebEdu-3B, seed 1337 & 10 & 100 \\
C: 1B @ 100M, WikiText-103, seed 1337 & 12 & 60 \\
D: 124M @ 100M, WikiText-103, seeds 7/42/253/1337/2026 & 130 & 646 \\
\midrule
Suites A--D & 167 & 1,106 \\
\midrule
Historical detach intervention (Appendix~\ref{app:detach}) & 16 & 28 eval.\ records \\
Detach replication, LERP $K{=}32$, seeds 7/42 (Appendix~\ref{app:detach}) & 2 & 10 \\
Single-channel, strict-forward, tail-policy and reduced-precision runs (Appendices~\ref{app:detach},
\ref{app:strict}, \ref{app:taildecomp}, \ref{app:lowprec}) & 12 & 60 \\
Bin-mean control (Appendix~\ref{app:binmean}) & 5 & 25 \\
\bottomrule
\end{tabular}
\end{table}

\paragraph{Runs besides suites A--D.} Nineteen further training runs use the suite-D schedule (124M,
WikiText-103, 100M tokens, five checkpoints each): the two detach-replication runs and the four single-channel runs of
Appendix~\ref{app:detach} (seeds 7 and 42), the four strict-forward runs of Appendix~\ref{app:strict}, the two
tail-policy runs of Appendix~\ref{app:taildecomp}, the two reduced-precision baselines of
Appendix~\ref{app:lowprec} and the five bin-mean control runs of Appendix~\ref{app:binmean} (seed 7; four trained
from scratch and one forward--backward decomposition run). They are reported in those appendices and are not pooled with the five-seed statistics of
suite D. Table~\ref{tab:assets} counts runs and checkpoints; the \rbn{I.runs} runs of Appendix~\ref{app:rb} are listed
in Table~\ref{tab:rb-overview}.

\paragraph{Names in the released code.} Result files, run names, run directories and code fields keep their
original identifiers; the released code maps them to the names used in this paper (\texttt{QRM}, \texttt{FW} and
\texttt{range\_mode=relevant\_zero} denote \FW; \texttt{nearest} denotes \CH; \texttt{\_fullgrad} denotes the full
calibration gradient and does not identify the surrogate). The surrogate of a run is identified by its saved
configuration together with the SHA-256 of its attention code, which the run manifests record.

\paragraph{Manifests and reproduction.} Each suite has a manifest that gives the operator axes, seed, code and
checkpoint path of every run; evaluation and analysis identify runs through it. \texttt{REPRODUCTION\_INDEX.md} in the
released code lists the manifests, code and result files of every run, maps each table and figure
to its source files, and gives the commands for evaluation, analysis, tables and figures; every figure and table in
this paper is regenerated from the stored result files by the scripts listed there.

\section{Extended related work}
\label{app:related}

\textbf{Approximate softmax operators.} Softermax \citep{softermax2021} and I-BERT \citep{ibert2021} replace the
exponential with a base-2 piecewise or integer-polynomial approximation and fine-tune downstream; their scales are static
and never receive a gradient. ConSmax \citep{consmax2024} trains a 6-layer model from scratch with learned per-head
parameters in place of the row reduction, keeps the exact exponential (so its calibration quantities are
optimizer-owned) and attributes its own early instability to non-unit normalization. Two base-2 softmax papers
\citep{zhang2022base2,zhang2023base2hp} study the classifier softmax: the first argues trainability from gradient
structure (a base change scales the cross-entropy Jacobian by $\ln 2$); the second finds that i.i.d.\ error above
$10^{-6}$ injected into an exact-backward pipeline breaks training, whereas our forward error is a deterministic function
of the scores. Our operators keep $\sum_j P_j = 1$ in exact arithmetic by normalizing after reconstruction, a
first-order correctness criterion in \citet{zhang2022base2}. IndexSoftmax \citep{intattention2026} replaces the
exponential with a fixed-window 32-entry lookup table and normalizes in integers, without retraining.

Under per-row calibrated reconstruction the Jacobian terms are neither constant, as in the base-2 case
\citep{zhang2022base2}, nor absorbable into the learning rate. Non-exponential attention (ReLU, sigmoid and linear
attention) is trainable with a stabilizer in each case \citep{wortsman2023relu,shen2023relu,rela2021,ramapuram2024sigmoid,katharopoulos2020};
our LERP at $K{=}1$ is a single-interval affine reconstruction, and no linear-complexity benefit is claimed.

\textbf{Straight-through estimators and calibration gradients.} The straight-through estimator was named and analysed by
\citet{bengio2013ste} for stochastic binary neurons; binarized networks \citep{bnn2016} made it the standard training rule
for hard forwards; \citet{yin2019ste} analyse a two-layer model with binary activations. The temperature of
straight-through Gumbel-softmax \citep{jang2017gumbel,maddison2017concrete} is the closest analogue of our $h$, except
that it is annealed by the practitioner whereas $h$ under MinMax is set by the model's own score geometry. TQT shows that
the TensorFlow fake-quant path zeroes the threshold gradient inside the interval, so the bounds only move outward
\citep{tqt2020}; the PyTorch fake-quantization path computes its observer statistics on a detached tensor
\citep{pytorch_fakequant}. ITA \citep{ita2023}, the closest attention-side instance, computes a base-2 softmax on 8-bit
integers by shifts, observes that above a certain input scale all entries but the row maximum quantize to zero, and tunes
the clipping range by quantization-aware training with that softmax in the loop; it learns a quantizer scale, not a
per-row quantity inside the operator, and does not analyse the backward. LSQ's step-size gradient \citep{lsq2020} is the per-layer analogue of our per-row $\partial w/\partial M$,
$\partial w/\partial m$, but dropping it freezes a parameter, whereas dropping ours changes $\partial L/\partial s$
itself. These observers are running buffers, so the
analogy is structural; what it locates is the graph choice we test: whether the derivative of a per-row extremum used
inside the forward is propagated back to the scores that produced it.

\textbf{\FW, IndexSoftmax and EXAQ.} \FW uses the same construction as IndexSoftmax \citep{intattention2026}, a
fixed window anchored at the row maximum with zero weight below it. Table~\ref{tab:threeops}
lists how the three operators differ as implemented (definitions read from the code: our own for \FW, the PyTorch
implementation released with IntAttention for IndexSoftmax, and the reimplementation of EXAQ \citep{exaq2024} in the
IntAttention artifact; Appendices~\ref{app:rb-index} and~\ref{app:rb-impl}).

\begin{table}[htb]
\centering\scriptsize
\caption{Three lookup operators anchored at the row maximum, as implemented; the constants are read from the code.}
\label{tab:threeops}
\setlength{\tabcolsep}{3pt}\begin{tabular}{@{}p{2.2cm}>{\raggedright\arraybackslash}p{3.0cm}>{\raggedright\arraybackslash}p{3.7cm}>{\raggedright\arraybackslash}p{4.4cm}@{}}
\toprule
 & \FW (this paper) & IndexSoftmax (PyTorch implementation released with IntAttention) & EXAQ (reimplementation in the IntAttention artifact) \\
\midrule
anchor & row maximum & row maximum & row maximum \\
window & fixed, $\tau=6$ nats & fixed, 6.6 nats; table entries are zero from 6.39 nats & lower bound $\max(C^*,\text{tensor minimum})$, $C^*=-1.75\sigma-2.06$ at 3 bits; $\sigma$ and the minimum over the whole attention tensor \\
levels & $K{+}1$ (main results: $K=4,16$) & 32 & 8 at 3 bits \\
rounding & nearest grid centre, or linear interpolation & truncated index & nearest level \\
table values & fp32 exponential & $\mathrm{round}(255\,e^{-6.6n/31})$, 8-bit & exponential of the quantized input \\
below the window & weight 0 & weight 0 & clamped to the window edge \\
normalization & fp32 & float, or UINT8 probabilities $\mathrm{round}(255p)/255$ & float \\
calibration quantity & none beyond the row maximum & none beyond the row maximum & $\sigma$ and tensor minimum, recomputed at every call \\
backward & defined in \S\ref{sec:grad} & none (training-free) & none \\
\bottomrule
\end{tabular}
\end{table}

\clearpage
\section{Additional baselines, interventions and seeds}
\label{app:rb}

The result files of these experiments, the scripts that generate the tables, figures and numbers of this appendix,
and an index from each table and figure to its runs are in the released code.

\FloatBarrier
\subsection{Overview and protocol}
\label{app:rb-overview}

This subsection gives the protocol of the additional experiments. Table~\ref{tab:rb-overview} lists the \rbn{I.runs}
training runs, grouped by the question they answer, with the suite (\S\ref{sec:protocol}) whose protocol each run
follows.

\textbf{Stack and code.} All runs were trained on two RTX~5090 machines, a 2-GPU and an 8-GPU machine (one GPU per run),
with Python 3.12.11, PyTorch 2.13.0+cu132 and CUDA 13.2 (stack S1 in Table~\ref{tab:hardware}). They use the training
code of suite D and of the runs in Appendices~\ref{app:strict} and~\ref{app:lowprec}, with the new operators added; on
CPU this code reproduces the loss and every gradient of the code it extends bit for bit.

\textbf{Pairing and statistics.} Each 100M-token condition is run on seeds 7, 42 and 1337 and paired with a softmax control
of the same seed trained on the same stack; the three paired differences give the mean and a 95\% $t$ interval.
Every 250M-token run is paired with a same-stack 250M softmax control, which reproduces the suite-B softmax
run bit for bit; 2.5B-token runs are paired with the suite-A softmax run, whose re-evaluation on the 8-GPU machine is
bitwise identical to its original evaluation. Single-run intervals are the circular moving-block bootstrap of \S\ref{sec:protocol};
benchmarks and held-out corpora use the methods of Appendix~\ref{app:full}. Intervals are per comparison and not
adjusted for multiplicity. Training took \rbn{I.gpuh} GPU-hours on RTX~5090; the validation, probe and held-out evaluations
took a further \rbn{I.evalh} GPU-hours.

\begin{table}[htb]
\centering\scriptsize
\caption{The additional experiments, grouped by the question they answer: suite protocol, seeds, number of training
runs and training GPU-hours. The total is computed before rounding.}
\label{tab:rb-overview}
\begin{tabular}{@{}p{6.6cm}p{2.6cm}lrr@{}}
\toprule
Question and experiment & Suite protocol & Seeds & Runs & GPU-h \\
\midrule
\multicolumn{5}{@{}p{12.9cm}@{}}{\emph{Controls on the software stack of these experiments (Appendix~\ref{app:rb-overview})}} \\
\quad same-stack softmax controls (100M, three seeds) & D (100M) & 7, 42, 1337 & 3 & 4.1 \\
\quad same-stack softmax control (250M) & B (250M) & 1337 & 1 & 3.5 \\
\addlinespace[3pt]
\multicolumn{5}{@{}p{12.9cm}@{}}{\emph{Can LERP be paired with the fixed window, and does the calibration-gradient result carry over to a static grid? (\ref{app:rb-fwmlerp})}} \\
\quad \FW--LERP, 100M and 250M & B (250M), D (100M) & 7, 42, 1337 & 8 & 17.7 \\
\quad \FW--LERP $K{=}4$ at 2.5B & A (2.5B) & 1337 & 1 & 40.3 \\
\quad row maximum detached under \FW & B (250M), D (100M) & 7, 42, 1337 & 9 & 20.2 \\
\addlinespace[3pt]
\multicolumn{5}{@{}p{12.9cm}@{}}{\emph{Is the surrogate rule a cause of the failure under MinMax? (\ref{app:rb-interv})}} \\
\quad surrogate switch, frozen step & B (250M) & 1337 & 10 & 49.3 \\
\quad the same on the 2.5B schedule & A (2.5B) & 1337 & 3 & 34.9 \\
\quad surrogate switch on three seeds & D (100M) & 7, 42, 253 & 12 & 21.0 \\
\addlinespace[3pt]
\multicolumn{5}{@{}p{12.9cm}@{}}{\emph{Does a window that tracks the scores without being differentiated train? (\ref{app:rb-ema})}} \\
\quad per-head exponential-moving-average (EMA) window & B (250M), D (100M) & 7, 42, 1337 & 8 & 32.5 \\
\addlinespace[3pt]
\multicolumn{5}{@{}p{12.9cm}@{}}{\emph{Do the rounding rules of standard number formats train? (\ref{app:rb-formats})}} \\
\quad rounding rules of five formats, incl. MXFP8, NVFP4 & B (250M), D (100M) & 7, 42, 1337 & 18 & 37.7 \\
\quad bf16 row sum and division & B (250M), D (100M) & 7, 42, 1337 & 4 & 7.9 \\
\addlinespace[3pt]
\multicolumn{5}{@{}p{12.9cm}@{}}{\emph{Does the calibration-gradient result hold for a published operator with tensor-level calibration? (\ref{app:rb-exaq})}} \\
\quad EXAQ, statistics detached or in the graph & B (250M), D (100M) & 7, 42, 1337 & 8 & 31.5 \\
\addlinespace[3pt]
\multicolumn{5}{@{}p{12.9cm}@{}}{\emph{Does a published lookup operator train, with and without reduced-precision normalization? (\ref{app:rb-index})}} \\
\quad IndexSoftmax, (i) float, (ii) UINT8 normalization & B (250M), D (100M) & 7, 42, 1337 & 8 & 21.1 \\
\quad IndexSoftmax at 2.5B & A (2.5B) & 1337 & 1 & 47.3 \\
\addlinespace[3pt]
\multicolumn{5}{@{}p{12.9cm}@{}}{\emph{Which downstream differences persist on a second seed? (\ref{app:rb-seed})}} \\
\quad second softmax seed at 2.5B & A (2.5B) & 7 & 1 & 33.2 \\
\quad second LERP $K{=}16$ seed at 2.5B & A (2.5B) & 7 & 1 & 46.6 \\
\addlinespace[3pt]
\multicolumn{5}{@{}p{12.9cm}@{}}{\emph{Does TF32 in training change the result? (\ref{app:rb-evalset})}} \\
\quad TF32 off in training & D (100M) & 7 & 2 & 5.5 \\
\midrule
Total &  &  & 98 & 454.1 \\
\bottomrule
\end{tabular}
\end{table}

\FloatBarrier
\subsection{\FW--LERP, and the row maximum detached under \FW}
\label{app:rb-fwmlerp}

This subsection asks whether LERP can be paired with the fixed window, and whether the calibration-gradient result
carries over to a static grid.

\textbf{\FW--LERP} combines the \FW window ($\tau=6$, zero tail) with LERP reconstruction and its exact Jacobian; it needs
no straight-through estimator, and the row maximum stays on the graph. At 100M tokens it is within noise of softmax for
$K{=}4$ and $K{=}16$ (Table~\ref{tab:rb-e1}). At 250M tokens $K{=}4$ ends \rbn{E1.250.final} \rbn{E1.250.final.ci};
against suite B's \FW--\CH runs it is \rbn{E1.250.vsW} \rbn{E1.250.vsW.ci} (Weight-STE) and \rbn{E1.250.vsP}
\rbn{E1.250.vsP.ci} (Prob-STE), the latter excluding zero but below the margin we interpret
(Table~\ref{tab:rb-e1b}). At 2.5B tokens it ends \rbn{E1b.final} \rbn{E1b.final.ci}, \rbn{E1b.vsLERP} \rbn{E1b.vsLERP.ci}
from MinMax--LERP $K{=}4$ (below the margin), and \rbn{E1b.vsW} and \rbn{E1b.vsP} from \FW--Weight and \FW--Prob $K{=}4$;
its trajectory is in Figure~\ref{fig:k}b. On the three held-out corpora it is worse than softmax (\rbn{E1b.held.wt},
\rbn{E1b.held.ptb} and \rbn{E1b.held.c4} nats on WikiText-103, PTB and C4), although indistinguishable from MinMax--LERP
$K{=}4$ in validation NLL (Table~\ref{tab:rb-e1bdown}). One 2.5B run.

\begin{table}[htb]
\centering\scriptsize
\caption{\FW--LERP minus the same-stack softmax control: 100M per seed (point estimates; the per-seed intervals are in the
released code), three-seed mean (95\% $t$ interval) and 250M (seed 1337, 95\% block-bootstrap CI).}
\label{tab:rb-e1}
\input{tables/tab_rb_e1}
\end{table}

\begin{table}[htb]
\centering\scriptsize
\caption{\FW--LERP $K{=}4$ at 250M and 2.5B against softmax and the runs of suites B and A (seed 1337, 976 blocks).}
\label{tab:rb-e1b}
\begin{tabular}{@{}lll@{}}
\toprule
Comparison (seed 1337) & NLL & $\dNLL$ [95\% CI] \\
\midrule
\FW--LERP ($K{=}4$), 250M\enspace$-$\enspace suite B \FW--Weight ($K{=}4$) & 3.8924 & $-0.0347$ [$-0.0361$, $-0.0333$] \\
\FW--LERP ($K{=}4$), 250M\enspace$-$\enspace suite B \FW--Prob ($K{=}4$) & 3.8924 & $-0.0028$ [$-0.0039$, $-0.0016$] \\
\FW--LERP ($K{=}4$), 2.5B\enspace$-$\enspace suite A softmax & 3.1997 & $+0.0069$ [$+0.0055$, $+0.0083$] \\
\FW--LERP ($K{=}4$), 2.5B\enspace$-$\enspace suite A MinMax--LERP ($K{=}4$) & 3.1997 & $+0.0022$ [$+0.0007$, $+0.0037$] \\
\FW--LERP ($K{=}4$), 2.5B\enspace$-$\enspace suite A \FW--Weight ($K{=}4$) & 3.1997 & $-0.0358$ [$-0.0375$, $-0.0342$] \\
\FW--LERP ($K{=}4$), 2.5B\enspace$-$\enspace suite A \FW--Prob ($K{=}4$) & 3.1997 & $-0.0124$ [$-0.0137$, $-0.0110$] \\
\bottomrule
\end{tabular}
\end{table}

\begin{table}[htb]
\centering\scriptsize
\caption{\FW--LERP $K{=}4$ at 2.5B: benchmark accuracy (pp; LAMBADA NLL in nats) and held-out NLL differences, statistics of
Appendix~\ref{app:full}.}
\label{tab:rb-e1bdown}
\input{tables/tab_rb_e1b_down}
\end{table}

\begin{figure}[htb]
\centering
\includegraphics[width=\textwidth]{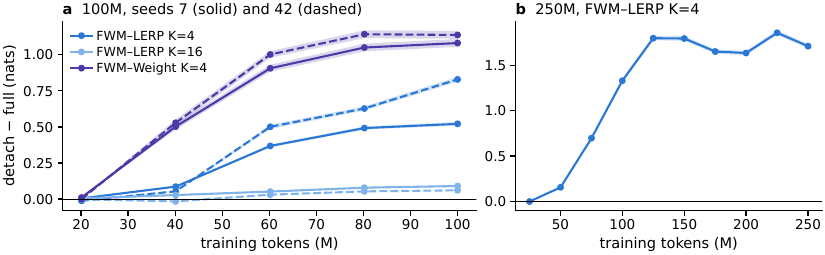}
\caption{Row maximum detached under \FW: NLL of each detached run minus its full-gradient pair at every
evaluated checkpoint, 95\% block-bootstrap bands. (a) 100M tokens; (b) \FW--LERP $K{=}4$, 250M tokens.}
\label{fig:rb-e11}
\end{figure}

\textbf{Row maximum detached under \FW.} With the row maximum detached, the forward is unchanged and the backward
drops the derivative through the row maximum; since the \FW step is constant, only the arg-max term of Eq.~\eqref{eq:full}, the
derivative through the grid origin, is removed. Training is worse for all three operators: every 100M detach-minus-full interval has its
lower bound above $+0.01$ (Table~\ref{tab:rb-e11}, Figure~\ref{fig:rb-e11}). The size depends strongly on $K$ (two-seed
means \rbn{E11.K4.mean} at $K{=}4$ and \rbn{E11.K16.mean} at $K{=}16$ for \FW--LERP), and at 250M tokens the $K{=}4$
penalty is \rbn{E11.250} nats with a mean span of \rbn{E11.250.span}. The detached runs violate the zero-sum identity
(relative row-sum residual \rbn{E11.zs.lo}--\rbn{E11.zs.hi}, against below \rbn{E11.zs.full} for the full runs). The full-gradient \FW--Weight seed-7
run of this appendix reproduces, block by block and bit for bit, the strict-forward run of Appendix~\ref{app:strict} with
the same configuration (difference \rbn{E11.06.vsC2}). The detached variant relies on \texttt{torch.clamp} passing a gradient of 1 at
its bounds (Appendix~\ref{app:rb-impl}).

\begin{table}[htb]
\centering\scriptsize
\caption{Row maximum detached under \FW: detach minus full at the endpoint, and detached vs full values of the
layer-mean span, largest probability $p_{\max}$, fraction of entries below the window, and relative row-sum residual of
the implemented score gradient (8 validation blocks).}
\label{tab:rb-e11}
\begin{tabular}{@{}lrllllll@{}}
\toprule
Operator & Seed & detach $-$ full [95\% CI] & span & $p_{\max}$ & clip fraction & zero-sum residual \\
\midrule
\FW--LERP $K{=}4$ & 7 & \begin{tabular}[c]{@{}l@{}}$+0.5215$\\{[$+0.5096$, $+0.5340$]}\end{tabular} & 660 vs 28 & 0.59 vs 0.19 & 0.90 vs 0.58 & 0.17 vs 7e-9 \\\addlinespace[2pt]
\FW--LERP $K{=}4$ & 42 & \begin{tabular}[c]{@{}l@{}}$+0.8294$\\{[$+0.8124$, $+0.8476$]}\end{tabular} & 785 vs 28 & 0.60 vs 0.20 & 0.88 vs 0.59 & 0.17 vs 7e-9 \\\addlinespace[2pt]
\FW--LERP $K{=}16$ & 7 & \begin{tabular}[c]{@{}l@{}}$+0.0907$\\{[$+0.0860$, $+0.0953$]}\end{tabular} & 57 vs 25 & 0.24 vs 0.20 & 0.63 vs 0.57 & 0.03 vs 6e-9 \\\addlinespace[2pt]
\FW--LERP $K{=}16$ & 42 & \begin{tabular}[c]{@{}l@{}}$+0.0609$\\{[$+0.0580$, $+0.0633$]}\end{tabular} & 56 vs 25 & 0.23 vs 0.20 & 0.61 vs 0.58 & 0.03 vs 6e-9 \\\addlinespace[2pt]
\FW--Weight $K{=}4$ & 7 & \begin{tabular}[c]{@{}l@{}}$+1.0812$\\{[$+1.0547$, $+1.1091$]}\end{tabular} & 762 vs 19 & 0.86 vs 0.11 & 0.97 vs 0.43 & 0.24 vs 6e-9 \\\addlinespace[2pt]
\FW--Weight $K{=}4$ & 42 & \begin{tabular}[c]{@{}l@{}}$+1.1373$\\{[$+1.1106$, $+1.1650$]}\end{tabular} & 810 vs 19 & 0.88 vs 0.11 & 0.99 vs 0.43 & 0.24 vs 6e-9 \\\addlinespace[2pt]
\FW--LERP $K{=}4$, 250M & 1337 & \begin{tabular}[c]{@{}l@{}}$+1.7143$\\{[$+1.6915$, $+1.7386$]}\end{tabular} & 2970 vs 29 & 0.95 vs 0.24 & 0.99 vs 0.67 & 0.25 vs 8e-9 \\\addlinespace[2pt]
\bottomrule
\end{tabular}

\end{table}

\FloatBarrier
\subsection{Interventions from saved checkpoints}
\label{app:rb-interv}

This subsection tests the proposed mechanism by intervention.

\textbf{Surrogate-switch and frozen-step branches.} From the suite-B MinMax $K{=}4$ runs with Weight-STE and Prob-STE we branched at 50M and at 125M
tokens and continued to 250M under the suite-B schedule. A branch starts from the weights, optimizer state and step of
the saved checkpoint. We write W for Weight-STE and P for Prob-STE, and label a
branch X$\to$Y when it uses surrogate X before the branch point and surrogate Y after it. W$\to$W and P$\to$P keep their
rule, W$\to$P and P$\to$W switch it, and ``W$\to$W, frozen step'' continues under Weight-STE with a per-layer constant step $h_0$, the layer's mean native
MinMax width at the branch checkpoint (attention probe, 8 validation blocks), on a window $Kh_0$ below the row maximum
whose grid continues into the tail instead of truncating it. All branches use the strict hard forward. As a check on
the branching procedure, W$\to$W and P$\to$P should retrace the original runs; they end within 0.01 nats of them:
\rbn{E2.50.valid.ww} and \rbn{E2.50.valid.pp} from 50M, \rbn{E2.125.valid.ww} and \rbn{E2.125.valid.pp} from 125M
(Table~\ref{tab:rb-e2chain}).

\begin{figure}[htb]
\centering
\includegraphics[width=\textwidth]{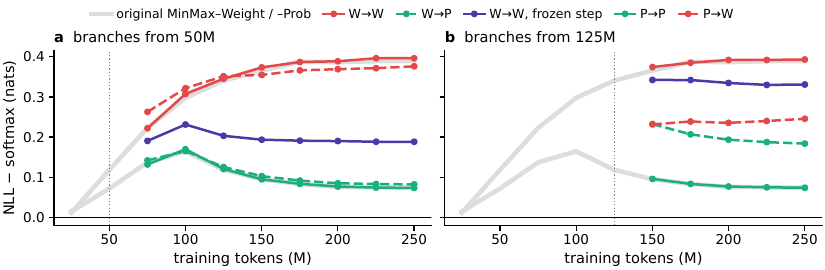}
\caption{Branches from 50M (a) and 125M (b): NLL minus the 250M softmax control at the checkpoints the two
share; grey: the original suite-B MinMax--Weight and MinMax--Prob runs; dotted: branch point. W: Weight-STE; P: Prob-STE;
X$\to$Y: surrogate X before the branch point, surrogate Y after it.}
\label{fig:rb-e2ab}
\end{figure}

\textbf{Results.} Table~\ref{tab:rb-e2ab} and Figure~\ref{fig:rb-e2ab} give the endpoints and trajectories. From 50M, the
switch to Prob-STE brings the span to \rbn{E2.50.span.wp}, against \rbn{E2.50.span.pp} for P$\to$P and
\rbn{E2.50.span.ww} for W$\to$W; the switch to Weight-STE raises it to \rbn{E2.50.span.pw}; with the frozen step it is
\rbn{E2.50.span.freeze}. The frozen-step branch leaves \rbn{E2.50.freezeleft} nats above P$\to$P from 50M and
\rbn{E2.125.freezeleft} from 125M. The frozen-step operator differs from MinMax in its forward (a resolved tail and a
per-layer constant step), so its effect is that of the width together with the resolved tail. One run per branch.

\begin{table}[htb]
\centering\scriptsize
\caption{Branches of the 250M MinMax $K{=}4$ runs (seed 1337): endpoint NLL, difference from the same-stack 250M softmax
control, and layer-mean span and native width $h$ at 250M.}
\label{tab:rb-e2ab}
\begin{tabular}{@{}lllllr@{}}
\toprule
Branch & Rule & NLL, 250M & $-$ softmax control [95\% CI] & span & $h$ \\
\midrule
50M & W$\to$W & 4.2847 & \begin{tabular}[c]{@{}l@{}}$+0.3964$\\{[$+0.3923$, $+0.4005$]}\end{tabular} & 55.8 & 13.96 \\\addlinespace[2pt]
 & W$\to$P & 3.9705 & \begin{tabular}[c]{@{}l@{}}$+0.0822$\\{[$+0.0807$, $+0.0838$]}\end{tabular} & 21.6 & 5.40 \\\addlinespace[2pt]
 & W$\to$W, frozen step & 4.0766 & \begin{tabular}[c]{@{}l@{}}$+0.1883$\\{[$+0.1863$, $+0.1904$]}\end{tabular} & 18.9 & 4.76 \\\addlinespace[2pt]
 & P$\to$P & 3.9616 & \begin{tabular}[c]{@{}l@{}}$+0.0733$\\{[$+0.0717$, $+0.0750$]}\end{tabular} & 21.9 & 5.49 \\\addlinespace[2pt]
 & P$\to$W & 4.2644 & \begin{tabular}[c]{@{}l@{}}$+0.3761$\\{[$+0.3722$, $+0.3800$]}\end{tabular} & 62.9 & 15.73 \\\addlinespace[2pt]
\midrule
125M & W$\to$W & 4.2815 & \begin{tabular}[c]{@{}l@{}}$+0.3932$\\{[$+0.3891$, $+0.3973$]}\end{tabular} & 54.4 & 13.61 \\\addlinespace[2pt]
 & W$\to$P & 4.0720 & \begin{tabular}[c]{@{}l@{}}$+0.1837$\\{[$+0.1818$, $+0.1856$]}\end{tabular} & 30.3 & 7.58 \\\addlinespace[2pt]
 & W$\to$W, frozen step & 4.2192 & \begin{tabular}[c]{@{}l@{}}$+0.3309$\\{[$+0.3276$, $+0.3341$]}\end{tabular} & 37.1 & 11.44 \\\addlinespace[2pt]
 & P$\to$P & 3.9620 & \begin{tabular}[c]{@{}l@{}}$+0.0737$\\{[$+0.0721$, $+0.0753$]}\end{tabular} & 22.1 & 5.52 \\\addlinespace[2pt]
 & P$\to$W & 4.1339 & \begin{tabular}[c]{@{}l@{}}$+0.2457$\\{[$+0.2429$, $+0.2484$]}\end{tabular} & 48.4 & 12.09 \\\addlinespace[2pt]
\bottomrule
\end{tabular}

\smallskip
\begin{minipage}{\linewidth}\scriptsize W: Weight-STE; P: Prob-STE. X$\to$Y: surrogate X before the branch point, surrogate Y after it.\end{minipage}
\end{table}

\begin{table}[htb]
\centering\scriptsize
\caption{Branch analysis at 250M: validity of the unswitched branches, and the share of the W$\to$W $-$ P$\to$P endpoint
gap that each intervention removes or reproduces.}
\label{tab:rb-e2chain}
\begin{tabular}{@{}lllllll@{}}
\toprule
Branch & W$\to$W $-$ W$^{\dagger}$ & P$\to$P $-$ P$^{\dagger}$ & W$\to$P removes & P$\to$W reproduces & frozen step removes & frozen $-$ P$\to$P \\
\midrule
50M & $+0.0072$ & $-0.0002$ & 97.3\% & 93.7\% & 64.4\% & $+0.1150$ \\
125M & $+0.0040$ & $+0.0001$ & 65.6\% & 53.8\% & 19.5\% & $+0.2573$ \\
\bottomrule
\end{tabular}

\smallskip
\begin{minipage}{\linewidth}\scriptsize W: Weight-STE; P: Prob-STE. X$\to$Y: surrogate X before the branch point, surrogate Y after it. $^{\dagger}$The original suite-B run with that surrogate: MinMax--Weight for W$^{\dagger}$, MinMax--Prob for P$^{\dagger}$.\end{minipage}
\end{table}

\textbf{Three seeds.} From the suite-D MinMax $K{=}4$ runs of seeds 7, 42 and 253 at 40M tokens the four switch branches were run to 100M. The switch effect has the same
sign on every seed (Table~\ref{tab:rb-e2d}). Each switched branch ends near the branch that used its new rule throughout
(three-seed means \rbn{E2d.WPmPP.mean} \rbn{E2d.WPmPP.ci} and \rbn{E2d.PWmWW.mean} \rbn{E2d.PWmWW.ci}); three per-seed
intervals of these two contrasts exclude zero and are listed in the table.

\begin{table}[htb]
\centering\scriptsize
\caption{Surrogate switch on three seeds: validation NLL at 100M of the four branches from 40M (suite-D MinMax $K{=}4$ parents) and their paired
differences per seed (95\% block-bootstrap CI) and over the three seeds (mean, 95\% $t$ interval).}
\label{tab:rb-e2d}
\input{tables/tab_rb_e2d}

\smallskip
\begin{minipage}{\linewidth}\scriptsize W: Weight-STE; P: Prob-STE. X$\to$Y: surrogate X before the branch point, surrogate Y after it.\end{minipage}
\end{table}

\begin{figure}[htb]
\centering
\includegraphics[width=\textwidth]{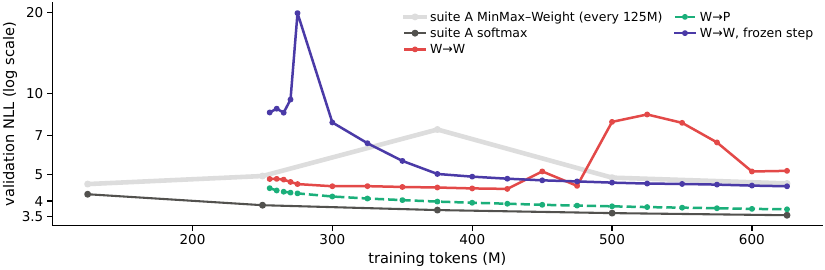}
\caption{Validation NLL of the three branches from 250M on the 2.5B schedule (log scale), with the suite-A
MinMax--Weight $K{=}4$ and softmax runs (evaluated every 125M). W: Weight-STE; P: Prob-STE; X$\to$Y: surrogate X before
the branch point, surrogate Y after it.}
\label{fig:rb-e2c}
\end{figure}

\textbf{Branches on the 2.5B schedule.} From the suite-A MinMax--Weight $K{=}4$ run at 250M we continued the 2.5B schedule to
625M as W$\to$W, W$\to$P and W$\to$W with a frozen step. W$\to$W collapses again, with its highest
NLL \rbn{E2c.ww.peak} at \rbn{E2c.ww.peakm}M; the original run, evaluated every 125M, has its highest evaluated NLL
\rbn{E2c.orig.peak} at \rbn{E2c.orig.peakm}M. The collapse recurs at a later point than in the original run, and
we report this group as supporting evidence. W$\to$P shows no loss spike and ends at \rbn{E2c.wp.end} (\rbn{E2c.wp.vsww}
nats from W$\to$W). The frozen-step branch jumps to \rbn{E2c.freeze.first} at its first checkpoint
(\rbn{E2c.freeze.firstm}M) and recovers to \rbn{E2c.freeze.end} (Table~\ref{tab:rb-e2c}, Figure~\ref{fig:rb-e2c}). One run per branch.

To see how much of this jump comes from changing the operator alone, we took the three MinMax--Weight $K{=}4$
checkpoints from which a frozen-step branch was started (250M tokens of the 2.5B run; 50M and 125M tokens of the 250M
run) and evaluated each one twice with its weights unchanged: once with the MinMax forward that the run had used up to
that point, and once with the frozen-step forward that the ``W$\to$W, frozen step'' branch uses from that point on (the
per-layer constant step $h_0$ measured at that checkpoint). The frozen-step forward gives a
higher NLL than the MinMax forward by \rbn{S0.A250.diff} nats \rbn{S0.A250.ci} on the 250M checkpoint of the 2.5B run, by
\rbn{S0.B50.diff} \rbn{S0.B50.ci} on the 50M checkpoint of the 250M run, and by \rbn{S0.B125.diff} \rbn{S0.B125.ci} on the
125M checkpoint of the 250M run. Changing the operator therefore changes the function before any training step, and by
the same amount in the 2.5B case and in the 125M case.

The two cases differ in the training after the branch point. We compare each frozen-step branch with the W$\to$W branch
from the same checkpoint, which sees the same batches. For the branches from the 250M checkpoint of the 2.5B run, the
gradient norm of the frozen-step branch at its first optimizer step is \rbn{S0.A250.gnratio} times that of the W$\to$W
branch, and its training loss exceeds that of the W$\to$W branch by \rbn{S0.A250.dl1} nats at the first step and by
\rbn{S0.A250.dlN} nats after \rbn{S0.nsteps} steps (about \rbn{S0.tokM}M tokens). For the branches from the 125M checkpoint
of the 250M run, the same excess falls from \rbn{S0.B125.dl1} to \rbn{S0.B125.dlN} nats over the same \rbn{S0.nsteps}
steps. Two differences between the checkpoints are candidate causes: the frozen step $h_0$ (layer mean \rbn{S0.A250.h0} nats for
the 250M checkpoint of the 2.5B run, \rbn{S0.B125.h0} and \rbn{S0.B50.h0} for the 125M and 50M checkpoints of the 250M
run) and the position on the learning-rate schedule.

\begin{table}[htb]
\centering\scriptsize
\caption{Branches from 250M on the 2.5B schedule (seed 1337, to 625M): highest validation NLL and where it occurs,
and the NLL at 625M.}
\label{tab:rb-e2c}
\begin{tabular}{@{}llll@{}}
\toprule
Branch from 250M & highest NLL & at & NLL at 625M \\
\midrule
W$\to$W & 8.361 & 525M & 5.1690 \\
W$\to$P & 4.460 & 255M & 3.7233 \\
W$\to$W, frozen step & 19.842 & 275M & 4.5315 \\
original MinMax--Weight (every 125M) & 7.354 & 375M & --- \\
\bottomrule
\end{tabular}

\smallskip
\begin{minipage}{\linewidth}\scriptsize W: Weight-STE; P: Prob-STE. X$\to$Y: surrogate X before the branch point, surrogate Y after it.\end{minipage}
\end{table}

\FloatBarrier
\subsection{A per-head EMA window}
\label{app:rb-ema}

This subsection tests a semi-static calibration: a window that tracks the scores but is not differentiated. Each attention layer keeps one fp32 buffer per head (144
windows); its width $\tau_h$ is an exponential moving average (EMA; decay 0.999) of the mean row span $M-m$ over rows $i\ge1$,
updated on every training micro-batch forward before it is used, frozen at evaluation and never on the autograd graph.
The operator is \FW with $\tau$ replaced by $\tau_h$: step $\tau_h/K$, zero tail below $-\tau_h$, \CH at $K{=}4$, with
Weight-STE or Prob-STE and the row maximum on the graph.

Both surrogates end well above softmax, and the Weight-minus-Prob difference is much smaller than under per-row MinMax
(Table~\ref{tab:rb-e4}): \rbn{E4.WmP.mean} \rbn{E4.WmP.ci} at 100M against \rbn{E4.hist.100}, and \rbn{E4.WmP.250} at
250M against \rbn{E4.hist.250}. The EMA width follows the span (Table~\ref{tab:rb-e4probe}); here it is the Prob-STE runs
whose span grows. The per-row MinMax gaps are those of suite D (100M) and suite B (250M). The experiment covers $K{=}4$
and one EMA design.

\begin{table}[htb]
\centering\scriptsize
\caption{EMA window at $K{=}4$: difference from the same-stack softmax control and Weight-STE minus Prob-STE, per seed
(point estimates), three-seed mean (95\% $t$) and 250M (95\% CI).}
\label{tab:rb-e4}
\input{tables/tab_rb_e4}

\smallskip
\begin{minipage}{\linewidth}\scriptsize $^{\dagger}$Per-row MinMax, \CH $K{=}4$, from suite D (100M) and suite B (250M).\end{minipage}
\end{table}

\begin{table}[htb]
\centering\scriptsize
\caption{EMA window at the endpoints: layer-mean span, EMA width (mean over layers and heads) and fraction of entries below
the window.}
\label{tab:rb-e4probe}
\begin{tabular}{@{}lrrrrr@{}}
\toprule
Surrogate & Seed & Tokens & span & EMA width & clip fraction \\
\midrule
Weight-STE & 7 & 100M & 51.4 & 52.7 & 0.136 \\
Weight-STE & 42 & 100M & 51.7 & 52.7 & 0.138 \\
Weight-STE & 1337 & 100M & 54.1 & 54.0 & 0.134 \\
Weight-STE & 1337 & 250M & 1089.1 & 1074.4 & 0.156 \\
Prob-STE & 7 & 100M & 800.0 & 799.6 & 0.166 \\
Prob-STE & 42 & 100M & 684.7 & 686.1 & 0.164 \\
Prob-STE & 1337 & 100M & 677.3 & 668.6 & 0.167 \\
Prob-STE & 1337 & 250M & 41743.4 & 41016.9 & 0.172 \\
\bottomrule
\end{tabular}
\end{table}

\FloatBarrier
\subsection{Reduced-precision format baselines}
\label{app:rb-formats}

This subsection uses the rounding rules of standard number formats as baselines. Each baseline rounds $e^z$,
$z=s-M$, with the rule of Table~\ref{tab:rb-formats} and normalizes in fp32; the backward is straight-through on the
rounding with the fp32 exponential derivative and the normalization Jacobian at the rounded weights (native autograd for
BF16-exp). The bf16 chain computes the exponential, the row sum and the division in bf16 with native autograd.
Each format baseline applies the format's rounding rule to $e^z$ in fp32; the cited recipes themselves keep
softmax at its original precision. The MXFP8-style scale exponent is rounded up, unlike the OCP MX rule
\citep{rouhani2023mx}; the NVFP4-style tensor scale is the constant 2688 because every row's maximum key has $e^0=1$. The
count of non-zero values is, for a key in the block that holds the row maximum, the number of distinct non-zero values
the rule gives to $e^z$; for BF16 it is \rbn{E0c.bf16exp.levels}, from enumerating every bf16 $z\le0$.

The three-seed means lie between \rbn{E5.mean.lo} and \rbn{E5.mean.hi} nats and every $t$ interval includes zero; the bf16
chain gives \rbn{E5n.mean} \rbn{E5n.ci} (Table~\ref{tab:rb-e5}). The seed-7 BF16-exp and FP8 runs are the
Appendix~\ref{app:lowprec} runs, paired with the suite-D softmax run of that seed; both were trained on stack S3
(Table~\ref{tab:hardware}), so these two three-seed means mix two training stacks.

\begin{table}[htb]
\centering\scriptsize
\caption{Rounding rules of the format baselines (applied in fp32), the number of distinct non-zero values for a key in
the row-maximum block, and the $z$ at and below which the rule gives 0.}
\label{tab:rb-formats}
\begin{tabular}{@{}lp{7.0cm}rr@{}}
\toprule
Format & Rule on $e^z$ & non-zero values & zero for $z\le$ \\
\midrule
BF16-exp & $\mathrm{bf16}(\exp(\mathrm{bf16}(z)))$ & 1021 & $-93$ \\
FP8-E4M3 & E4M3 of $e^z$, round to nearest even & 56 & $-6.9315$ \\
MXFP8-style & blocks of 32 keys, scale $2^{e}$ with $e$ the smallest integer such that block max $\le 448\cdot2^{e}$; E4M3 elements & 120 & $-12.4767$ \\
NVFP4-style & tensor scale $2688/\max w$ (always 2688), blocks of 16 keys with E4M3 block scale, E2M1 elements & 7 & $-3.1781$ \\
UINT8 & $\mathrm{round}(255\,e^z)/255$ (rounds $e^z$, not $p$) & 255 & $-6.2344$ \\
\bottomrule
\end{tabular}
\end{table}

\begin{table}[htb]
\centering\scriptsize
\caption{Format baselines minus the same-stack softmax control: 100M per seed (point estimates), three-seed mean (95\%
$t$) and 250M (95\% CI).}
\label{tab:rb-e5}
\input{tables/tab_rb_e5}

\smallskip
\begin{minipage}{\linewidth}\scriptsize $^{\ast}$The Appendix~\ref{app:lowprec} run minus the suite-D softmax run of seed 7, both trained on stack S3
(Table~\ref{tab:hardware}).\end{minipage}
\end{table}

\FloatBarrier
\subsection{EXAQ}
\label{app:rb-exaq}

This subsection trains a published operator whose calibration is tensor-level. EXAQ at 3 bits
shifts each row by its maximum and quantizes $x=s-M$ to 8 uniform levels on $[lo,0]$ with
$lo=\max(C^*,\min x)$, $C^*=-1.75\sigma-2.06$, where $\sigma$ and the minimum are taken over the whole attention tensor
of the layer and recomputed at every call (not calibrated once and frozen); keys below $lo$ are clamped to the lowest
level, not zeroed, and $P=\mathrm{softmax}(\hat x)$ in fp32. The backward is Prob-STE with a LERP surrogate on the same
grid; with the statistics detached (d) $lo$ and the step are constants, in the graph (g) the gradient also reaches every
valid score of the layer through $\sigma$ and the minimum.

Condition d is worse than g (Table~\ref{tab:rb-e3b}): \rbn{E3b.dg.mean} at 100M (three seeds) and \rbn{E3b.dg.250} at
250M. Both are clearly worse than softmax, and with the statistics detached the 250M span reaches \rbn{E3b.d.250span}
against \rbn{E3b.g.250span} in the graph (Table~\ref{tab:rb-e3bprobe}).

\begin{table}[htb]
\centering\scriptsize
\caption{EXAQ 3-bit with the window statistics detached (d) or in the graph (g): differences from the same-stack softmax
control, d $-$ g, and g minus IndexSoftmax (i) with float normalization (per seed: point estimates).}
\label{tab:rb-e3b}
\input{tables/tab_rb_e3b}
\end{table}

\begin{table}[htb]
\centering\scriptsize
\caption{EXAQ at the endpoints: layer-mean span, window statistics and the clamped tail (fraction of entries and their
softmax probability mass).}
\label{tab:rb-e3bprobe}
\begin{tabular}{@{}lrrrrrrp{1.3cm}p{1.5cm}@{}}
\toprule
Stats & Seed & Tokens & span & $\sigma$ & lower bound & step & clamped entries & softmax mass clamped \\
\midrule
detach & 7 & 100M & 113.7 & 66.95 & -119.22 & 17.032 & 0.170 & 0.0015 \\
detach & 42 & 100M & 114.9 & 68.03 & -121.11 & 17.302 & 0.180 & 0.0008 \\
detach & 1337 & 100M & 118.1 & 68.39 & -121.75 & 17.392 & 0.166 & 0.0008 \\
detach & 1337 & 250M & 8258.6 & 3814.77 & -6677.90 & 953.986 & 0.231 & 0.0004 \\
graph & 7 & 100M & 25.5 & 13.44 & -25.58 & 3.655 & 0.177 & 0.0002 \\
graph & 42 & 100M & 25.6 & 13.08 & -24.95 & 3.564 & 0.188 & 0.0001 \\
graph & 1337 & 100M & 25.8 & 13.75 & -26.12 & 3.731 & 0.179 & 0.0003 \\
graph & 1337 & 250M & 33.1 & 19.73 & -36.59 & 5.226 & 0.250 & 0.0000 \\
\bottomrule
\end{tabular}
\end{table}

\FloatBarrier
\subsection{IndexSoftmax}
\label{app:rb-index}

This subsection trains a published lookup operator, with and without reduced-precision normalization.

\textbf{Implementation of the two published operators.} Our IndexSoftmax follows the PyTorch implementation released with
IntAttention \citep{intattention2026}. With floating-point normalization its forward pass reproduces their
floating-point function (\texttt{idx\_softmax\_fp}) bit for bit on fp32 scores, on CPU and GPU. The UINT8 variant adds the output normalization of
their integer path, $\mathrm{round}(255\cdot e_j/\sum e)/255$, and reproduces the integer codes of their
integer function (\texttt{idx\_softmax}) on integer-valued scores. Their full pipeline also quantizes $Q$, $K$ and $V$; here these stay in floating point and
the 6.6 threshold is applied in floating point, so the UINT8 variant is a composition of two parts of their code. The original EXAQ source code is no longer publicly available; our EXAQ implementation follows the
EXAQ paper \citep{exaq2024} and the reimplementation released in the IntAttention artifact repository. Its forward pass reproduces that
reimplementation bit for bit when both run on the same machine (stack S3, Table~\ref{tab:hardware}); on the RTX~5090
machines our outputs are within $2.0{\times}10^{-6}$ of the reference outputs computed on S3, with the same grid indices.
As in that reimplementation, the window statistics are recomputed at every call. Both operators are defined for
inference; the backward rules (Prob-STE with the row maximum on the graph, and for EXAQ the choice between detaching and
differentiating $\sigma$ and the minimum) are ours. The agreement tests in our released code load the IntAttention code from a path given on the
command line.

\textbf{Definition.} With $d_j=\min(M-s_j,6.6)$ and $n_j=\lfloor31d_j/6.6\rfloor$ (truncated, as in their released code; the formula in \citet{intattention2026} rounds this
index to nearest), the unnormalized weight is the 8-bit
table entry $\mathrm{round}(255\,e^{-6.6n_j/31})$; entries 30 and 31 are 0, so a key is set to zero from $d_j\ge30\cdot6.6/31
\approx6.39$ nats, and the lookup returns the cell end nearer the row maximum. (i) normalizes in floating point; (ii)
rounds the normalized probability to an 8-bit code, does not renormalize the row, and gives an all-zero row when the
integer row sum reaches $510\cdot255$ (at least 510 keys). The surrogate is \FW--LERP on a 32-level grid over 6.6 nats
with exact exponentials.

\begin{figure}[htb]
\centering
\includegraphics[width=\textwidth]{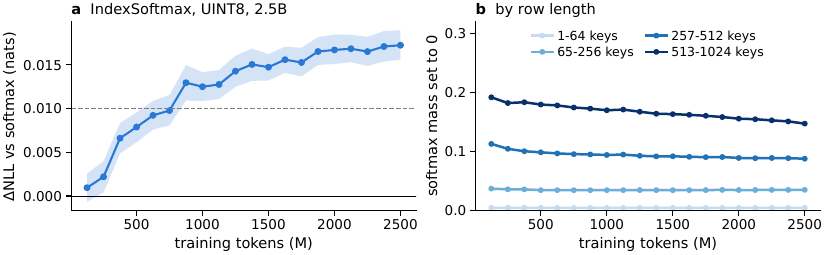}
\caption{IndexSoftmax (UINT8) at 2.5B. (a) NLL minus suite-A softmax along training, 95\% block-bootstrap band
(dashed: the 0.01-nat margin). (b) Softmax probability mass on keys the operator sets to zero, by row length (layer
mean, 8 validation blocks).}
\label{fig:rb-e3a}
\end{figure}

\textbf{Results.} Both variants are within noise of softmax at 100M and 250M (Table~\ref{tab:rb-e3a}); the UINT8 mean at
100M, \rbn{E3a.ii.mean} \rbn{E3a.ii.ci}, has an interval that includes zero. The 2.5B run uses the UINT8 variant, whose 250M
difference from softmax is \rbn{E3a.ii.250}. It ends \rbn{E3a25.final}
\rbn{E3a25.final.ci} (one seed), level with \FW--Prob $K{=}4$ (\rbn{E3a25.vsFWMP}; Table~\ref{tab:rb-e3a25}), and its gap
is still rising slowly at the end of training (\rbn{E3a25.at1000} at 1B and \rbn{E3a25.at2000} at 2B tokens;
Figure~\ref{fig:rb-e3a}). The gap is consistent with the probability mass that the UINT8 normalization drops
(Appendix~\ref{app:rb-zeroed}); at 2.5B only the UINT8 variant was trained. Evaluated with exact
softmax or float normalization in place of its own operator, the trained model loses \rbn{E3a25.swap.lo}--\rbn{E3a25.swap.hi}
nats. On the benchmarks its BLiMP difference is \rbn{E3a25.blimp} pp \rbn{E3a25.blimp.ci} (Table~\ref{tab:rb-e3adown}).

\begin{table}[htb]
\centering\scriptsize
\caption{IndexSoftmax, (i) float normalization and (ii) UINT8 probabilities: differences from the same-stack softmax
control and (ii) $-$ (i) (per seed: point estimates).}
\label{tab:rb-e3a}
\input{tables/tab_rb_e3a}
\end{table}

\begin{table}[htb]
\centering\scriptsize
\caption{IndexSoftmax (UINT8) at 2.5B: the endpoint under other forwards with the weights unchanged, and against three
suite-A runs (976 blocks).}
\label{tab:rb-e3a25}
\begin{tabular}{@{}p{4.6cm}lll@{}}
\toprule
Forward at evaluation (weights unchanged) & NLL & $-$ native & $-$ suite A softmax (or as stated) \\
\midrule
native: IndexSoftmax (ii), UINT8, as trained & 3.2100 & --- & $+0.0172$ [$+0.0156$, $+0.0189$] \\
IndexSoftmax (i), float normalization & 3.2188 & $+0.0088$ [$+0.0082$, $+0.0094$] & $+0.0260$ [$+0.0244$, $+0.0277$] \\
exact fp32 softmax & 3.2177 & $+0.0077$ [$+0.0072$, $+0.0083$] & $+0.0250$ [$+0.0234$, $+0.0266$] \\
native\enspace$-$\enspace suite A \FW--Prob ($K{=}4$) & 3.2100 &  & $-0.0020$ [$-0.0036$, $-0.0005$] \\
native\enspace$-$\enspace suite A \FW--Weight ($K{=}16$) & 3.2100 &  & $+0.0134$ [$+0.0115$, $+0.0152$] \\
native\enspace$-$\enspace suite A MinMax--LERP ($K{=}4$) & 3.2100 &  & $+0.0125$ [$+0.0107$, $+0.0143$] \\
\bottomrule
\end{tabular}
\end{table}

\begin{table}[htb]
\centering\scriptsize
\caption{IndexSoftmax (UINT8) at 2.5B: benchmark (pp; LAMBADA NLL in nats) and held-out NLL differences, statistics of
Appendix~\ref{app:full}.}
\label{tab:rb-e3adown}
\input{tables/tab_rb_e3a_down}
\end{table}

\FloatBarrier
\subsection{Probability mass set to zero}
\label{app:rb-zeroed}

This subsection reports the probability mass that each truncating operator removes. Table~\ref{tab:rb-e0b} gives, at
the trained endpoints of these experiments, the softmax probability mass on keys each operator sets to zero and the row sum of
its output; Table~\ref{tab:rb-e0blen} splits the UINT8 normalization of IndexSoftmax by row length, where the lost mass
grows with the number of keys. In the 2.5B IndexSoftmax (UINT8) run, the fraction of all-zero rows among rows with
more than 512 keys peaks at \rbn{E0b.allzero.max}\% (near \rbn{E0b.allzero.maxB}B tokens) and is \rbn{E0b.allzero.end}\%
at the endpoint.

\begin{table}[htb]
\centering\scriptsize
\caption{Softmax probability mass on keys the operator sets to zero, and output row sum, at the trained endpoints
(layer mean, 8 validation blocks; a range is over the runs at that length).}
\label{tab:rb-e0b}
\begin{tabular}{@{}lll@{}}
\toprule
Operator (range over the runs at that length) & softmax mass set to zero & row sum \\
\midrule
\FW--LERP $K{=}4$, 100M & 0.0094--0.0100 & 1.0000 \\
\FW--LERP $K{=}16$, 100M & 0.0100--0.0103 & 1.0000 \\
\FW--LERP $K{=}4$, 250M & 0.0136 & 1.0000 \\
\FW--LERP $K{=}16$, 250M & 0.0134 & 1.0000 \\
\FW--Weight $K{=}4$, 100M & 0.0064--0.0069 & 1.0000 \\
\FW--Weight $K{=}4$, row maximum detached, 100M & 0.0013--0.0019 & 1.0000 \\
IndexSoftmax, float normalization, 100M & 0.0067--0.0071 & 1.0000 \\
IndexSoftmax, UINT8, 100M & 0.1161--0.1259 & 0.9136--0.9212 \\
IndexSoftmax, float normalization, 250M & 0.0103 & 1.0000 \\
IndexSoftmax, UINT8, 250M & 0.1187 & 0.9210 \\
IndexSoftmax, UINT8, 2.5B & 0.1023 & 0.9336 \\
EMA window, Weight-STE, 100M & 0.0010--0.0013 & 1.0000 \\
EMA window, Prob-STE, 100M & 0.0003--0.0004 & 1.0000 \\
FP8-E4M3, 100M & 0.0038--0.0039 & 1.0000 \\
NVFP4-style, 100M & 0.0254--0.0255 & 1.0000 \\
FP8-E4M3, 250M & 0.0050 & 1.0000 \\
NVFP4-style, 250M & 0.0275 & 1.0000 \\
UINT8 on $e^z$, 100M & 0.0076--0.0081 & 1.0000 \\
MXFP8-style, 100M & 2e-7 & 1.0000 \\
UINT8 on $e^z$, 250M & 0.0099 & 1.0000 \\
MXFP8-style, 250M & 3e-7 & 1.0000 \\
\bottomrule
\end{tabular}
\end{table}

\begin{table}[htb]
\centering\scriptsize
\caption{IndexSoftmax with UINT8 normalization, by row length: softmax probability mass on keys set to zero, output row
sum, and fraction of all-zero rows.}
\label{tab:rb-e0blen}
\setlength{\tabcolsep}{3pt}\begin{tabular}{@{}>{\raggedright\arraybackslash}p{3.1cm}*{9}{>{\raggedleft\arraybackslash}p{0.95cm}}@{}}
\toprule
 & \multicolumn{2}{c}{1--64 keys} & \multicolumn{2}{c}{65--256 keys} & \multicolumn{2}{c}{257--512 keys} & \multicolumn{3}{c@{}}{513--1024 keys} \\
\cmidrule(lr){2-3}\cmidrule(lr){4-5}\cmidrule(lr){6-7}\cmidrule(l){8-10}
 & mass set to zero & row sum & mass set to zero & row sum & mass set to zero & row sum & mass set to zero & row sum & all-zero rows \\
\midrule
IndexSoftmax UINT8, 100M, seed 7 & 0.0039 & 0.9990 & 0.0379 & 0.9830 & 0.1094 & 0.9359 & 0.1778 & 0.8697 & 3.39e-6 \\
IndexSoftmax UINT8, 100M, seed 42 & 0.0039 & 0.9991 & 0.0393 & 0.9820 & 0.1123 & 0.9331 & 0.1803 & 0.8677 & 0 \\
IndexSoftmax UINT8, 100M, seed 1337 & 0.0039 & 0.9991 & 0.0377 & 0.9833 & 0.1047 & 0.9378 & 0.1651 & 0.8800 & 1.70e-6 \\
IndexSoftmax UINT8, 250M & 0.0042 & 0.9995 & 0.0379 & 0.9837 & 0.1090 & 0.9392 & 0.1680 & 0.8787 & 7.97e-5 \\
IndexSoftmax UINT8, 2.5B & 0.0042 & 0.9999 & 0.0347 & 0.9860 & 0.0876 & 0.9568 & 0.1471 & 0.8941 & 2.21e-3 \\
softmax control, seed 7$^{\ast}$ & 0.0039 & 0.9991 & 0.0371 & 0.9834 & 0.1021 & 0.9390 & 0.1634 & 0.8816 & 1.70e-6 \\
softmax control, seed 42$^{\ast}$ & 0.0037 & 0.9992 & 0.0369 & 0.9838 & 0.1046 & 0.9374 & 0.1678 & 0.8776 & 0 \\
softmax control, seed 1337$^{\ast}$ & 0.0038 & 0.9991 & 0.0374 & 0.9833 & 0.1039 & 0.9375 & 0.1677 & 0.8771 & 1.70e-6 \\
\bottomrule
\end{tabular}

\smallskip
\begin{minipage}{\linewidth}\scriptsize No row of up to 512 keys is all-zero. $^{\ast}$The UINT8 forward applied to the scores of the trained softmax
controls.\end{minipage}
\end{table}

\FloatBarrier
\subsection{A second seed at 2.5B tokens}
\label{app:rb-seed}

\begin{figure}[htb]
\centering
\includegraphics[width=\textwidth]{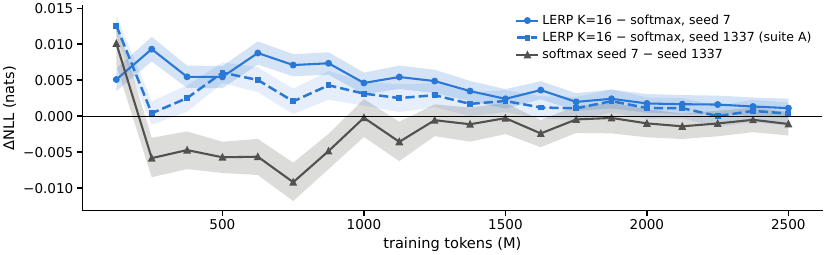}
\caption{MinMax--LERP $K{=}16$ minus softmax along 2.5B tokens on seed 7 and on seed 1337 (suite A), and
softmax seed 7 minus seed 1337; 95\% block-bootstrap bands.}
\label{fig:rb-e8}
\end{figure}

This subsection asks which downstream differences persist on a second seed. A softmax run and a
MinMax--LERP $K{=}16$ run were trained on seed 7 with the suite-A protocol. The softmax run ends
\rbn{E8a.vs1337} \rbn{E8a.vs1337.ci} from the seed-1337 softmax run; LERP $K{=}16$ minus softmax is \rbn{E8.val.s7}
\rbn{E8.val.s7.ci} on seed 7 and \rbn{E8.val.s1337} \rbn{E8.val.s1337.ci} on seed 1337 (Figure~\ref{fig:rb-e8}).
Table~\ref{tab:rb-e8} sets the two seeds side by side with the difference between the two softmax runs. That difference
excludes zero on LAMBADA accuracy (\rbn{E8.ss.lambada} pp \rbn{E8.ss.lambada.ci}) and on WikiText-103 and PTB
(\rbn{E8.ss.wt} and \rbn{E8.ss.ptb} nats), not on C4 (\rbn{E8.ss.c4} \rbn{E8.ss.c4.ci}). On WikiText-103 the LERP
$K{=}16$ difference changes sign between the seeds (\rbn{E8c.held.wt} on seed 7, \rbn{E8p.held.wt} on seed 1337). On
BLiMP the LERP $K{=}16$ difference is \rbn{E8.blimp.s7.3} pp \rbn{E8.blimp.s7.3ci} on seed 7 and \rbn{E8.blimp.s1337.3} pp
\rbn{E8.blimp.s1337.3ci} on seed 1337: the same sign at about half the size, with a seed-7 interval that includes zero. Softmax and LERP $K{=}16$ have two seeds each; the other
near-baseline conditions have one.

\begin{table}[htb]
\centering\scriptsize
\caption{Second seed at 2.5B: benchmark accuracy (pp; LAMBADA NLL in nats) and held-out NLL differences, statistics of
Appendix~\ref{app:full}.}
\label{tab:rb-e8}
\input{tables/tab_rb_e8_down}
\end{table}

\FloatBarrier
\subsection{Operation counts, bit widths and a roofline estimate}
\label{app:rb-cost}

This subsection gives operation counts, bit widths and a cost estimate for deployment. Table~\ref{tab:rb-e0a}
counts the operations of a minimal fused forward per key, read from the training code; the training code itself materializes
the full score tensor and computes surrogates, so its cost is higher than this count. Table~\ref{tab:rb-e0d} applies the published
roofline inputs of FlashAttention-4 (FA4) \citep{zadouri2026fa4} to a lookup operator: an estimate from published numbers, to
which the lookup's own integer and floating-point work would be added. A single-pass kernel anchors the grid at the
running maximum and computes a different function (Appendix~\ref{app:rb-evalset}); a kernel that first finds the row
maximum is bounded at \rbn{E0d.twopass}$\times$ FA4's cycle count. IntAttention, an integer attention pipeline built on
IndexSoftmax, runs up to $2.0\times$ faster than a conventional INT8 attention pipeline that keeps a floating-point softmax
and up to $3.7\times$ faster than FP16 attention on an \mbox{RK3588S2} (Armv8) CPU \citep{intattention2026}; the $3.7\times$
includes the gain from INT8 matrix products.
Table~\ref{tab:rb-e0aacc} lists the measured 2.5B accuracy of the same operators.

\begin{table}[htb]
\centering\scriptsize
\caption{Per-key operations of a minimal fused forward (counts read from the code).}
\label{tab:rb-e0a}
\begin{tabular}{@{}p{2.3cm}p{4.5cm}p{6.1cm}@{}}
\toprule
Operator & Per row & Per key \\
\midrule
exact softmax & max and sum (fusable online), 1 reciprocal & subtract, exponential, multiply \\
MinMax--LERP & max and min before any key is indexed (two passes), reciprocal of the step, $K{+}1$ exponentials, sum, 1 reciprocal & subtract, scale, floor and clamp, fractional part, 2 table reads, interpolation, multiply \\
\FW--\CH & max, sum, 1 reciprocal; constant table, no exponential at run time & subtract, fused multiply-add, round to an index, 1 tail compare, 1 table read, multiply \\
\FW--LERP & as \FW--\CH & subtract, fused multiply-add, floor, fractional part, tail compare, 2 table reads, interpolation, multiply \\
IndexSoftmax (float) & max, integer sum, 1 reciprocal & subtract, scale with saturation, truncate, 1 table read (8-bit), multiply \\
IndexSoftmax (UINT8) & as above & as above, plus a scale by $255/\mathrm{sum}$ and a round to an 8-bit code \\
\bottomrule
\end{tabular}
\end{table}

\begin{table}[htb]
\centering\scriptsize
\caption{FA4 roofline applied to a lookup operator: cycles per $128^3$ forward tile under ideal overlap (an estimate).}
\label{tab:rb-e0d}
\begin{tabular}{@{}p{4.3cm}rrrrlp{2.6cm}@{}}
\toprule
Implementation (per 128$^3$ tile) & matmul & exp unit & shared mem. & bound & vs FA4 & function \\
\midrule
FA4, exact softmax, single pass & 1024 & 1024 & 768 & 1024 & 1.00$\times$ & exact softmax \\
lookup operator, single pass, grid anchored at the running max & 1024 & 0 & 768 & 1024 & 1.00$\times$ & different from the trained operator \\
lookup operator, two passes (full-row max first) & 1536 & 0 & 1280 & 1536 & 1.50$\times$ & same as the trained operator \\
\bottomrule
\end{tabular}
\end{table}

\begin{table}[htb]
\begin{minipage}[t]{0.32\textwidth}\centering\scriptsize
\caption{Levels and index bits of a $K$-interval grid.}
\label{tab:rb-e0c}
\begin{tabular}{@{}rrr@{}}
\toprule
$K$ & levels & index bits \\
\midrule
2 & 3 & 2 \\
3 & 4 & 2 \\
4 & 5 & 3 \\
7 & 8 & 3 \\
8 & 9 & 4 \\
15 & 16 & 4 \\
16 & 17 & 5 \\
31 & 32 & 5 \\
32 & 33 & 6 \\
63 & 64 & 6 \\
64 & 65 & 7 \\
\bottomrule
\end{tabular}

\end{minipage}\hfill
\begin{minipage}[t]{0.64\textwidth}\centering\scriptsize
\caption{Measured accuracy at 2.5B tokens (976 blocks).}
\label{tab:rb-e0aacc}
\begin{tabular}{@{}ll@{}}
\toprule
Operator (124M @ 2.5B, seed 1337) & $\dNLL$ vs softmax [95\% CI] \\
\midrule
MinMax--LERP K4 & $+0.0047$ [$+0.0031$, $+0.0063$] \\
MinMax--LERP K16 & $+0.0003$ [$-0.0013$, $+0.0020$] \\
MinMax--LERP K32 & $+0.0019$ [$+0.0005$, $+0.0034$] \\
\FW--Prob K4 & $+0.0193$ [$+0.0179$, $+0.0207$] \\
\FW--Weight K4 & $+0.0427$ [$+0.0410$, $+0.0445$] \\
\FW--Weight K16 & $+0.0039$ [$+0.0023$, $+0.0055$] \\
\FW--LERP K4 & $+0.0069$ [$+0.0055$, $+0.0083$] \\
IndexSoftmax, UINT8 & $+0.0172$ [$+0.0156$, $+0.0189$] \\
\bottomrule
\end{tabular}
\end{minipage}
\end{table}

\FloatBarrier
\subsection{Evaluation settings and TF32-off training}
\label{app:rb-evalset}

This subsection measures the sensitivity of the results to the evaluation setting and to TF32 in training. Six existing
\CH endpoints were re-evaluated under each setting of Table~\ref{tab:rb-e6}; no setting changes the NLL by more than
\rbn{E6.maxabs} nats, and the same evaluation on the RTX~3090 differs from the RTX~5090 by at most \rbn{E6.max3090}. The
largest change, \rbn{E6.online.max} \rbn{E6.online.max.ci} (\FW--Weight $K{=}4$ at 2.5B), is the single-pass \FW variant, which processes keys in blocks of 128 and anchors each block at the running
maximum: a different function from the trained operator whenever the row maximum lies after the first block. The
two-pass variant first takes the full-row maximum and only accumulates the denominator in blocks; it is the same function.
Training without TF32 changes the seed-7 \FW--Weight $K{=}4$ endpoint by \rbn{E7.fwm} \rbn{E7.fwm.ci} against the
same code with TF32, and the MinMax--Weight $K{=}4$ endpoint by \rbn{E7.mm} against the Appendix~\ref{app:strict} run,
trained with the same code (mean only); one seed and two conditions.

\begin{table}[htb]
\centering\scriptsize
\caption{Evaluation settings: NLL change from the reference evaluation (TF32 off, batch 1, fp32 scores, native forward).
Endpoints: suite A at 2.5B and two suite-D seed-7 runs audited in Appendix~\ref{app:strict} (100M). Last column: largest
difference between the RTX~3090 and RTX~5090 evaluation over all settings.}
\label{tab:rb-e6}
\setlength{\tabcolsep}{4pt}\begin{tabular}{@{}llrrrrrrr@{}}
\toprule
Endpoint & ref NLL & TF32 on & batch 8 & bf16 scores & direct & two-pass & single-pass & 3090 $|\Delta|$ \\
\midrule
A \FW--Prob $K{=}4$ & 3.2121 & $+0.0000$ & $+0.0000$ & $+0.0002$ & $+0.0001$ & $-0.0001$ & $-0.0002$ & 0.0002 \\
A \FW--Weight $K{=}4$ & 3.2355 & $+0.0000$ & $+0.0002$ & $+0.0003$ & $-0.0001$ & $+0.0000$ & $+0.0041$ & 0.0003 \\
A \FW--Weight $K{=}16$ & 3.1967 & $-0.0002$ & $-0.0001$ & $+0.0000$ & $+0.0000$ & $-0.0001$ & $+0.0001$ & 0.0001 \\
A MinMax--Weight $K{=}4$ & 4.0831 & $+0.0005$ & $+0.0004$ & $+0.0010$ & $+0.0001$ & $+0.0001$ & --- & 0.0002 \\
C.2 \FW--Weight $K{=}4$ & 4.1691 & $+0.0000$ & $+0.0000$ & $+0.0015$ & $-0.0001$ & $+0.0002$ & $+0.0018$ & 0.0002 \\
C.2 MinMax--Weight $K{=}4$ & 4.3615 & $+0.0003$ & $-0.0001$ & $+0.0003$ & $-0.0001$ & $-0.0001$ & --- & 0.0002 \\
\bottomrule
\end{tabular}
\end{table}

\FloatBarrier
\subsection{Implementation definitions and verification}
\label{app:rb-impl}

\textbf{Conventions.} Every operator except the native softmax disables autocast inside attention and computes in fp32;
the hard forward is computed without gradient and the straight-through compositions are strict (the forward value equals
the hard value bit for bit). Prob-STE is $P=\sg(P_{\mathrm{hard}})+P_{\mathrm{sur}}-\sg(P_{\mathrm{sur}})$, Weight-STE
$w=\sg(w_{\mathrm{hard}})+w_{\mathrm{sur}}-\sg(w_{\mathrm{sur}})$ followed by normalization. ``Row maximum in the graph''
means that $z=s-M$ sends $-\partial L/\partial z$ to the arg-max key. Where the code differs from what the names suggest:
IndexSoftmax sets keys to zero from 6.39 nats, not 6.6, and takes the cell end nearer the maximum; its UINT8 rows are not
renormalized and an all-zero row is not handled; EXAQ's detached variant detaches only the window statistics, its window
is tensor-level and its tail is clamped, not zeroed; the EMA window is per layer and head and is updated on every training
micro-batch before use; the UINT8 format baseline rounds $e^z$, not the probability; the NVFP4-style tensor scale is
constant; the MXFP8-style scale exponent is rounded up.

\textbf{Verification.} The tests of the training mechanics and of the operators (\FW, the EMA window, the format
rules, IndexSoftmax and EXAQ) passed on stack S3 (Table~\ref{tab:hardware}) and on both RTX~5090 machines; the test
scripts are in the released code.
The agreement of IndexSoftmax and EXAQ with the code released in the IntAttention artifact is described in
Appendix~\ref{app:rb-index}. Detaching the row maximum under \FW relies on \texttt{torch.clamp} passing a gradient of 1
at its bounds, which holds in PyTorch 2.13.0 (used here) but not in PyTorch 2.14.1; it was checked on the CPU and the GPU
before every run of that experiment, and the released code pins PyTorch 2.13.0.

\FloatBarrier

\section{A reconstruction rule with a consistent backward}
\label{app:binmean}

This appendix reports a control operator, not a method of this paper: a \emph{bin-mean reconstruction} (``Centroid'' in the
result files) whose backward is the exact derivative of its forward, trained and analysed under the suite-D protocol.

\subsection{Definition}
\label{app:binmean-def}
Take the same calibration (MinMax or \FW) and the same \CH rounding grid as in \S\ref{sec:grad}: $K{+}1$ bins centred
on the grid points, with the \FW zero tail belonging to no bin. For each bin $b$ the representative weight is the
arithmetic mean of the exponential over its members, $w^{*}_b=\frac{1}{|b|}\sum_{j\in b}e^{z_j}$, and every key in the
bin receives $\hat w_j=w^{*}_{b(j)}$. Bin mass is therefore conserved, $\sum_{j\in b}\hat w_j=\sum_{j\in b}e^{z_j}$, and
the normalizer equals $Z=\sum_j e^{z_j}$ (under \FW, the $Z$ of the window). The backward is the exact derivative of
this forward, $\partial w^{*}_b/\partial z_j=e^{z_j}/|b|$, so no surrogate is needed; the calibration statistics do not
enter the normalized gradient (full and detached calibration gradients differ by at most $2\times10^{-15}$ in fp64). With
$e_j=e^{z_j}$, $g_j=\partial L/\partial P_j$, $\bar g_b$ the unweighted mean of $g_j$ over bin $b$ and $\langle g\rangle$
the $\hat P$-weighted mean,
\begin{equation}
\text{bin-mean:}\quad \frac{\partial L}{\partial s_j}=\frac{e_j\,(\bar g_b-\langle g\rangle)}{\hat W},
\qquad
\text{exact softmax:}\quad \frac{\partial L}{\partial s_j}=\frac{e_j\,(g_j-\langle g\rangle)}{Z}.
\label{eq:binmean}
\end{equation}
The bin-mean gradient replaces each key's upstream gradient by the bin average and drops the within-bin contrast
$p_j(g_j-\bar g_b)$: keys in one bin can only be pushed in the same direction, with magnitude proportional to $e_j$.
A second variant (Centroid-2 in the result files) approximates the log-mean-exp of a bin from the mean and variance of
its $z$ values, $w^{*}_b\approx e^{\bar z_b+\sigma_b^2/2}$; under \FW it is not separable from bin-mean (training endpoint
difference $-0.001$ nats), and in wide bins its representative can exceed the largest member's weight and overflow in
fp32, so it is not reported further. Verification: the forward matches an independent NumPy
reference to a relative difference of $7\times10^{-15}$, bin mass is conserved to $6\times10^{-16}$, the closed-form gradient matches autograd to
$1\times10^{-15}$, and every existing operator path of the code root is bitwise unchanged.

\subsection{Training from scratch}
\label{app:binmean-train}
Table~\ref{tab:binmean-train} gives the 100M-token endpoints at 124M on WikiText-103 (seed 7, $K{=}4$, suite-D protocol;
every run evaluated on one RTX 3090, 243 blocks, batch 1, TF32 off; 95\% circular block-bootstrap CIs, block 16, 4000
resamples, evaluation sampling only). The paired differences on the same blocks are \FW bin-mean $-$ \FW--Prob (strict)
$+0.1418$ [+0.1361, +0.1480], \FW bin-mean $-$ \FW--Weight $+0.1059$ [+0.1011, +0.1109], MinMax bin-mean $-$ MinMax--Prob (strict) $+0.1882$ [+0.1839, +0.1926] and MinMax
bin-mean $-$ MinMax--Weight $+0.1093$ [+0.1037, +0.1151]. These gaps exceed the seed SD of the corresponding suite-D cells (0.007--0.013)
by more than a factor of ten. At 20M tokens the four bin-mean runs sit at the level of \FW--Weight ($+0.010$ to $+0.014$);
the separation occurs between 20M and 40M, stops widening after 60M under \FW and widens to the endpoint under MinMax.
At 100M (layer mean over 8 validation blocks) the \FW bin-mean model has row span 15.1 nats and softmax entropy 4.62,
against 26.2 / 3.74 for the softmax model and 26.6 / 3.88 for \FW--Prob. The control runs come from the training stacks
recorded in Appendix~\ref{app:protocol} (Prob-STE on an RTX 5090, the others on the local RTX 3090); the stack
differences audited there are about 0.01 nats or less, far below these gaps.

\begin{table}[htb]
\centering\footnotesize
\caption{Bin-mean reconstruction trained from scratch: 124M @ 100M tokens, WikiText-103, seed 7, $K{=}4$; $\dNLL$ vs the
same-seed softmax run (NLL 4.1368), paired per validation block, 95\% circular block-bootstrap CI (evaluation sampling only).
Control rows are the same-seed runs of Appendix~\ref{app:strict} and suite D, evaluated on the same GPU.}
\label{tab:binmean-train}
\begin{tabular}{lrl}
\toprule
Condition (seed 7, $K{=}4$, 100M) & $\dNLL$ & 95\% CI \\
\midrule
\FW bin-mean & $+0.1382$ & [+0.1338, +0.1429] \\
MinMax bin-mean & $+0.3338$ & [+0.3267, +0.3413] \\
\midrule
\FW--Prob (strict forward, Appendix~\ref{app:strict}) & $-0.0036$ & control \\
\FW--Weight (historical) & $+0.0323$ & control \\
MinMax--Prob (historical) & $+0.1452$ & control \\
MinMax--Weight (historical) & $+0.2245$ & control \\
LERP $K{=}4$ & $+0.0942$ & control \\
\bottomrule
\end{tabular}
\end{table}

\subsection{Forward--backward decomposition}
\label{app:binmean-decomp}
A supplementary run keeps the forward bitwise equal to the strict hard \CH forward of \FW--Prob and replaces only the
LERP branch of the backward by the bin-mean branch, $P=\sg(P^N)+\big(P^C-\sg(P^C)\big)$, where $P^C$ are the bin-mean
probabilities on the same bins. Before training, the whole-model loss was verified bitwise equal to the strict \FW--Prob
code root on CPU and GPU, and the backward equal to the vector--Jacobian product of the bin-mean probabilities.
Table~\ref{tab:binmean-2x2} gives the $2{\times}2$ (\FW, $K{=}4$, seed 7, 100M). With a joint block bootstrap, the
backward effect (\CH-forward/bin-mean-backward $-$ strict \FW--Prob) is $+0.1455$ [+0.1397, +0.1514], the forward effect (\FW bin-mean $-$
\CH-forward/bin-mean-backward) is $-0.0037$ [-0.0070, -0.0004], and the backward share is 1.03 [1.00, 1.05]: backward-dominated under the
pre-registered rule. The forward effect is below the pre-declared 0.02-nat stack-difference threshold and is not
interpreted. At initialization the layer-0 $Q$/$K$ projection gradient norms (one backward on the first micro-batch)
are 0.099 for softmax, 0.084 for strict \FW--Prob, 0.0154 for \FW bin-mean and 0.0159 for the supplementary run, with the
same proportions in every layer: at initialization the gradient magnitude is set by the backward alone. The
supplementary run's trajectory ($+0.018$ / $+0.114$ / $+0.162$ / $+0.146$ / $+0.142$ at 20/40/60/80/100M) nearly coincides
with \FW bin-mean's ($+0.014$ / $+0.105$ / $+0.155$ / $+0.139$ / $+0.138$); its 100M geometry is span 13.7, entropy 4.69.

\begin{table}[htb]
\centering\footnotesize
\caption{Forward--backward decomposition at \FW, $K{=}4$, seed 7, 100M tokens: $\dNLL$ vs softmax. The
(bin-mean forward, LERP-surrogate backward) cell was not trained.}
\label{tab:binmean-2x2}
\begin{tabular}{lrr}
\toprule
forward $\backslash$ backward & LERP surrogate (Prob-STE) & bin-mean \\
\midrule
hard \CH & $-0.0036$ & $+0.1419$ [+0.1377, +0.1464] \\
bin-mean & not trained & $+0.1382$ \\
\bottomrule
\end{tabular}
\end{table}

\subsection{Gradient alignment on frozen models}
\label{app:binmean-grad}
Table~\ref{tab:binmean-grad} evaluates, on the five suite-D softmax endpoints (4 validation blocks each, layer mean),
the score gradient of each backward rule against the exact softmax gradient under the same scores, $V$ and upstream
gradient taken from the exact model (LERP's score gradient under a fixed upstream gradient equals Prob-STE's). The
bin-mean norm ratio nearly equals its cosine: the exact gradient minus an approximately orthogonal component. As a
descriptive association, at \FW $K{=}4$ the cosine ordering of the three backward rules (Prob-STE $>$ Weight-STE $>$
bin-mean) matches the ordering of the same-seed training endpoints ($-0.004<+0.032<+0.138$); three points, not a
mechanism.

\begin{table}[htb]
\centering\footnotesize
\caption{Score-gradient alignment with the exact softmax gradient on the five suite-D softmax endpoints: cosine /
norm ratio, layer mean over 4 validation blocks.}
\label{tab:binmean-grad}
\begin{tabular}{lrlll}
\toprule
Calibration & $K$ & bin-mean cos / norm ratio & Prob-STE cos / norm ratio & Weight-STE cos \\
\midrule
FWM & 2 & 0.452 / 0.459 & 0.675 / 1.006 & 0.535 \\
FWM & 4 & 0.633 / 0.641 & 0.908 / 1.008 & 0.837 \\
FWM & 8 & 0.754 / 0.758 & 0.975 / 1.007 & 0.953 \\
FWM & 16 & 0.833 / 0.838 & 0.993 / 1.007 & 0.987 \\
FWM & 64 & 0.923 / 0.930 & 0.999 / 1.006 & 0.999 \\
MinMax & 4 & 0.362 / 0.363 & 0.577 / 0.639 & 0.455 \\
MinMax & 16 & 0.618 / 0.618 & 0.863 / 0.887 & 0.777 \\
\bottomrule
\end{tabular}
\end{table}

\subsection{Forward fidelity on frozen models}
\label{app:binmean-fwd}
Table~\ref{tab:binmean-fwd} replaces all 12 attention layers of the five suite-D softmax endpoints by each
reconstruction and reports the full-validation $\dNLL$ (five-seed mean). Under \FW the bin-mean forward is better than
\CH at every $K$ (same direction on all five seeds) and worse than LERP; under MinMax it is slightly worse than \CH for
$K\le16$. The training ordering is the reverse of the forward-fidelity ordering, consistent with
Appendix~\ref{app:binmean-decomp}.

\begin{table}[htb]
\centering\footnotesize
\caption{Frozen-model substitution, all 12 layers, five suite-D softmax endpoints: $\dNLL$ vs the exact model, five-seed
mean.}
\label{tab:binmean-fwd}
\begin{tabular}{lrrrr}
\toprule
Calibration & $K$ & \CH & bin-mean & LERP \\
\midrule
FWM & 4 & +0.0280 & +0.0208 & +0.0045 \\
FWM & 8 & +0.0073 & +0.0045 & +0.0005 \\
FWM & 16 & +0.0019 & +0.0008 & +0.0001 \\
FWM & 32 & +0.0005 & +0.0002 & +0.0000 \\
MinMax & 4 & +0.628 & +0.639 & +0.691 \\
MinMax & 16 & +0.119 & +0.122 & +0.140 \\
\bottomrule
\end{tabular}
\end{table}

\subsection{Scope}
\label{app:binmean-scope}
One seed, $K{=}4$, 100M tokens; the decomposition was run under \FW only; the (bin-mean forward, LERP backward) cell was
not trained; the control runs come from different training stacks; no kernel was implemented, so no speed or storage
conclusion is drawn. The MinMax bin-mean failure does not pass through the calibration--gradient coupling (the
calibration statistics do not enter the gradient), so under MinMax a second channel exists besides the one discussed
in \S\ref{sec:res-geom}: averaging within wide bins removes the contrast signal. The bin-mean rule is MSE-optimal in the
weight domain (Lloyd's centroid condition), which did not translate into training quality.

\end{document}